\documentclass[lettersize,journal]{IEEEtran}
\usepackage{amsmath,amsfonts,amssymb}
\usepackage{array}
\usepackage[caption=false,font=footnotesize,labelfont=rm,textfont=rm]{subfig}
\usepackage{textcomp}
\usepackage{xpatch}
\usepackage{diagbox}
\usepackage{url}
\usepackage{verbatim}
\usepackage{graphicx}
\usepackage{cite}
\usepackage[table]{xcolor}
 \usepackage{threeparttable}
\usepackage{makecell}
\usepackage{hyperref}
\usepackage{amssymb}
\usepackage{multirow}
\usepackage{stmaryrd}
\usepackage{booktabs}
\usepackage{xcolor}
\usepackage{epstopdf}
\usepackage{wasysym}
\usepackage{pifont}
\usepackage[ruled,vlined]{algorithm2e}
\usepackage{algorithmic}
\usepackage{amsthm}
\usepackage{indentfirst}
\usepackage{ragged2e}
\usepackage{soul}
\usepackage{tikz}
\usetikzlibrary{patterns}
\usepackage{xcolor}
\newcommand{\shadecell}[1]{%
\tikz[baseline=(node.base)]{
\node[
    inner sep=1.5pt,
    rectangle,
    rounded corners=1pt,
    fill=gray!8,
    pattern=north east lines,
    pattern color=gray!35
] (node) {#1};
}}
\usepackage[dvipsnames]{xcolor}
\usepackage{hyperref}
 \usepackage{mathtools}
\SetKwFor{SSS}{Server:}{}{endfor}
\SetKwFor{BBB}{$S_2$:}{}{endfor}
\SetKwFor{clients}{Node:}{}{endfor}
\newtheorem{assumption}{Assumption}

\newtheorem{Remark}{Discussion}

\newtheorem{thm}{\bf Theorem}

\usepackage{color, xcolor}
\usepackage{wrapfig}
\begin{document}
\title{FL-OA: A Byzantine-Robust Federated Learning Framework  with  Outsourced Auditing for \\ Intelligent Devices}
\author{Hongliang Zhang,  Zhongyuan Yu, Fenghua Xu, Teng Hu,  Jian Meng, Jiguo Yu,~\IEEEmembership{Fellow,~IEEE}
\thanks{This work was partially supported by NSF of China under Grants 62272256 and 62202250,  and the Shandong Province Youth Innovation Team Project under Grant 2024KJH032. ($\textit{Corresponding author}$: $\textit{Jiguo Yu}$)}
\thanks{H. Zhang is with the Key Laboratory of Computing Power Network and Information Security, Ministry of Education, Shandong Computer Science Center, Qilu University of Technology (Shandong Academy of Sciences), Jinan, 250353, China, Email: b1043123004@stu.qlu.edu.cn.}
\thanks{Z. Yu is with the College of computer science and technology, China University of Petroleum,  Qingdao, 266580, China, Email:  yuzhy24601@gmail.com.}
\thanks{F. Xu is with the Cyber Security Institute, University of Science and Technology of China,  Hefei, 230026, China, Email: nstlxfh@gmail.com.}
\thanks{T. Hu is with Institute of Computer Application, China Academy of Engineering Physics, Mianyang, 621900, China,  Email: mailhuteng@foxmail.com.}
\thanks{J. Meng is with Inspur Software Group Ltd., Jinan, 250101,    China, Email: mengjian@inspur.com}
\thanks{J. Yu is with  School of Computer Science and Engineering, University of Electronic Science and Technology of China, Chengdu, 611731, China, and also with the Big Data Institute, Qilu University of Technology, Jinan, 250353, China, Email: jiguoyu@sina.com.}
}

\markboth{Journal of \LaTeX\ Class Files,~Vol.~14, No.~8, August~2021}%
{Shell \MakeLowercase{\textit{et al.}}: A Sample Article Using IEEEtran.cls for IEEE Journals}
\maketitle
\begin{abstract}
Federated learning (FL) enables multiple intelligent devices  to collaboratively train a high-accuracy model without sharing raw data.
However, due to its distributed nature, FL is vulnerable to Byzantine attacks. 
Existing defense methods rely on strong assumptions, such as the proportion of malicious devices not exceeding 50\%, or the server having an additional root dataset  that matches the training task.
Moreover, they show limited efficacy as they overlook $(i)$ the divergence among benign  updates and $(ii)$ the curse of dimensionality involved in comparing two high-dimensional updates.
To solve these concerns, we propose FL-OA,  a Byzantine-robust  federated learning   framework utilizing outsourced auditing.
In FL-OA, the server collaborates with third-party organization that holds an additional root dataset to perform outsourced auditing, thereby enabling the server to achieve robust aggregation without strong assumptions.
Additionally,  FL-OA introduces a gradient ascent step and a correction term during local training to mitigate the divergence  among benign updates, and designs a parameter importance indicator to extract critical parameters for auditing, alleviating the curse of dimensionality.
We further provide a detailed theoretical analysis of FL-OA.
Extensive experiments  demonstrate that   FL-OA outperforms existing defense methods against  Byzantine attacks.
\end{abstract}
 
\section{INTRODUCTION}
\sloppy
With the rapid  development of internet of things, the intelligent device   market has expanded rapidly. 
They offer advanced intelligence and communication capabilities.
By collecting large amounts of data generated by these devices, their service provider  (i.e., server) can train artificial intelligence (AI) models.
However, such data contains sensitive user information, which poses potential privacy risks.
\begin{wrapfigure}{t}{0.22\textwidth}
  \centering   
  {
      \label{3301642}  \includegraphics[width=0.9\linewidth]{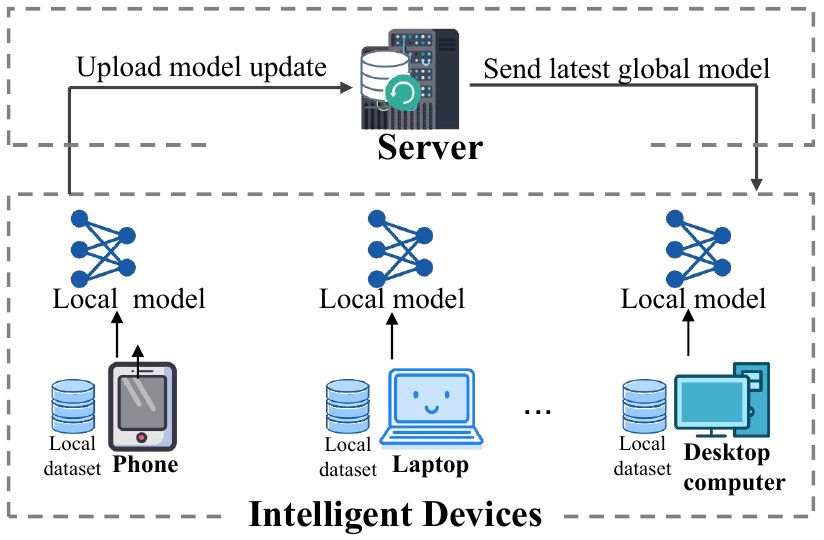}
  }
  \vspace{-5pt}
\caption{A typical FL architecture. The server,  operated by a technology company (e.g., Huawei or Apple), coordinates global model training, while multiple intelligent devices (e.g., cell phones, laptops, and desktops) collaboratively train the model in edge  scenarios.
}
  \vspace{-5pt}
\label{3141141}
\end{wrapfigure}
To this end, federated learning (FL) has emerged as a distributed computing paradigm that enables multiple intelligent devices to collaboratively train a global AI model, and has been widely  applied in scenarios  such as healthcare, autonomous driving, and financial  systems.
As shown in Fig. \ref{3141141},  each  device trains its local model using its own local data and submits its model update to  the server without explicitly sharing  raw training data.
However, due to the distributed nature of FL, its training process is inherently vulnerable to Byzantine attacks, in which some  devices controlled by an adversary (malicious devices)  send malicious   updates to the server,  degrading the performance of the global model \cite{wang2022threats}\cite{kumar2023impact}.
\fussy

To resist  Byzantine attacks in FL,  several  Byzantine-robust strategies \cite{NIPS2017f4b9ec30,huang2023multi,krauss2023mesas,11202428,yin2018byzantine,FLtrust,miao2022privacy,10475552,9798217,10713463,xia2024byzantine} have been proposed.
These strategies  are mainly divided into two angles \cite{uddin2025systematic}.
The first angle  leverages statistical knowledge  to audit  model  updates submitted by  devices \cite{NIPS2017f4b9ec30,huang2023multi,krauss2023mesas,11202428,yin2018byzantine}.  
Nevertheless, these methods  typically  assume that the majority of devices  are benign.
If this assumption is not met, these methods  fail to effectively resist Byzantine attacks.
The second type  assumes that the  server holds  an additional root dataset,  and  uses that dataset as a benchmark   to identify malicious updates \cite{FLtrust,miao2022privacy,10475552,9798217,10713463,xia2024byzantine}.
Compared to the first angle, the second  is more accurate in auditing malicious behavior from devices and does  not require the assumption about the proportion of benign devices. 
However, due to  privacy and regulations,  constructing the additional dataset required by the second type  is difficult in real-world scenarios, as the server cannot directly obtain data samples from intelligent devices \cite{10458320}.
This inspires the  question: \textbf{\textit{Can the  server collaborate with a third-organization holding an additional root dataset to outsource the auditing process, enabling Byzantine-robust aggregation without directly acquiring that dataset? }}

Besides the above mentioned, in the practical world, the Byzantine-robust strategies are confronted with the following challenges.
$i)$ \textit{Divergence of model updates}.
In distributed scenarios, FL data is usually heterogeneous, i.e., Non-Independent and Identically Distributed (Non-IID) \cite{lu2024federated,huang2024federated,10492865,10891500}.
Existing study  \cite{sun2023fedspeed} observes that Non-IID data leads to different optimal solutions across  devices, which causes the divergence among their model updates.
However, this divergence  interferes with the differences calculated using  cosine similarity or Euclidean distance, making it difficult for defense strategies   (e.g., \cite{NIPS2017f4b9ec30,huang2023multi,krauss2023mesas,11202428,yin2018byzantine,10648998,11421423,10495004,FLtrust,miao2022privacy,10475552,9798217,10713463,xia2024byzantine}) to distinguish whether the difference comes from divergence  or Byzantine attacks.
$ii)$ \textit{Curse of dimensionality}.
We notice that  existing defense strategies (e.g., \cite{NIPS2017f4b9ec30,huang2023multi,krauss2023mesas,11202428,yin2018byzantine,10648998,11421423,10495004,FLtrust,miao2022privacy,10475552,9798217,10713463,xia2024byzantine}) evaluate the differences  among  model updates  using metrics such as cosine similarity or Euclidean distance.
However, some studies  \cite{1093}\cite{beyer1999nearest} observe  that calculating the difference  between two high-dimensional vectors leads to  a curse of dimensionality,  i.e., the maximum difference between two vectors decreases with increasing dimensionality of vectors. 
 With the rapid expansion of model architectures in recent years, model updates have become higher-dimensional \cite{9451544}, making this curse phenomenon  more pronounced. 
Therefore, both the divergence of model updates and the curse of dimensionality should  be carefully addressed in Byzantine-robust FL defenses.

To tackle the above  issues, this paper proposes a Byzantine-robust FL framework utilizing outsourced auditing, named  FL-OA.
In this framework, the  server collaborates  with a third-party organization that holds an additional root dataset to defend against Byzantine attacks from malicious devices.
Notably, collaboration between two service providers is reasonable in practical  FL scenarios.
Moreover, to mitigate the divergence among benign updates from different devices, we introduce  both a gradient ascent step and a correction term into the local optimization process. 
In addition, to address the curse of dimensionality in  auditing,  we design a parameter importance indicator to extract critical parameters of each model update  for auditing analysis.
The main contributions of this paper are summarized as follows:
\begin{itemize}
\item 
We propose FL-OA, a Byzantine-robust federated learning framework. It eliminates the server's reliance on assumptions regarding the root dataset and the proportion of malicious devices through outsourced auditing.
 
 \item  We theoretically analyze FL-OA to provide theoretical support for its effectiveness.
 
\item We evaluate the performance of FL-OA against various Byzantine attacks  under different data distribution settings across multiple datasets.
Experimental results demonstrate that FL-OA outperforms existing schemes.
\end{itemize}

\section{RELATED WORK} 
In this section, we introduce the Byzantine-robust FL works, which  are  divided into statistical knowledge-based   and the additional root  dataset-based approaches.
\subsubsection{Statistical Knowledge-based Approaches} 
Statistical knowledge-based approaches  typically assume that malicious  updates submitted by Byzantine devices  are geometrically   far away from those of benign devices \cite{NIPS2017f4b9ec30,huang2023multi,krauss2023mesas,11202428,yin2018byzantine,10648998,11421423,10495004}.
Specifically,  Krum   selects a single model update from all submitted updates as the trusted one \cite{NIPS2017f4b9ec30}.
It calculates the sum of  Euclidean distances between each model update and the others   to determine a score, and then  selects the  update with the lowest score as  the global  update.
Similarly,  the work in \cite{yin2018byzantine} propose Median, which removes the updates with the largest and smallest Euclidean distances from the mean point, and then takes the median of the remaining updates as the global  update. 
However, their effectiveness is built on the assumption that the majority of devices are benign, which limits their applicability.

\subsubsection{Additional Root  Dataset-based Approaches} 
Additional Root dataset-based approaches typically assume  the   server holds  a   root dataset to identify  malicious  updates \cite{FLtrust,miao2022privacy,10475552,9798217,10713463,xia2024byzantine}, thereby avoiding the assumption that the majority of devices are benign.
Specifically, FLTrust is the first to propose maintaining an additional root dataset on the  server and generate a trusted root model update \cite{FLtrust}. 
More specifically, the server calculates a trust score by comparing the device update with the root model update, and then uses this score to determine whether the update is benign or malicious.
Similarly, the work in \cite{miao2022privacy} uses the root  update  as a reference for scoring submitted model updates, and then determines the aggregation weight of each device according to the resulting scores.
However, these works  \cite{FLtrust}\cite{miao2022privacy}  rely on the    assumption that the  server holds a clean root dataset, hindering their real-world applicability.
To this end,   the work in  \cite{10475552} propose FL-Auditor, which  uses a third-party auditor  to review    model updates from devices,  thereby assisting the  server obtain a more robust global model.
The core idea of FL-Auditor is that the   server cooperates with a third-party that holds  the  root dataset  to audit,  avoiding the  requirement that the server itself owns the  root dataset.  
Although this design has inspired our work, in practical scenarios, the data distribution of the root dataset differs  from that of the device's local dataset.
This causes the root update deviate from  benign updates, thereby increasing the difficulty of detecting malicious updates.
In addition,  these works \cite{FLtrust,miao2022privacy,10475552,9798217,10713463,xia2024byzantine} struggle to the  divergence among model updates caused by Non-IID data, thereby limiting their defense effectiveness.
To this end,  existing survey \cite{uddin2025systematic} reviews local objective regularization as an effective approach to mitigating  the  divergence of model updates.
For example, the work in \cite{li2021model} incorporates a model-contrastive term into the FL objective to align the local model with the global model.
Meanwhile, the work in \cite{li2020federated} introduces  a proximity term (prox term) during local training to adjust the direction of the local model closer to the global model.
Similarly, the work in \cite{sun2023fedspeed} propose  a correction term to constrain the direction of model updates at  each device.
Although these  works  \cite{sun2023fedspeed}\cite{li2021model}\cite{li2020federated} cannot defend against Byzantine attacks, they inspire us to mitigate the divergence of benign updates, thereby contributing to robust aggregation.


\sloppy
\begin{wrapfigure}{t}{0.19\textwidth}
     \vspace{-11pt}
  \centering
 \includegraphics[width=0.4\columnwidth]{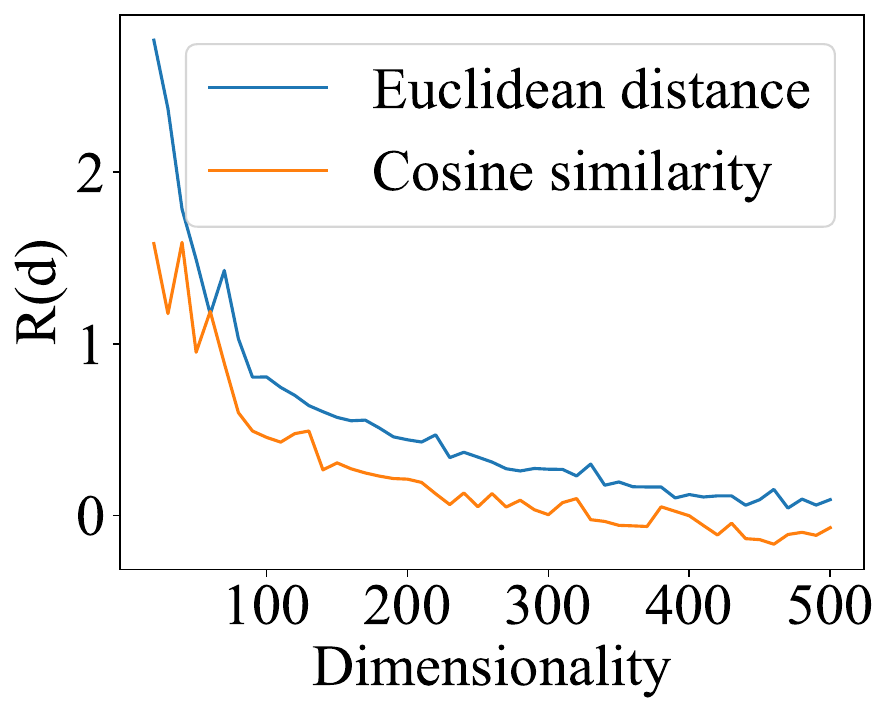}
     \vspace{-25pt}
  \caption{The relative maximum difference with respect to dimensionality.}\label{12181112341}
   \vspace{-10pt}
\end{wrapfigure} 
Notably,  existing works \cite{NIPS2017f4b9ec30,huang2023multi,krauss2023mesas,11202428,yin2018byzantine,10648998,11421423,10495004,FLtrust,miao2022privacy,10475552,9798217,10713463,xia2024byzantine} confront the curse of dimensionality with increasing model scales, regardless of whether they rely on statistical knowledge or an additional   root dataset.
Specifically, when model updates have high dimensionality, the Euclidean distance or cosine similarity computed by these works yields the curse of dimensionality.
To empirically validate this issue, we conduct the following experiment. 
The relative maximum difference  is defined as follows:
\begin{equation}
R(d) = \log\frac{D_{\max}(d) - D_{\min}(d)}{D_{\min}(d)},
\end{equation}
where $D_{\max}(d)$ and $D_{\min}(d)$ denote the maximum and minimum pairwise distances, respectively, in a $d$-dimensional space.
From Fig. \ref{12181112341}, it is observed that   $R(d)$ gradually decreases  with increasing dimensionality,   reflecting the curse of dimensionality in high-dimensional spaces. 
Thus, mitigating the curse is crucial for auditing model updates.
\fussy

\section{PROBLEM FORMULATION}
In this section, we  provide the system architecture, potential threats, and design goals for FL-OA.

\subsection{System Architecture}

\begin{figure}[t]
  \centering   
 \includegraphics[width=0.9\linewidth]{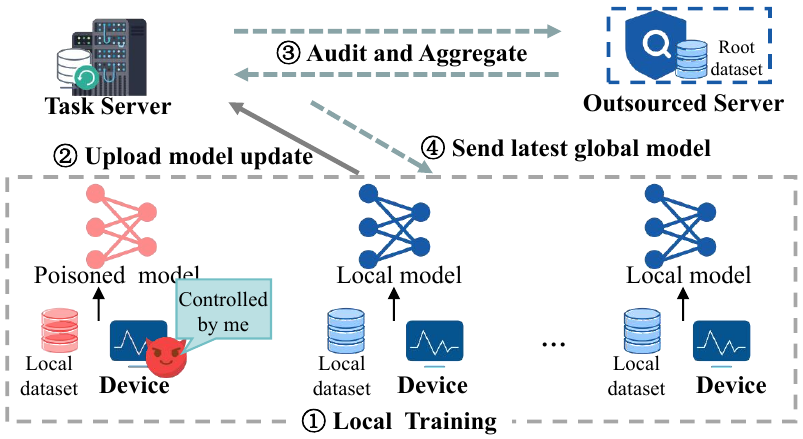}
 \vspace{-5pt}
\caption{System architecture of FL-OA.}
\label{4122201}
\vspace{-15pt}
\end{figure}

As shown in Fig. \ref{4122201}, FL-OA consists of a Task Server (TS), an Outsourced Server (OS), and intelligent devices.
The effectiveness of  FL-OA is predicated on the assumption that the task server cooperates with an outsourced server that maintains the  additional  root datasets required for training tasks.
The concrete roles in FL-OA are elaborated as follows.
\begin{itemize}
  \item The Task Server (TS) is the central server (service provider) in the FL system, responsible for publishing  training tasks to the entire FL system.
  \item The Outsourced Server (OS) is a third-party server (service provider) that maintains the additional root dataset  required by TS's task.  
      It is responsible for auditing the model updates from devices, thereby providing guidance for the TS to perform robust aggregation.
  \item Intelligent devices   are responsible for performing local training based on their own private datasets.
\end{itemize}

In  traditional FL process, each device  $k\in\mathcal{K}$ cooperatively assist  the TS for training   a global model using its own   private dataset $\mathcal{D}_k$, where  \scalebox{0.85}{$\mathcal{D} = \bigcup_{k\in\mathcal{K}}\mathcal{D}_k$} is the union of the training samples held by all devices.
Formally, FL aims to seek the optimal  global model parameters $\textit{\textbf{W}}^\star \in \mathbb{R}^d$ by  minimizing  the global objective function $\mathcal{L}(\cdot;\cdot)$,  denoted as:
\begin{equation}\scalebox{1}{$
\textit{\textbf{W}}^\star = \mathop{\arg\min}_{\textit{\textbf{W}}}\mathcal{L}(\textit{\textbf{W}};\mathcal{D}) = \frac{1}{|\mathcal{K}|}\sum_{k \in\mathcal{K}}\mathcal{L}_k(\textit{\textbf{W}};\mathcal{D}_k), \label{4131653}$}
\end{equation}
where 
\begin{equation}\scalebox{1}{$
\mathcal{L}_k(\textit{\textbf{W}};\mathcal{D}_k) =\frac{1}{|\mathcal{D}_k|}l(\textit{\textbf{W}};\mathcal{D}_k) , \label{4131653}$}
\end{equation}
where $l(\cdot;\cdot) $ denotes the empirical loss function, $\mathcal{L}_k(\cdot;\cdot)$ denotes the local objective function of device $k$.
Each device trains a local model  by minimizing the empirical loss over its    private dataset $\mathcal{D}_k$.
After completing local training, the device uploads its  model update to the TS.
At the $t$-th  round, let  $\textit{\textbf{u}}_k^t$ be  the model update of device $k$, which is calculated as the difference between the local model parameters and the current global model parameters.
The TS then aggregates the model updates from  devices to compute the global model.

\subsection{Potential Threats}
In practice, since the training process is initiated by the TS, it is assumed to execute its task honestly.
In addition, it is reasonable in practice for OS to honestly execute the audit rule  due to legal regulations and company reputation.
However,  the TS cannot guarantee that all intelligent devices  follow the protocol.
Thus, we assume that the intelligent devices   consist  of both benign and malicious devices.
Specifically, each benign device  honestly performs local training and uploads its model update.
Instead,  malicious devices disturb the global model  by tampering their  training data or  model update, thereby degrading the performance of the global model.

\subsection{Design Goals}
Considering the above  threats, FL-OA is designed to achieve the following goals in defending against Byzantine attacks:
\begin{itemize}
  \item \textit{Consistency}. 
FL-OA should mitigate the divergence among benign  updates during local training under various data distributions.
  \item \textit{Fidelity}. 
FL-OA should not sacrifice the accuracy of the global model in the absence of  Byzantine attacks.
Specifically, without Byzantine attacks, FL-OA should learn a global model whose accuracy is close to that learned by  averaging the model updates from all devices.
  \item \textit{Robustness}. 
FL-OA should  learn a global model with higher accuracy than existing defense schemes  under various Non-IID data  when subjected to Byzantine attacks.
\end{itemize}

\begin{algorithm}[t]
\begin{small}
    \caption{High-Level Idea of  FL-OA}
    \label{algorithm:test123}
    \LinesNumbered
\KwIn{\parbox[t]{0.82\linewidth}{Device set $\mathcal{K}$,     initial global model parameters $\textit{\textbf{W}}^{\textit{init}}$, the local iterations  $E$, the number of rounds $T$, local learning rate  $\eta$, weighting coefficients $\beta$ and $\alpha$, the proportion of selected coordinates $\kappa$, and global learning rate  $\mu$.
}}
    \KwOut {Final global model parameters $\textit{\textbf{W}}^{\textit{final}}$.}
Initialize $T,E,\textit{\textbf{W}}^{\textit{init}}$;\\
\For{{\rm each  round} $t \in \{0,\cdots,T-1\}$}{
\hspace*{-1em}\textit{Step I}: $\mathtt{Select\ and\ synchronize}$.\\
TS  selects the set of devices  $\mathcal{K}^t \subset \mathcal{K}$;\\
TS sends $\textit{\textbf{W}}^{t}$ to devices in  $\mathcal{K}^t$ and OS;\\
\hspace*{-1em}\textit{Step II}: $\mathtt{Local  \ Training}$.\\
// $\mathtt{Device}$:\\
\For{{\rm each selected device} $k \in \mathcal{K}^t$}{
$\textit{\textbf{u}}^t_k \leftarrow  \operatorname {Local\_Training}(\textit{\textbf{W}}^{t},\mathcal{D}_k,\eta, E,\beta,\alpha)$;\\
{\rm Send  $\textit{\textbf{u}}^t_k$ to TS};\\
}
// $\mathtt{OS}$:
$\textit{\textbf{u}}^t_{\textit{root}} \leftarrow \operatorname {Local\_Training}(\textit{\textbf{W}}^{t},\mathcal{D}_{\textit{root}},\eta, E,\beta,\alpha)$;\\
\hspace*{-1em}\textit{Step III}:  $\mathtt{Audit \ Updates}$.\\
// $\mathtt{TS}$:
{\rm Send  $\textit{\textbf{u}}^t_k$ to OS};\\
// $\mathtt{OS}$:
$ s^t_k, \widetilde{\textit{\textbf{u}}}^t_k\leftarrow \operatorname {Update\_Audit}(\textit{\textbf{u}}^t_k, \kappa, \textit{\textbf{u}}^t_\textit{root})$;\\
{\rm Send  $s^t_k$ and $\widetilde{\textit{\textbf{u}}}^t_k$ to TS};\\
\hspace*{-1em}\textit{Step IV}:  $\mathtt{Robust \ Aggregation}$.\\
// $\mathtt{TS}$:
$\textit{\textbf{W}}^{t+1} \leftarrow \textit{\textbf{W}}^{t}+\mu \frac{1}{\sum_{k \in \mathcal{K}^t}s^t_k}\sum_{k \in \mathcal{K}^t}s^t_k\widetilde{\textit{\textbf{u}}}^t_k$;\\
}
$\textit{\textbf{W}}^{\textit{final}} \leftarrow \textit{\textbf{W}}^{T}$;\\
\Return Final global model parameters $\textit{\textbf{W}}^{\textit{final}}$.
\end{small}
\end{algorithm}

\section{DESIGN OF OUR FRAMEWORK} 
In this section, we  introduce the high-level idea of FL-OA, and then describe the concrete construction. 
\subsection{High-Level Idea of FL-OA}
The FL-OA is formally presented in Algorithm \ref{algorithm:test123}, where the TS, OS, and intelligent devices jointly execute  the federated   protocol.
Specifically, FL-OA iteratively performs the following  steps:
\begin{itemize}
  \item \textit{Step I. Select  and  Synchronize}. During  the $t$-th   round, TS randomly selects  a subset of devices, denoted by   $\mathcal{K}^t$ for local  training.
       The main reason is that selecting partial devices  reduces the communication overhead of TS.
It then synchronizes  the latest global model parameters $\textit{\textbf{W}}^t$ to  the selected devices  and OS.
  \item \textit{Step II. Local  Training}. 
  Each selected device   $k \in \mathcal{K}^t$  locally optimizes the received global model using its  private dataset $\mathcal{D}_k$, and then sends its model update to  TS.
  Meanwhile, OS trains its root model on the clean root dataset $\mathcal{D}_\textit{root}$.
   \item \textit{Step III. Audit Updates}.
Upon receiving model updates, TS forwards them to OS for auditing. 
OS then evaluates these updates using its root model update as a reference and returns the audit results to TS.
  \item \textit{Step IV. Robust Aggregation}. 
TS aggregates the model  updates  using  the audit results provided by OS to calculate  the updated global model parameters $\textit{\textbf{W}}^{t+1}$.
\end{itemize}
This  process continue until the predefined number of rounds is reached,   yielding  the final global model parameters $\textit{\textbf{W}}^{\textit{final}}$.
In the following, we describe the \textit{Steps II},  \textit{III}, and \textit{IV} in detail.

\subsection{Local  Training}
In this step, the selected devices and the OS separately execute Algorithm \ref{algorithm:test3445}.
The devices use it to obtain improved model updates, while the OS uses it to generate the root model update for auditing the devices' updates in \textit{Step III}.
For clarity, we describe the local training process only from the perspective of a device, as the OS follows the same procedure.

\begin{algorithm}[t]
\begin{small}
    \caption{$\operatorname{Local\_Training}$}
    \label{algorithm:test3445}
    \LinesNumbered
\KwIn{\parbox[t]{0.82\linewidth}{
Local dataset $\mathcal{D}_k$, local learning rate $\eta$, local iteration $E$, 
global model parameters $\textit{\textbf{W}}^t$, and weighting coefficients 
$\beta$ and $\alpha$.
}}
    \KwOut {Model update  $\textit{\textbf{u}}^t_{k}$.}
	$\textit{\textbf{w}}^t_{k,(0)} \leftarrow \textit{\textbf{W}}^t$;\\
	  \For{{\rm each local iteration} $e$ $\in$ $\{0,1,\cdots,E-1\}$}{
	    	Randomly sample  $\mathcal{D}^t_{k,(e)}\subset \mathcal{D}_k$;\\
		$\textit{\textbf{g}}_{k,(e+1), 1}^{t} \leftarrow \nabla\mathcal{L}_k(\textit{\textbf{w}}^t_{k,(e)}; \mathcal{D}^t_{k,(e)})$;\\
		$\widetilde{\textit{\textbf{w}}}^t_{k,(e+1)} \leftarrow \textit{\textbf{w}}^t_{k,(e)} + \eta \textit{\textbf{g}}^t_{k,(e+1), 1}$;\\
		$\textit{\textbf{g}}_{k,(e+1), 2}^{t} \leftarrow \nabla\mathcal{L}_k(\widetilde{\textit{\textbf{w}}}^t_{k,(e+1)}; \mathcal{D}^t_{k,(e)})$;\\
   		$\widetilde{\textit{\textbf{g}}}_{k,(e+1)}^{t} \leftarrow (1-\beta)\textit{\textbf{g}}_{k,(e+1), 1}^{t}+ \beta \textit{\textbf{g}}_{k,(e+1), 2}^{t}$;\\
   		$\textit{\textbf{w}}_{k,(e+1)}^{t} \leftarrow \textit{\textbf{w}}_{k,(e)}^{t}-\eta(\widetilde{\textit{\textbf{g}}}_{k,(e+1)}^{t}- \widehat{\textit{\textbf{g}}}_{k}^{t-1}+ \frac{1}{\alpha}(\textit{\textbf{w}}_{k,(e)}^{t} - \textit{\textbf{W}}^t))$;\\
    }
$\widehat{\textit{\textbf{g}}}_{k}^{t} \leftarrow \widehat{\textit{\textbf{g}}}_{k}^{t-1} - \frac{1}{\alpha}(\textit{\textbf{w}}_{k,(E)}^{t} - \textit{\textbf{W}}^t)$;\\
$\textit{\textbf{u}}^t_{k} \leftarrow \textit{\textbf{w}}^t_{k,(E)} - \alpha\widehat{\textit{\textbf{g}}}_{k}^{t}-\textit{\textbf{w}}^t_{k,(0)}$;\\
\end{small}
\end{algorithm}
Specifically, data heterogeneity causes the divergence among model updates from  devices, making it difficult for existing defense schemes \cite{NIPS2017f4b9ec30,huang2023multi,krauss2023mesas,11202428,yin2018byzantine,10648998,11421423,10495004,FLtrust,miao2022privacy,10475552,9798217,10713463,xia2024byzantine}  to distinguish malicious updates from benign ones.
To address this issue, we design a correction term  into  local  training for enhancing the consistency of model updates.
More specifically,    after receiving the global model parameters $\textit{\textbf{W}}^t$, each selected device   iteratively updates its local model as follows:
\begin{itemize}
 \item \textit{Stage I}.
 At the $e$-th local iteration, each selected device  randomly samples mini-batch \scalebox{0.85}{$\mathcal{D}_{k,(e)}^t$} from its   dataset $\mathcal{D}_k$ and computes the  stochastic gradient  \scalebox{0.85}{$\textit{\textbf{g}}_{k,(e+1), 1}^{t}$} by  \scalebox{0.85}{$\nabla\mathcal{L}_k(\textit{\textbf{w}}^t_{k,(e)}; \mathcal{D}^t_{k,(e)})$}.
Using this gradient, the device  performs a gradient ascent  step with learning rate $\eta$  to obtain the intermediate parameters:  \scalebox{0.85}{$\widetilde{\textit{\textbf{w}}}^t_{k,(e+1)} = \textit{\textbf{w}}^t_{k,(e)} + \eta \textit{\textbf{g}}^t_{k,(e+1), 1}.$}
Notably, existing works \cite{andriushchenko2022towards}\cite{qu2022generalized}  have observed that applying vanilla gradient descent to local objective function  causes the global model to fall into a sharp valley, which  increases the inconsistency of  updates among devices.
 Thus, according to the analysis of works \cite{andriushchenko2022towards}\cite{qu2022generalized}, introducing a gradient ascent step during the local  training  can guide the optimization process toward flat minima.

 \item \textit{Stage II}.
The selected device reuses the same mini-batch  \scalebox{0.85}{$\mathcal{D}_{k,(e)}^t$} to compute a second  gradient \scalebox{0.85}{$\textit{\textbf{g}}_{k,(e+1), 2}^{t}$} by \scalebox{0.85}{$\nabla\mathcal{L}_k(\widetilde{\textit{\textbf{w}}}^t_{k,(e+1)}; \mathcal{D}^t_{k,(e)}).$} 
By combining the two gradients, we  calculate  the intermediate gradient \scalebox{0.85}{$\widetilde{\textit{\textbf{g}}}_{k,(e+1)}^{t}$} by
\scalebox{0.85}{$ (1-\beta)\textit{\textbf{g}}_{k,(e+1), 1}^{t}+ \beta \textit{\textbf{g}}_{k,(e+1), 2}^{t}.$}
Notably, the work in \cite{zhao2022penalizing} points out that the intermediate gradient \scalebox{0.85}{$\widetilde{\textit{\textbf{g}}}_{k,(e+1)}^{t}$} leverages the first-order approximation  to efficiently implement the corresponding gradient  to fit well in the gradient descent process. 
Moreover, the work in  \cite{zhao2022penalizing} further  applies  gradient normalization  to regularize the gradients $\textit{\textbf{g}}_{k,(e+1),1}^{t}$and $\textit{\textbf{g}}_{k,(e+1),2}^{t}$.
However, since the magnitude of model updates across devices carries important information that reflects the characteristics of their local data, normalizing gradients during local  training would lead to the loss of such information.
Thus, we directly combine the two gradients $\textit{\textbf{g}}_{k,(e+1),1}^{t}$and $\textit{\textbf{g}}_{k,(e+1),2}^{t}$ without applying normalization.
 \item \textit{Stage III}.
The selected device updates its model parameters by the following formula:
\begin{equation}\scalebox{0.85}{$\textit{\textbf{w}}_{k,(e+1)}^{t} = \textit{\textbf{w}}_{k,(e)}^{t}-\eta(\widetilde{\textit{\textbf{g}}}_{k,(e+1)}^{t}- \widehat{\textit{\textbf{g}}}_{k}^{t-1}+ \frac{1}{\alpha}(\textit{\textbf{w}}_{k,(e)}^{t} - \textit{\textbf{W}}^t)),\nonumber$} \end{equation}
where the deviation  \scalebox{0.85}{$(\textit{\textbf{w}}_{k,(e)}^{t} - \textit{\textbf{W}}^t)$} can be regarded  as  a proximal term  to align local model with the global model.
This term   facilitates a balance between local model parameters of each device and global model parameters.
However, as noted in work \cite{hanzely2020federated}, because local objectives differ across devices, their resulting local solutions inevitably exhibit distinct local offsets.
 Specifically, the local offset refers to the deviation
 \scalebox{0.85}{$(\textit{\textbf{w}}_{k,(E)}^{t} - \textit{\textbf{W}}^t)$} accumulated in  each  round $t$.
This indicates that the inconsistency of model updates has not been eliminated.
To this end, we introduce  \scalebox{0.85}{$\widehat{\textit{\textbf{g}}}_{k}^{t-1}$}  as the correction term to  counteract the  local  offset.
   This correction term could be considered as a momentum that accumulates previous local offset (see the Theorem \ref{16197}). 
   At the beginning of federated training, each device's \scalebox{0.85}{$\widehat{\textit{\textbf{g}}}_{i}^{-1}$} is initialized to 0.
    It is  updated in line 9 of Algorithm \ref{algorithm:test3445} to capture the historical local offset patterns between both local and global models.
    
\end{itemize}

 After repeating  the above stage  $E$ times, each device updates its correction term as 
\scalebox{0.85}{$\widehat{\textit{\textbf{g}}}_{k}^{t} := \widehat{\textit{\textbf{g}}}_{k}^{t-1} - \frac{1}{\alpha}(\textit{\textbf{w}}_{k,(E)}^{t} - \textit{\textbf{W}}^t),$}
where the term \scalebox{0.85}{$(\textit{\textbf{w}}_{k,(E)}^{t} - \textit{\textbf{W}}^t)$} can be considered as local offset. 
To prevent the correction term itself from introducing additional bias, we subtract the updated correction term \scalebox{0.85}{$\widehat{\textit{\textbf{g}}}_{k}^{t}$} from the local model parameters \scalebox{0.85}{$\textit{\textbf{w}}_{k,(E)}^{t}$}, thereby obtaining bias-corrected local model parameters, i.e.,
\scalebox{0.85}{$\textit{\textbf{w}}^t_{k,(E)} - \alpha\widehat{\textit{\textbf{g}}}_{k}^{t}.$}
Finally, the model update of device $k$ is computed as \scalebox{0.85}{$\textit{\textbf{u}}_{k}^{t}
:=
\textit{\textbf{w}}_{k,(E)}^{t}
-
\alpha \widehat{\textit{\textbf{g}}}_{k}^{t}
-
\textit{\textbf{w}}_{k,(0)}^{t}.$}

Upon executing Algorithm \ref{algorithm:test3445}, the selected devices submit their model updates to TS.
In parallel, OS also executes Algorithm \ref{algorithm:test3445} on its clean root dataset $\mathcal{D}_\textit{root}$ to generate the root model update $\textit{\textbf{u}}^t_{\textit{root}}$, which is retained locally rather than uploaded to TS.
The local  training step enhances the consistency of model updates under Non-IID data,  facilitating the identification of deviation caused by malicious updates.
The system then proceeds to \textit{Step III} (i.e., \textit{Audit Updates}).

\subsection{Audit Updates}
TS receives the set of model updates  $\{\textit{\textbf{u}}^t_k\}_{k \in \mathcal{K}^t}$ and forwards them to  OS for auditing.
The audit operation at OS consists of \textit{Critical Parameter Extraction}, \textit{Trust  Score  Calculation}, and \textit{Gradient   Normalization}, as outlined  in Algorithm \ref{algorithm:test5123}. 
Each device's model update is a high-dimensional vector characterized by both direction and magnitude.
 Malicious devices may manipulate the direction  of their model updates, thereby compromising the global model update.
Without using fundamental trust, it is difficult  to determine  which model updates can represent the global update direction.
Therefore, OS employs cosine similarity to measure the directional  difference  between its root  update and the device model update.
However, computing the cosine similarity between two high-dimensional vectors suffers from the curse of dimensionality.
 To address this issue, OS extracts key parameters from each model update for audit analysis, thereby alleviating the curse of dimensionality.
  In addition, since malicious devices may manipulate the magnitude of their model updates to amplify their impact \cite{shejwalkar2021manipulating}, OS normalizes each model update before auditing. The detailed procedure is described below.

 \textit{Stage I. Critical Parameter Extraction}.
To eliminate the curse of dimensionality caused by computing the distance between two  high-dimensional vectors, many dimensionality reduction methods are proposed, such as PCA \cite{mackiewicz1993principal} and t-SNE \cite{van2008visualizing}.
However, they  project high-dimensional parameters into a lower-dimensional space, which may obscure anomalous information in certain parameters caused by malicious updates.
Thus, we focus on extracting critical parameters  from model updates for auditing analysis.
Based on this consideration, we introduce the Parameter Importance Indicator ($\operatorname{PII}$) to measure the importance of each parameter within the model updates and select the critical parameters based on their $\operatorname{PII}$ values.

Specifically, $\operatorname{PIF}$ combines   the absolute  magnitude of each parameter and its deviation from other devices  at the same coordinate,   highlighting parameters with large magnitudes that deviate significantly from the majority of devices.
Formally, for the $j$-th parameter of device $k$ at round $t$, its $\operatorname{PII}_{k}^{t}[j]$  is defined as:
\begin{equation}\label{381514}
\operatorname{PII}_{k}^{t}[j]
=
|\textit{\textbf{u}}_{k}^{t}[j]|
+
\frac{
\left|\,|\textit{\textbf{u}}_{k}^{t}[j]|
-
\operatorname{med}\!\left(\left\{|\textit{\textbf{u}}_i^t[j]|\right\}_{i\in\mathcal K}\right)
\right|
}{
\operatorname{med}\!\left(\left\{|\textit{\textbf{u}}_i^t[j]|\right\}_{i\in\mathcal K}\right)+\epsilon
},
\end{equation} 
where $|\textit{\textbf{u}}_{k}^{t}[j]|$ denotes the absolute value of the $j$-th parameter in the model update of device $k$ at the $t$-th round, $\operatorname{med}\!\left(\left\{|\textit{\textbf{u}}_i^t[j]|\right\}_{i\in\mathcal K}\right)$ is the median absolute  magnitude at the  $j$-th parameter across all devices, and $\epsilon$ is a small constant to prevent division by zero.
Based on formula (\ref{381514}), each model update $\textit{\textbf{u}}_{k}^{t}$ yields a corresponding $\operatorname{PII}_{k}^{t}$ vector.
A parameter with a large $\operatorname{PIF}$ value has a substantial influence on the model update and deviates significantly from the parameters of most devices at the same coordinate, indicating that it is more likely to have been maliciously manipulated.
Thus, the critical parameter coordinates in each model update are selected according to their $\operatorname{PII}_{k}^{t}$ vectors.
Formally, we define the function  $\operatorname{top}(\operatorname{PII}_{k}^{t},\kappa)$ to map the $\operatorname{PII}_{k}^{t}$ vector to a binary mask vector   $M_{k}^{t} \in  \{0,1\}^d$, where the coordinates corresponding to the largest $\kappa$ proportion of values in $\operatorname{PII}_{k}^{t}$ are set to 1,  and the others are set to 0.
Hence, the coordinates with mask value 1 are marked  as the critical parameter coordinates in device $k$'s model update at round $t$.
Accordingly, $(\textit{\textbf{u}}^t_k \odot M_{k}^{t})$ selects the parameters used for auditing analysis from the update $\textit{\textbf{u}}^t_k$ according to the mask $M_{k}^{t}$, thereby achieving dimensionality reduction,  where $\odot$ denotes the Hadamard product.

\begin{algorithm}[t]
\begin{small}
    \caption{$\operatorname{Update\_Audit}$}
    \label{algorithm:test5123}
    \LinesNumbered
    \KwIn {Model update $\textit{\textbf{u}}^t_k$,   root model update $\textit{\textbf{u}}^t_\textit{root}$, and the   proportion of selected coordinates $\kappa$, 
}
    \KwOut {Trust score $s^t_k$,   normalized model update $\widetilde{\textit{\textbf{u}}}^t_k$.}
// $\mathtt{Dimensionality \ Reduction}$.\\
\For{{\rm each selected device} $k\in\mathcal{K}^t$}{$\operatorname{PII}_{k}^{t} \leftarrow \operatorname{Param\_ Imp\_Ind}(\{\textit{\textbf{u}}_k^t\}_{k\in\mathcal{K}}, \kappa)\hfill\lhd$  by Eq. (\ref{381514});\\
$M_{k}^{t} \leftarrow\operatorname{Top}(\operatorname{PII}_{k}^{t},\kappa)$;\\}{}
// $ \mathtt{Trust \ Score \ Calculation}$.\\
	  \For{{\rm each selected device} $k\in\mathcal{K}^t$ }{
$s^t_k \leftarrow  \operatorname{Relu}\left(\cos\big((\textit{\textbf{u}}^t_k \odot M_{k}^{t}),\,\textit{\textbf{u}}^t_\textit{root}\big)\right)\hfill\lhd$  by Eq. (\ref{3172004});\\
}
// $\mathtt{Gradient \ Normalization}$.

	  \For{{\rm each selected device} $k\in\mathcal{K}^t$ }{
$\widetilde{\textit{\textbf{u}}}^t_k \leftarrow \frac{\Vert\textit{\textbf{u}}_{\textit{root}}^t\Vert}{\Vert\textit{\textbf{u}}_{k}^t\Vert}\textit{\textbf{u}}^t_k$;\\
}
\Return  Trust  score $s^t_k$, normalized model update $\widetilde{\textit{\textbf{u}}}^t_k$.
\end{small}
\end{algorithm}

 \textit{Stage II. Trust  Score Calculation}.
To ensure that benign  updates contribute more to the global model, we compute a trust score for each device based on the  update derived from the root dataset. 
The trust score  is derived using a similarity-based evaluation method, whose core idea has been adopted in related FL works \cite{FLtrust}\cite{miao2022privacy}\cite{10475552}.
However, these methods suffer from the challenges posed by data heterogeneity and the curse of dimensionality.
To this end,   FL-OA introduces a correction term during local training and focuses on critical parameters during the auditing step, thereby  mitigating these challenges.
Specifically, after obtaining the mask vector   $M_{k}^{t}$, OS computes a trust score  for each selected device.
For device $k$, the trust score $s_k^t$ is defined as the cosine similarity between the root model update $\textit{\textbf{u}}^t_{\textit{root}}$ and its model update $\textit{\textbf{u}}^t_k$ over the critical parameters.
Formally,   $s_k^t$ is computed as follows: 
\begin{equation}\label{3141036}
s^t_k = \cos\big((\textit{\textbf{u}}^t_k \odot M_{k}^{t}),\,\textit{\textbf{u}}^t_\textit{root}\big),
\end{equation}
where $\cos(\cdot,\cdot)$ denotes the cosine similarity.
However, the cosine similarity may be negative. 
Specifically, malicious updates  may be in the opposite direction to the root model update,  resulting in a negative trust score (i.e., $s^t_k  <0$).
To solve the negative problem,  $\operatorname{Relu}(\cdot)$ function is used to clip $s^t_k$. Formally, the computation of the  score $s^t_k$ is modified as:
\begin{equation}\label{3172004}
s^t_k =  \operatorname{Relu}\left(\cos\big((\textit{\textbf{u}}^t_k \odot M_{k}^{t}),\,\textit{\textbf{u}}^t_\textit{root}\big)\right),
\end{equation}
where $\operatorname{Relu}(x) = x $ if $x>0$ and $\operatorname{Relu}(x) = 0 $ otherwise.

 \textit{Stage III. Gradient   Normalization}.
Some malicious devices may amplify their influence by submitting model updates with excessively large magnitudes. 
 To counter this behavior, we adopt a unified normalization operation that standardizes the magnitude of each  update  to match  the norm of the root update $ \Vert\textit{\textbf{u}}_{\textit{root}}^t\Vert$. 
 This design ensures that all updates have the same magnitude effect during aggregation, thereby preventing malicious devices from dominating the optimization of the global model by manipulating update magnitudes.
Similar normalization strategies have been adopted in existing studies \cite{FLtrust}\cite{miao2022privacy}\cite{10475552}.
Formally,  the  model update is  normalized as:
\scalebox{0.85}{$
\widetilde{\textit{\textbf{u}}}^t_k := \frac{\Vert\textit{\textbf{u}}_{\textit{root}}^t\Vert}{\Vert\textit{\textbf{u}}_{k}^t\Vert}\textit{\textbf{u}}^t_k,$}
where  $\widetilde{\textit{\textbf{u}}}^t_k$ is the normalized model update.
After  \textit{Step III}, OS obtains the trust score $s_k^t$ and the normalized model update $\widetilde{\textit{\textbf{u}}}^t_k$ for each selected device, and  sends them to TS to guide  the aggregation process.

\subsection{Robust Aggregation}
In the step,  after receiving  trust scores $s_k^t$ and normalized model updates  $\widetilde{\textit{\textbf{u}}}^t_k$, TS aggregates the normalized model updates  to obtain the global model.
Formally, the global model is updated as:
\begin{equation}\scalebox{0.85}{$
\textit{\textbf{W}}^{t+1} = \textit{\textbf{W}}^{t}+\mu \textit{\textbf{u}}_g^t,$}
\end{equation}
where $\mu$   is the global learning rate, $\textit{\textbf{u}}_g^t$ represents the global update, computed as: 
\scalebox{0.85}{$
\textit{\textbf{u}}_g^t =
\frac{1}{\sum_{k \in \mathcal{K}^t}s_k^t}
\sum_{k \in \mathcal{K}^t}
s_k^t \widetilde{\textit{\textbf{u}}}_k^t.
$}
The above process is repeated in each round until the predefined number of training rounds is reached.

\section{ANALYSES}

This section presents the convergence guarantees of FL-OA and analyzes the correction term in depth.
Specifically, we prove that the deviation between the global model parameters $\textit{\textbf{W}}^{t}$ learned by FL-OA (under Byzantine attacks) and the optimal global model parameters $\textit{\textbf{W}}^{*}$ is bounded.
In addition, we derive the relationship between the local offset and the correction term.
In the following, we  give  three assumptions, which are commonly adopted in convergence  analyses \cite{FLtrust}\cite{cho2020client}\cite{wang2020tackling}\cite{xu2025detecting}, followed by our theorems.

\begin{assumption}
\textit{The function} $\mathcal{L}(\textit{\textbf{W}})$ \textit{is} $\gamma$-\textit{strongly convex ($\gamma \textgreater 0$) and differentiable  over the parameter space}  $\Phi$ \textit{with} $L_1$-\textit{Lipschitz continuous gradient} ($L_1\textgreater 0$). \textit{Meanwhile, the empirical loss function} $l(\textit{\textbf{W}})$ \textit{is} $L_2$-\textit{Lipschitz probabilistically. Formally,  for any} $\textit{\textbf{W}}^{t_1},\textit{\textbf{W}}^{t_2} \in \Phi$, \textit{we have the following:}
\begin{equation}\scalebox{0.82}{$
\mathcal{L}(\textit{\textbf{W}}^{t_1}) \geq \mathcal{L}(\textit{\textbf{W}}^{t_2})+\nabla \mathcal{L}(\textit{\textbf{W}}^{t_2})\odot (\textit{\textbf{W}}^{t_1}-\textit{\textbf{W}}^{t_2})+ \frac{\gamma}{2}\Vert\textit{\textbf{W}}^{t_1}-\textit{\textbf{W}}^{t_2}\Vert^2,$}
\end{equation}
\begin{equation}\scalebox{0.9}{$
\nabla \mathcal{L}(\textit{\textbf{W}}^{t_2})-\nabla \mathcal{L}(\textit{\textbf{W}}^{t_1}) \leq L_1\Vert\textit{\textbf{W}}^{t_2}-\textit{\textbf{W}}^{t_1}\Vert,$}
\end{equation}
\textit{where} $\nabla$ \textit{is gradient,} $\textit{\textbf{x}} \odot \textit{\textbf{y}}$ \textit{denotes the inner product of vectors} $\textit{\textbf{x}}$ \textit{and} $\textit{\textbf{y}}$, \textit{and} $\Vert\cdot\Vert$ \textit{represents} $\ell_2$ \textit{norm}.
\textit{For any} $\delta \in (0,1)$ \textit{and} $\bar{\delta} = 1- \frac{\delta}{3}$, \textit{there exists an} $L_2$ \textit{such that:}
\begin{equation}\scalebox{1}{$
Pr\left\{\sup_{\textit{\textbf{W}}^{t_1}\neq\textit{\textbf{W}}^{t_2}}\frac{\|\nabla l(\textit{\textbf{W}}^{t_1})-\nabla l(\textit{\textbf{W}}^{t_2})\|}{\|\textit{\textbf{W}}^{t_1}-\textit{\textbf{W}}^{t_2}\|}\leq L_2\right\}\geq\bar{\delta}.$}
\end{equation}
\end{assumption}

\begin{assumption}
The  local dataset $\mathcal{D}_k$ of each device $k$ and the root  dataset $\mathcal{D}_\textit{root}$ owned by OS are independently sampled from the training dataset distribution $\chi$. 
\end{assumption}

\begin{assumption}
 \textit{The gradient of the experical loss function} $l(\textit{\textbf{W}}^\star;\mathcal{D})$ \textit{at the optimal global model parameter} $\textit{\textbf{W}}^\star$ \textit{is bounded. For any} $\textit{\textbf{W}}\in \Phi$, $h(\textit{\textbf{W}};\mathcal{D}) = \nabla l(\textit{\textbf{W}};\mathcal{D}) -\nabla l(\textit{\textbf{W}}^\star;\mathcal{D})$ \textit{is bounded. Moreover, for any unit vector} $\textit{\textbf{v}}$, \textit{there exist positive constants} $\sigma_1$ \textit{and} $\gamma_1$ \textit{such that} $\nabla l(\textit{\textbf{W}}^\star;\mathcal{D}) \odot\textit{\textbf{v}}$ \textit{is sub-exponential.
And  there exist positive constants} $\sigma_2$ \textit{and} $\gamma_2$ \textit{such that for any} $\textit{\textbf{W}}\in \Phi$ with $\textit{\textbf{W}}^\star \neq \textit{\textbf{W}}$ \textit{and  vector} $\textit{\textbf{v}}$, $\frac{\left(\left(h\left(\textit{\textbf{W}};\mathcal{D}\right)-\mathbb{E}[h\left(\textit{\textbf{W}};\mathcal{D}\right)]\right)\odot \textit{\textbf{v}}\right)}{\|\textit{\textbf{W}}-\textit{\textbf{W}}^{\star}\|}$ \textit{is sub-exponential with}  $\sigma_2$ \textit{and} $\gamma_2$.
\textit{Then, let} $G_1$ \textit{to denote} $\exp{({\sigma^2_1\xi^2/2})}$ \textit{and} $G_2$ \textit{to denote} $\exp{({\sigma^2_2\xi^2/2})}$, \textit{for any} $|\xi| \leq \frac{1}{\gamma_1}$, \textit{any} $|\xi| \leq \frac{1}{\gamma_2}$ \textit{and} $\textbf{B} = \{\textit{\textbf{v}}: \Vert\textit{\textbf{v}}\Vert = 1\}$, \textit{we have the following:}
\begin{equation}
\sup_{\textit{\textbf{v}}\in \textbf{B} }\mathbb{E}[\exp\left(\xi\left(\nabla l\left(\textit{\textbf{W}}^\star;\mathcal{D}\right)\odot \textit{\textbf{v}}\right)\right)]\leq G_1,
\end{equation}
\begin{equation}\scalebox{1}{$
\sup_{\textit{\textbf{v}}\in \textbf{B}}\mathbb{E}\left[\exp\left(\frac{\xi\left(\left(h\left(\textit{\textbf{W}};\mathcal{D}\right)-\mathbb{E}[h\left(\textit{\textbf{W}};\mathcal{D}\right)]\right)\odot \textit{\textbf{v}}\right)}{\|\textit{\textbf{W}}-\textit{\textbf{W}}^{\star}\|}\right)\right]\leq G_2.$}
\end{equation}
\end{assumption}

\begin{thm}
Suppose Assumptions 1-3 hold,  the  deviation between the global model parameters $\textit{\textbf{W}}^t$ learned by FL-OA and the optimal global model parameters $\textit{\textbf{W}}^{*}$ is bounded. Formally, in the $t$-th  round, 
the relationship between $\textit{\textbf{W}}^{t}$ and $\textit{\textbf{W}}^{*}$ satisfies the following formula:
\begin{equation}\scalebox{0.85}{$
\begin{aligned}
\Vert\textit{\textbf{W}}^{t}-\textit{\textbf{W}}^{*}\Vert &\leq(1-\tau)^t\Vert\textit{\textbf{W}}^{\textit{init}}-\textit{\textbf{W}}^{*}\Vert+\frac{12\mu\triangle_1}{\tau},
\end{aligned}$}
\end{equation}
where $\tau = 1- (\sqrt{(1+\mu^2L_1^2-\mu \gamma)}+2\mu L_1+24\mu\triangle_2)$,
$\triangle_1 = \sqrt{2}\sigma_1 \sqrt{(d\log6+\log(3/\delta))/|\mathcal{D}_{\textit{root}}|}$, and 
\begin{equation}\scalebox{0.85}{$\triangle_2 = \sigma_2\sqrt{2/|\mathcal{D}_{\textit{root}}|}\sqrt{d\log\frac{18L_3}{\delta_2}+\frac{d}{2}\log\frac{|\mathcal{D}_{\textit{root}}|}{d}+\log(\frac{6\delta_2^2r\sqrt{\mathcal{D}_{\textit{root}}}}{\gamma_2\sigma_1\delta})} \nonumber,$}\end{equation} $L_3 = max\{L_1,L_2\}$, and $|\mathcal{D}_{\textit{root}}|$ denotes  the size of the root dataset,  $d$ is dimension of \textit{\textbf{W}}. 
Proof: See Appendix I.
\end{thm} 
\begin{Remark}
After undergoing several  rounds, the deviation between the global model learned through FL-OA and the optimal global model will be confined to a relatively narrow range. In other words, FL-OA is theoretically robust to Byzantine attacks.
\end{Remark}

\begin{thm} \label{16197}
Let $\sum_{e = 0}^{E-1}\lambda_{(e)} = \sum_{e = 0}^{E-1}\frac{\eta}{\alpha}(1-\frac{\eta}{\alpha})^{E-e} = 1-(1-\frac{\eta}{\alpha})^E=\lambda$,
after $E$  local iterations in Algorithm \ref{algorithm:test3445}, the local offset for each selected device $k \in \mathcal{K}^t$ is denoted as:
\begin{equation}\scalebox{0.85}{$
\textit{\textbf{w}}^t_{k,(E)} -\textit{\textbf{W}}^t = \alpha\lambda\widehat{\textit{\textbf{g}}}_{k}^{t-1}- \alpha\sum_{e=0}^{E-1}\lambda_{(e)}\widetilde{\textit{\textbf{g}}}_{k,(e)}^{t},\label{7262044}$}
\end{equation}
and the update for correction term $\widehat{\textit{\textbf{g}}}_{k}^{t}$ can be rewritten  as:
\begin{equation}\scalebox{0.85}{$
\widehat{\textit{\textbf{g}}}_{k}^{t} = (1-\lambda)\widehat{\textit{\textbf{g}}}_{k}^{t-1} + \sum_{e = 0}^{E-1}\lambda_{(e)}\widetilde{\textit{\textbf{g}}}_{k,(e)}^{t}.\label{3421}$}
\end{equation}
\end{thm} 

\begin{Remark}
The proof details can be referred to the Appendix II.
Following Theorem 2, let $\widehat{\textit{\textbf{g}}}_{k}^{-1} = 0$, we can obtain $\textit{\textbf{w}}^0_{k,(E)} -\textit{\textbf{W}}^0 = - \alpha\sum_{e=0}^{E-1}\lambda_{(e)}\widetilde{\textit{\textbf{g}}}_{k,(e)}^{0}$, where $ - \alpha\sum_{e=0}^{E-1}\lambda_{(e)}$ is the same for each selected device, and $\widetilde{\textit{\textbf{g}}}_{k,(e)}^{0}$ is different for each selected device, which shows that the inconsistency of model updates is independent the local learning rate $\eta$ and the importance $\alpha$.
In addition, we observe that for $\eta< \alpha$, the previous $\widetilde{\textit{\textbf{g}}}_{k,(e)}^{t}$ is weakened in the local offset with increasing $E$.
Moreover,  formula (\ref{7262044}) indicates that the local offset is  transferred to a exponential average of previous local gradients when applying the prox term.
From formula (\ref{3421}), we notice that $\widehat{\textit{\textbf{g}}}_{k}^{t}$ performs as a momentum term of the historical  updates before round $t$,  which shows that the correction term  is considered as an estimation of the local offset.
\end{Remark}
\section{EVALUATION}
In this section, we  describe the experimental settings, and  examine  the consistency, fidelity, and robustness of FL-OA.

\subsection{Experimental Settings}
\subsubsection{Datasets and Model Architecture}
Similar to prior works \cite{10648998}\cite{10495004}\cite{FLtrust}\cite{10458320}\cite{10891500}, we evaluate  the performance of FL-OA  using CIFAR10 and CIFAR100.
    Additionally, we use  the ResNet18 architecture \cite{targ2016resnet} as the local model  for experiments on   CIFAR10 \cite{krizhevsky2009learning} and CIFAR100 \cite{krizhevsky2009learning}, where ResNet18 is a deep residual network consisting of   convolutional layers,    average pooling layers, and   fully connected layers.
Since  ResNet18 contains more than $2.00 \times 10^7$ parameters, its model updates can be regarded as vectors in a high-dimensional parameter space.

\subsubsection{Non-IID and IID Settings}
To  evaluate the FL-OA performance, we simulate  both Non-IID  and IID setting in our experiments, respectively.
Furthermore, we  extract  a random subset from each dataset's training samples to  establish  the  root dataset $\mathcal{D}_{\textit{root}}$ based on the following settings.
\begin{itemize}
  \item \textbf{Non-IID setting}. 
  For the Non-IID setting, we employ the Dirichlet distribution \cite{lin2016dirichlet} to simulate Non-IID data.
Notably, the Dirichlet distribution has been widely applied for simulating data partitioning in FL.
 Formally, let DIR($\iota$) be the data distribution, where $\iota$ controls the level of data heterogeneity. 
A smaller $\iota$ corresponds to a higher level of Non-IID.
Specifically, we extract $\textbf{q}^j \sim$ DIR($\iota$) for class $j$ from the Dirichlet distribution.
  Each component   $q^j_k$ of $\textbf{q}^j$ determines the percentage of the examples of class $j$ assigned to device $k$.
This partitioning  ensures heterogeneous data distributions and sample sizes among different devices.
Moreover, under the Non-IID setting, the construction of the clean root dataset $\mathcal{D}_{\textit{root}}$ still follows the Dirichlet distribution.
Specifically, the entire dataset $\mathcal{D}$ is partitioned according to DIR($\iota$) into $(|\mathcal{K}|+1)$  subsets.
 Among them, $|\mathcal{K}|$ subsets are allocated to the devices, while the remaining subset is assigned to the OS as $\mathcal{D}_{\textit{root}}$.
This design ensures statistical consistency between the root dataset and the devices' datasets. 
For example, under the CIFAR10  with $|\mathcal{K}|=100$ devices and DIR(0.1), the constructed root dataset contains 382 samples, with class proportions approximately [0.02, 0.79, 0.02, 0.04, 0.02, 0.02, 0.01, 0.02, 0.01, 0.05]. 
Notably, we do not adopt a fixed partitioning  for the root dataset.
Instead, its construction is adaptively adjusted based on the parameter $\iota$ in DIR($\iota$) to accommodate different degrees of data heterogeneity.

  \item \textbf{IID setting}. 
In the IID setting, we randomly shuffle  the training samples  and distribute them equally to each device and OS.

\end{itemize}
\subsubsection{Setting of Byzantine Attacks}
Since the number of malicious  devices affects FL performance to varying degrees, we set  different percentages of malicious  devices in our  experiments.
Formally, let  $\operatorname{Att}$ be  percentage of malicious  devices,  computed as $\operatorname{Att} = \frac{|\mathcal{K}_b|}{|\mathcal{K}|}$, where $|\mathcal{K}_b|$ denotes the number of malicious devices.
Since FL-OA does not rely on the assumption that the majority of devices are benign, we set $\operatorname{Att}=50\%$ by default.
In addition, malicious devices maintain consistent adversarial   behavior  throughout the  federated training.
We consider   three types of attacks to simulate Byzantine attack behaviors.
(i) Gaussian Attacks \cite{fraboni2021free}: Some devices are unwilling to contribute their own model updates.  To fulfill the task of TS, they  upload model updates  generated by Gaussian noise. 
 Specifically, each parameter in the model update is replaced with a randomly selected value  from a Gaussian distribution.
(ii) Neurotoxin Attacks \cite{zhang2022neurotoxin}: 
Malicious devices   project  their  updates onto the coordinate set defined by the top $75\%$  largest-magnitude parameters of the previous-round global model, and sets the update in  other coordinates to zero.
(ii) Focused-Flip Attacks \cite{fang2023vulnerability}: 
Malicious devices     flip the signs of a small portion (set to 50\%) of  parameters in their model updates, while the signs of the remaining parameters remain unchanged.

\subsubsection{Implementation Details}
We adopt the following hyper-parameter settings.
Specifically, the total number of devices is set to 100, and 20\% of the devices are randomly selected in each round of local training.
In addition, the batch size is set to 50, the number of local iterations $E$ is set to 5, the local learning rate $\eta$  is set to 0.1, and the global learning rate $\mu$ is set to 1. 
 The weighting coefficients $\alpha$ and  $\beta$ are set to 0.1, and the proportion of selected coordinates $\kappa$  is set as 30\%.
 The number of communication rounds $T$ is set to 500.
Furthermore, all benign devices use the same hyper-parameter configuration, and are initialized with the same model parameters $\textit{\textbf{W}}^{\textit{init}}$.
Unless otherwise specified, the above settings are used as the default configuration.


\subsubsection{Baselines}
To   evaluate the effectiveness of FL-OA, we compare it with  five defense schemes:  MKrum \cite{NIPS2017f4b9ec30}, FLTrust \cite{FLtrust}, FL-Auditor \cite{10475552}, AlignIns \cite{xu2025detecting}, and FLgym \cite{11298320}.

\subsubsection{Evaluation Metrics}
Since Byzantine attacks are intended to degrade the performance of the global model, we adopt the  accuracy of the global model on  datasets as the  metric to evaluate the defense effectiveness of different schemes. 
 A higher accuracy indicates better performance of the corresponding scheme.
Moreover, we define an additional metric, denoted as $\operatorname{Acc}_{50\%-0\%}$ to measure  the accuracy difference  between the setting with 50\% malicious devices and the setting without malicious devices.
A smaller value of $\operatorname{Acc}_{50\%-0\%}$ indicates that the scheme exhibits  less accuracy degradation when  subjected to Byzantine attacks, thereby demonstrating greater robustness.

\subsection{Experimental Results}
\begin{figure*}[!t]
  \centering   
  \subfloat[CIFAR10, IID.] 
  {
      \label{523102701}  \includegraphics[width=0.24\linewidth]{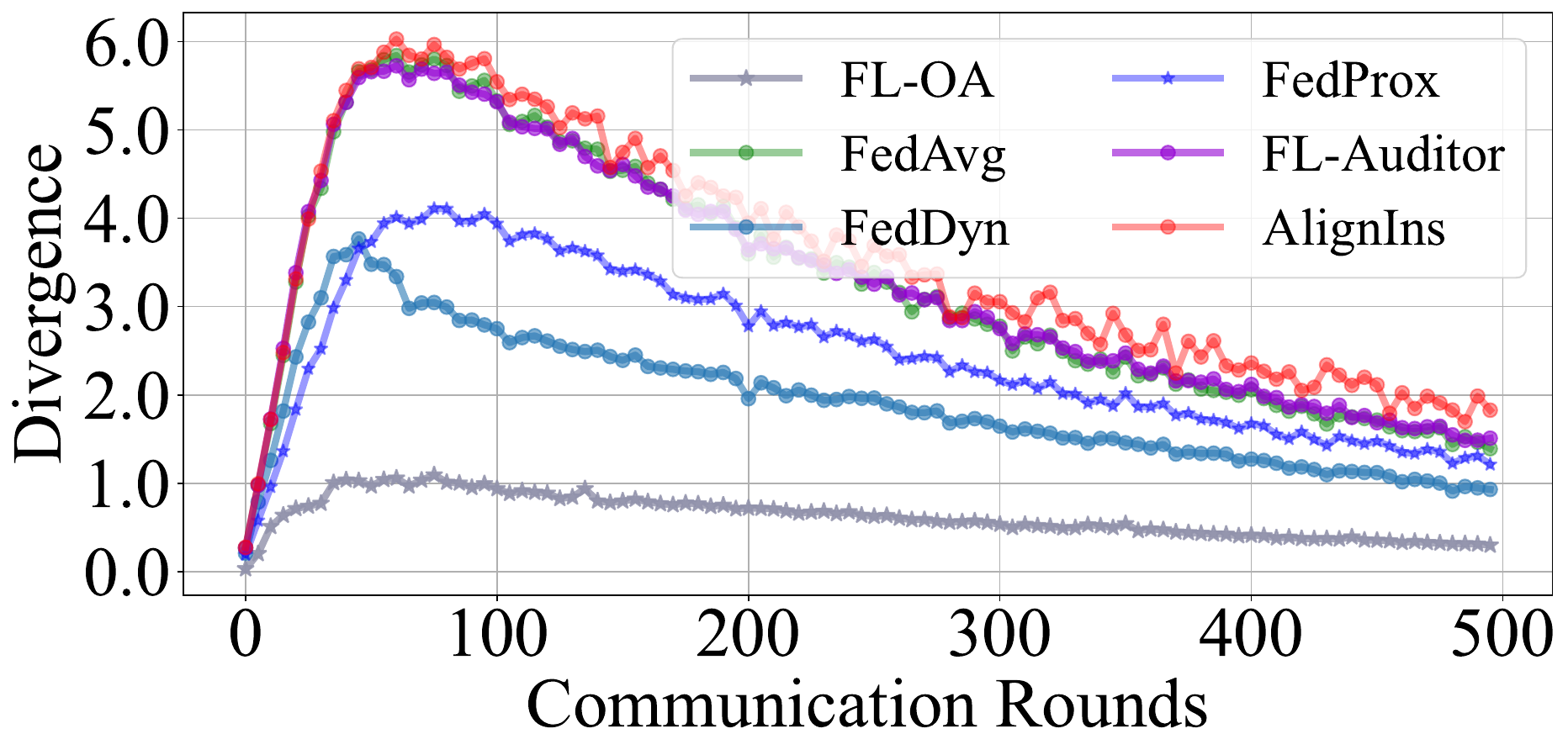}
  }
  \subfloat[CIFAR10, DIR(0.6).] 
  {
      \label{52310271}  \includegraphics[width=0.24\linewidth]{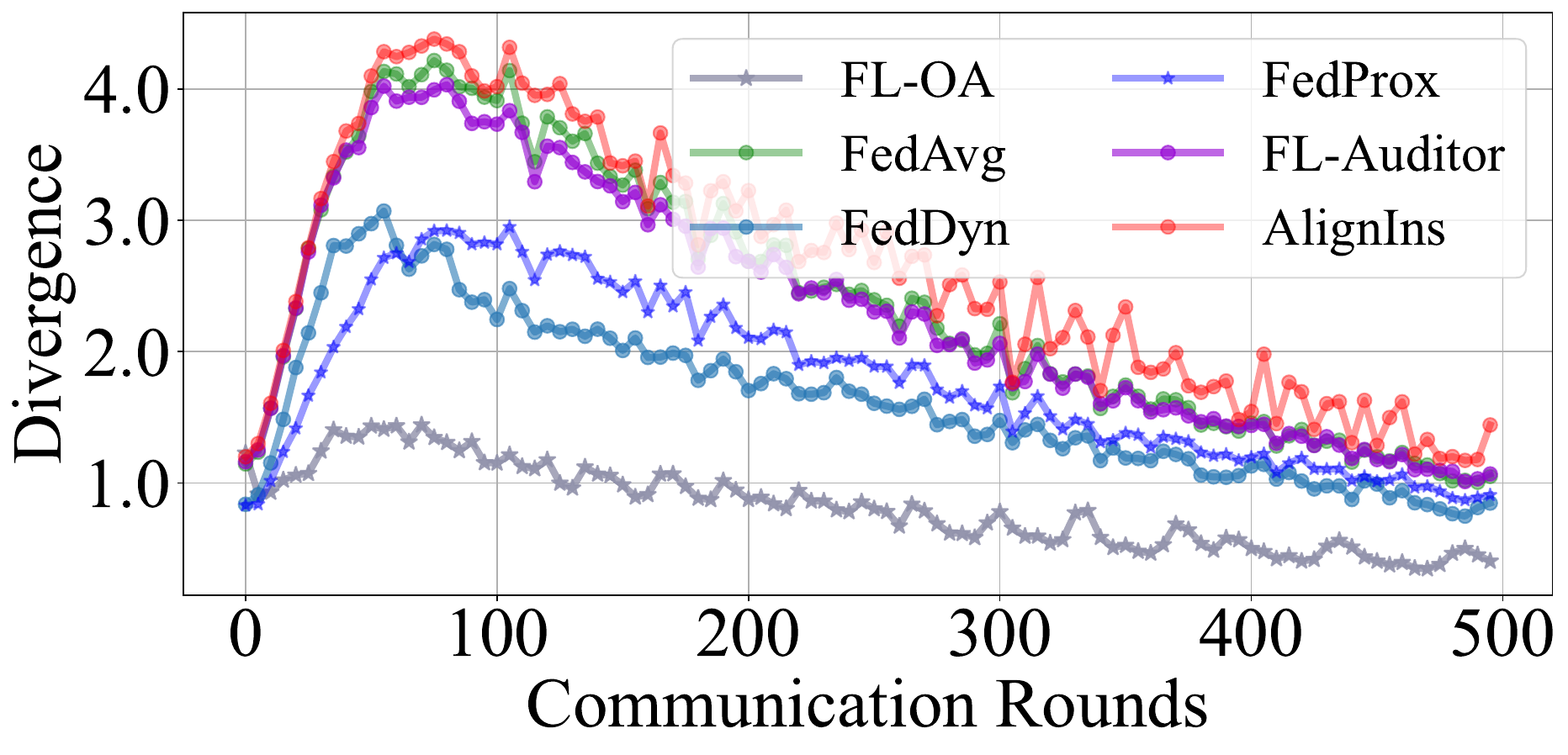}
  }
  \subfloat[CIFAR10, DIR(0.3).] 
  {
      \label{52310272}  \includegraphics[width=0.24\linewidth]{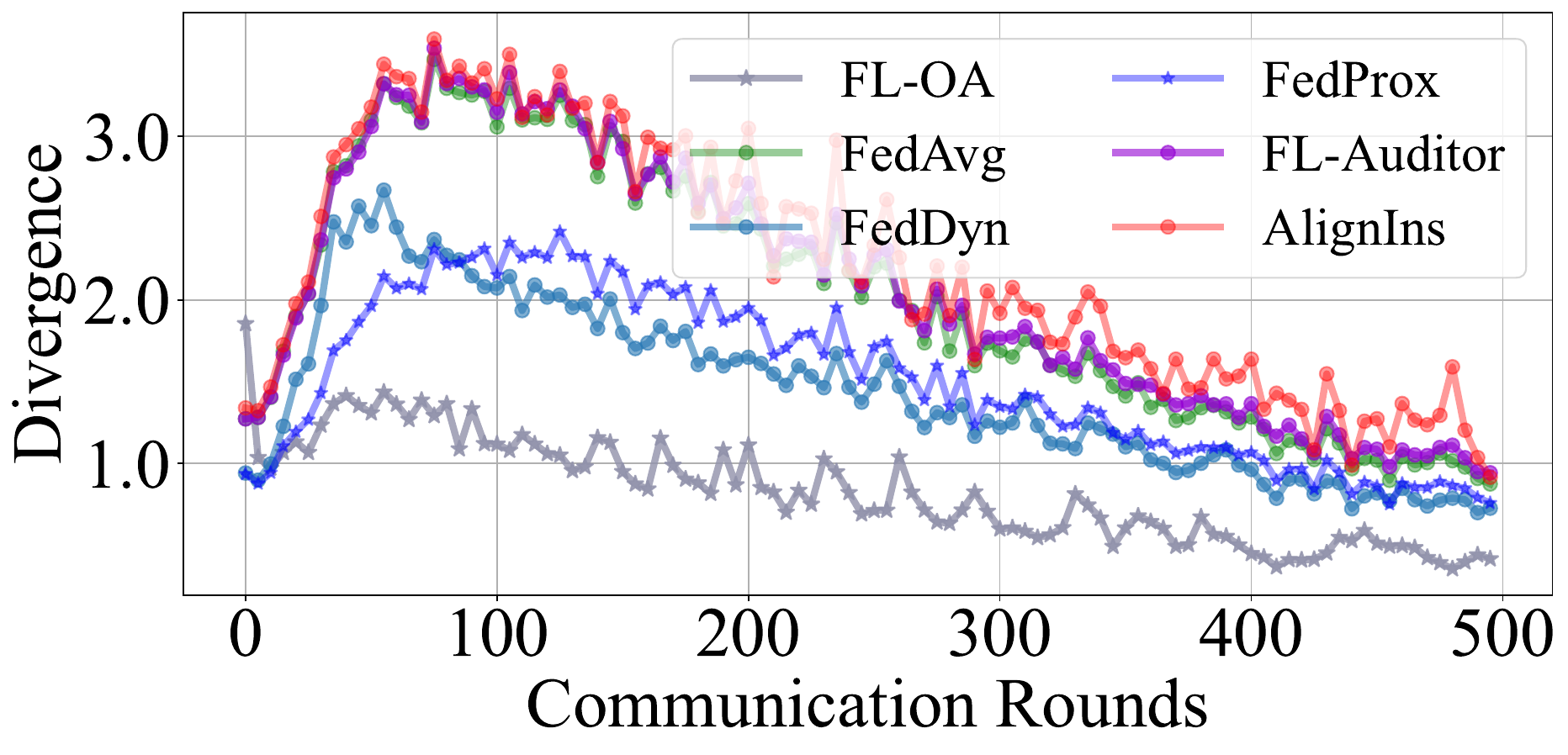}
  }
  \subfloat[CIFAR10, DIR(0.1).] 
  {
      \label{52310272}  \includegraphics[width=0.24\linewidth]{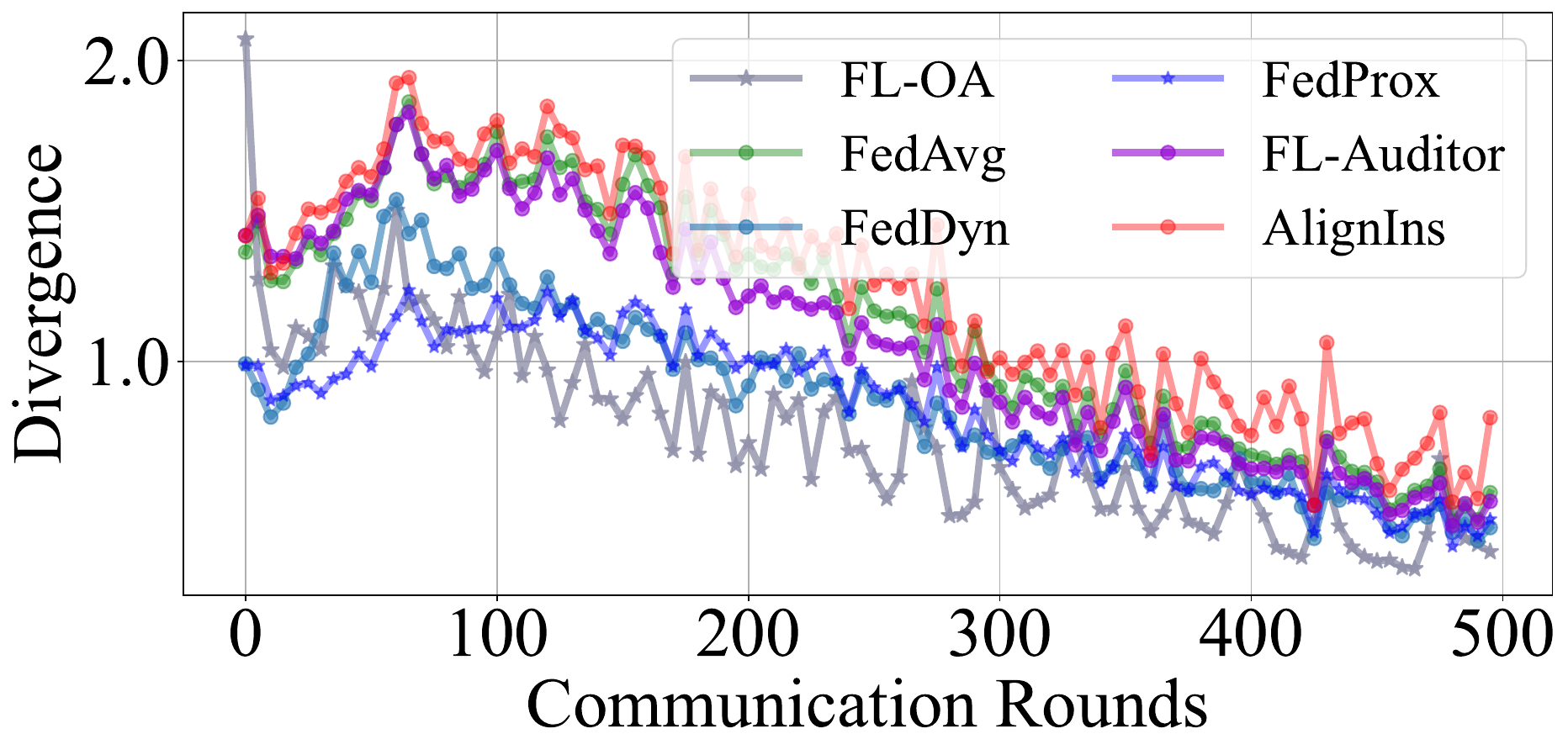}
  }

\vspace{0pt}
  \subfloat[CIFAR100, IID.] 
  {
      \label{523102701001}  \includegraphics[width=0.24\linewidth]{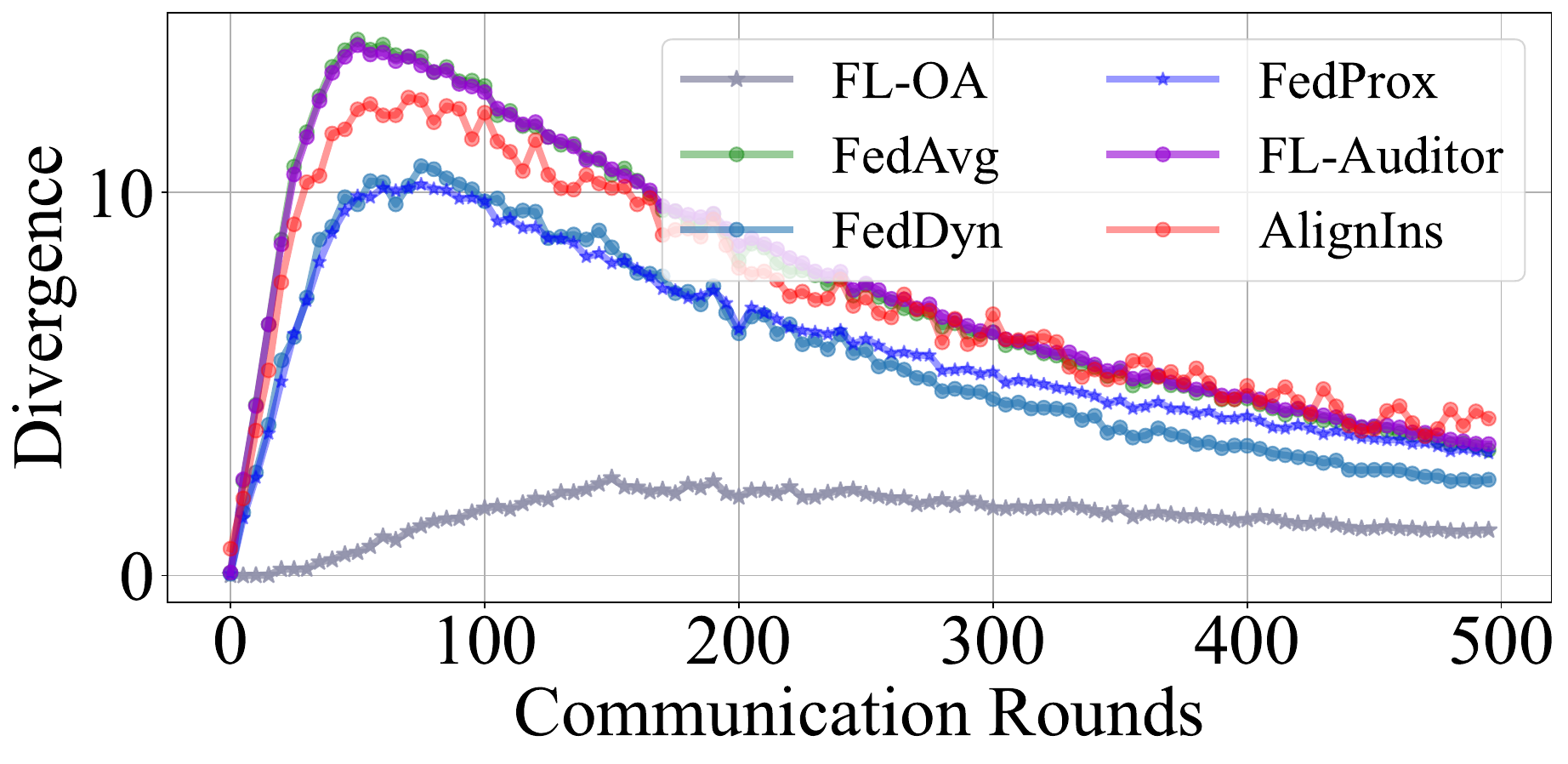}
  }
  \subfloat[CIFAR100, DIR(0.6).] 
  {
      \label{523102701002}  \includegraphics[width=0.24\linewidth]{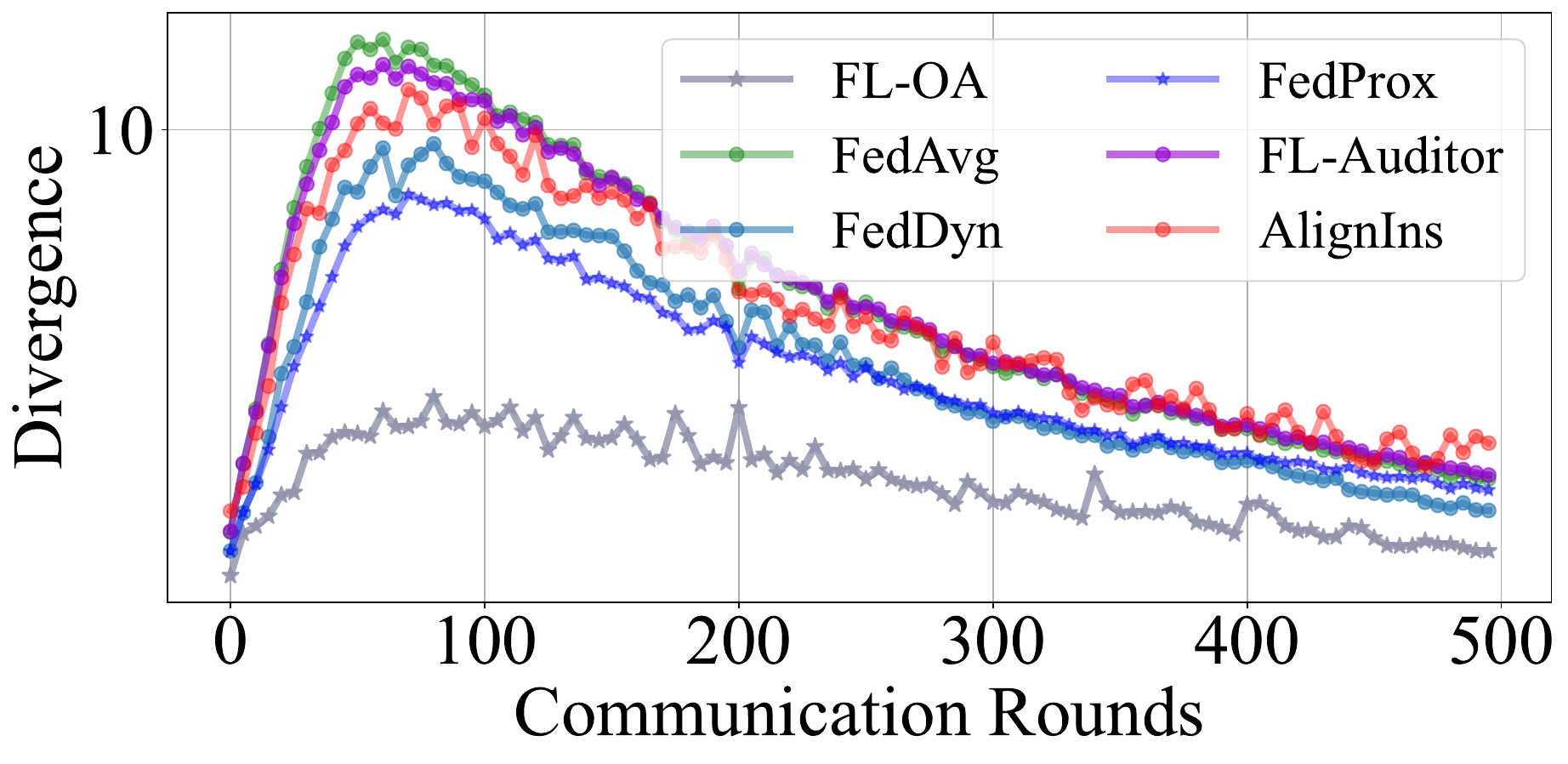}
  }
  \subfloat[CIFAR100, DIR(0.3).] 
  {
      \label{523102701003}  \includegraphics[width=0.24\linewidth]{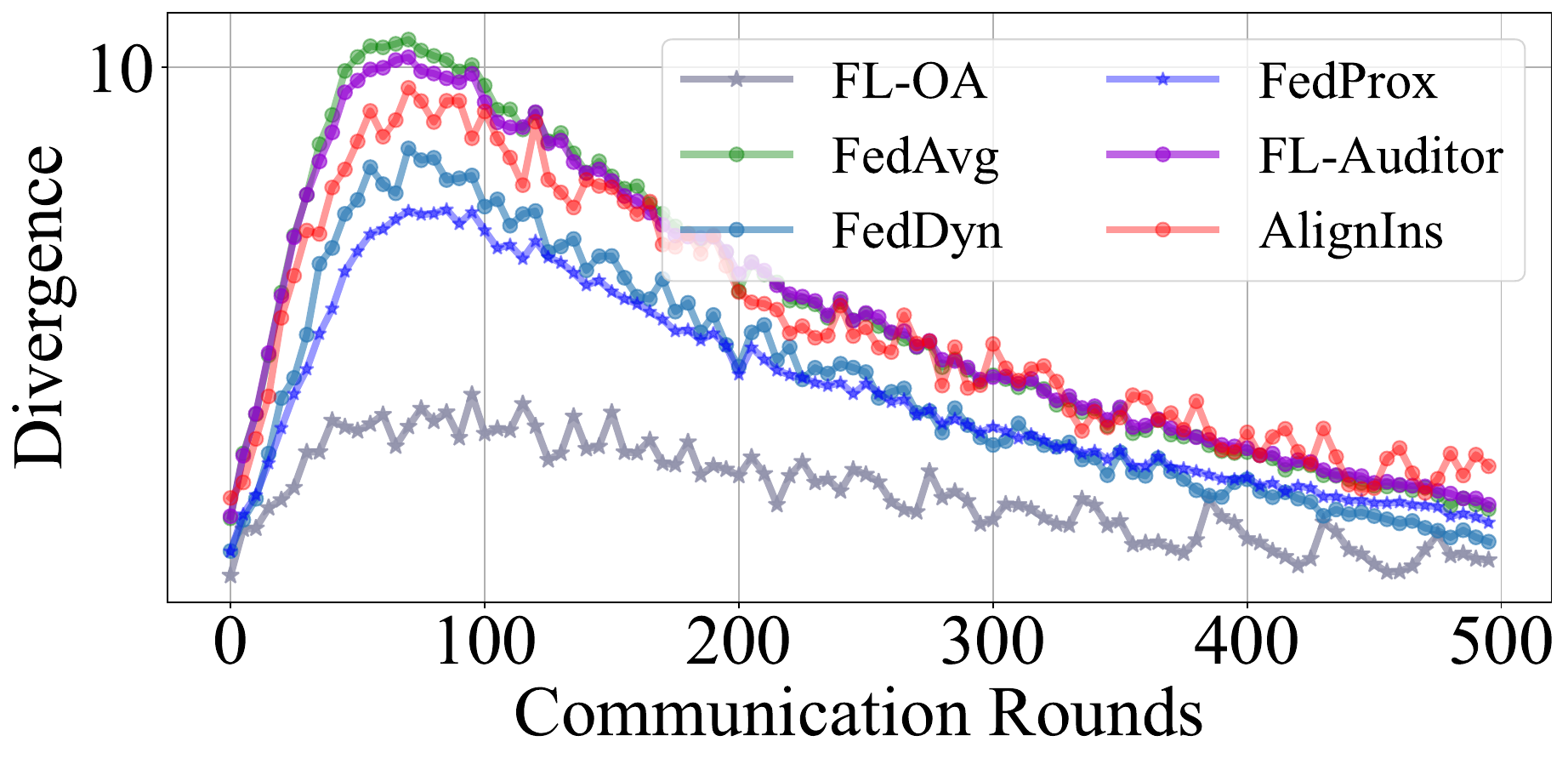}
  }
  \subfloat[CIFAR100, DIR(0.1).] 
  {
      \label{523102701003}  \includegraphics[width=0.24\linewidth]{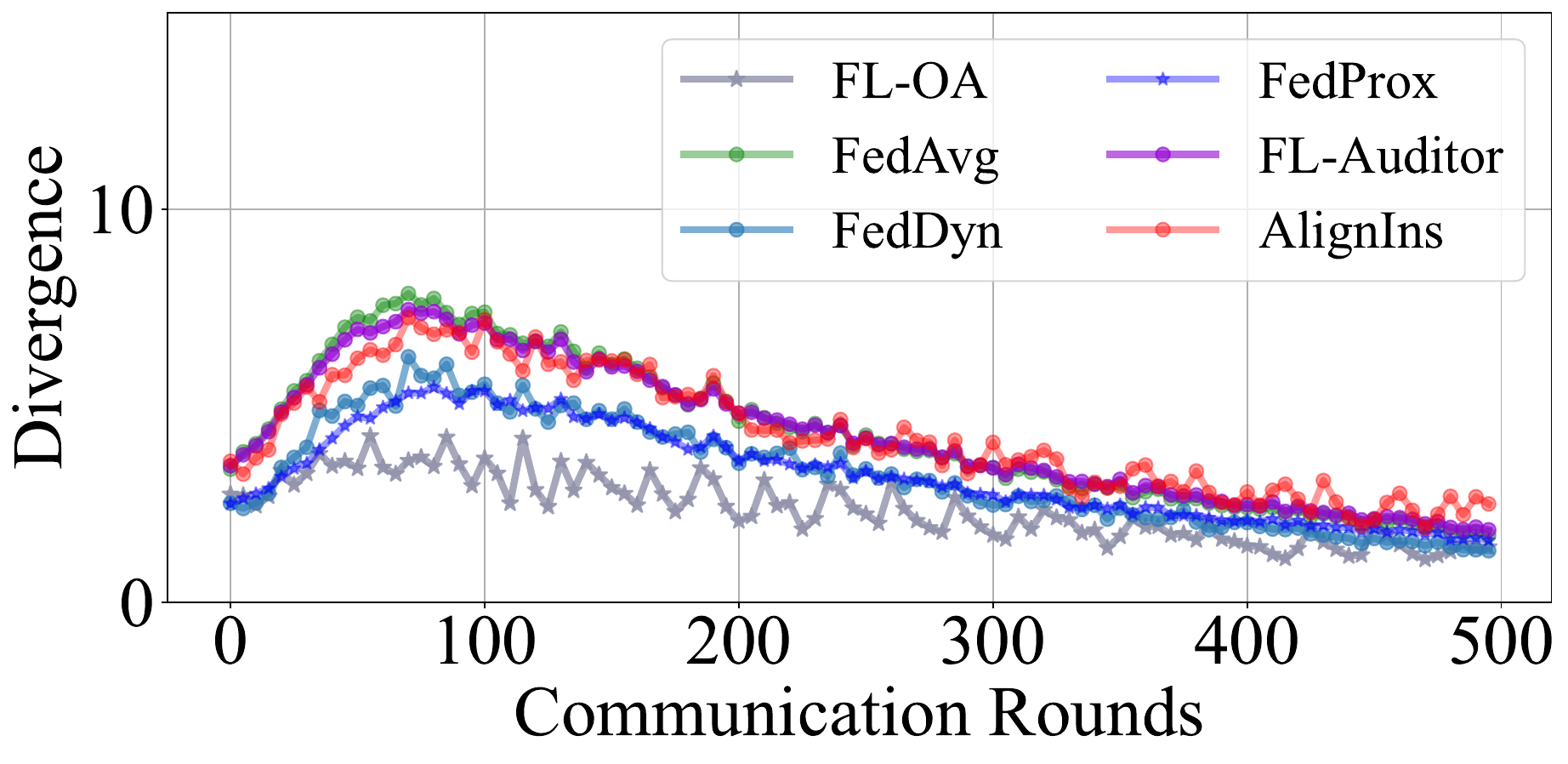}
  }
\caption{Divergence  comparison between FL-OA and existing schemes  in both   IID and  Non-IID settings without  attacks.}
\vspace{-10pt}
\label{5231027}
\end{figure*}
\subsubsection{Consistency of FL-OA}
To evaluate the consistency of model updates in FL, we introduce divergence as a evaluation metric, which is defined as:
\begin{equation}\scalebox{0.85}{$\operatorname{Div}:= \sum_{k\in \mathcal{K}^t}\frac{\|\widetilde{\textit{\textbf{u}}}^t_k-\bar{\textit{\textbf{u}}}^t\|_2^2}{|\mathcal{K}^t|},$}
\end{equation}
where $\bar{\textit{\textbf{u}}}^t= \mu \frac{1}{\sum_{k \in \mathcal{K}^t}s^t_k}\sum_{k \in \mathcal{K}^t}s^t_k\widetilde{\textit{\textbf{u}}}^t_k$.
A lower divergence value $\operatorname{Div}$ indicates a higher degree of consistency among model updates.
We evaluate the divergence of FL-OA  across  four data distributions, namely  IID,  DIR($0.6$), DIR($0.3$), and DIR($0.1$).
For  comparison, we evaluate several  non-robust FL schemes, including  FedAvg \cite{mcmahan2017communication}, FedDyn \cite{acar2021federated}, and FedProx \cite{li2020federated}. We also include several defense schemes, namely FL-Auditor \cite{10475552} and AlignIns \cite{xu2025detecting}.
Notably,   these experiments are conducted in the absence of  attacks, i.e., $\operatorname{Att} = 0\%$.

\begin{figure*}
  \centering   
  \subfloat[CIFAR10, IID.] 
  {
      \label{6232157}  \includegraphics[width=0.24\linewidth]{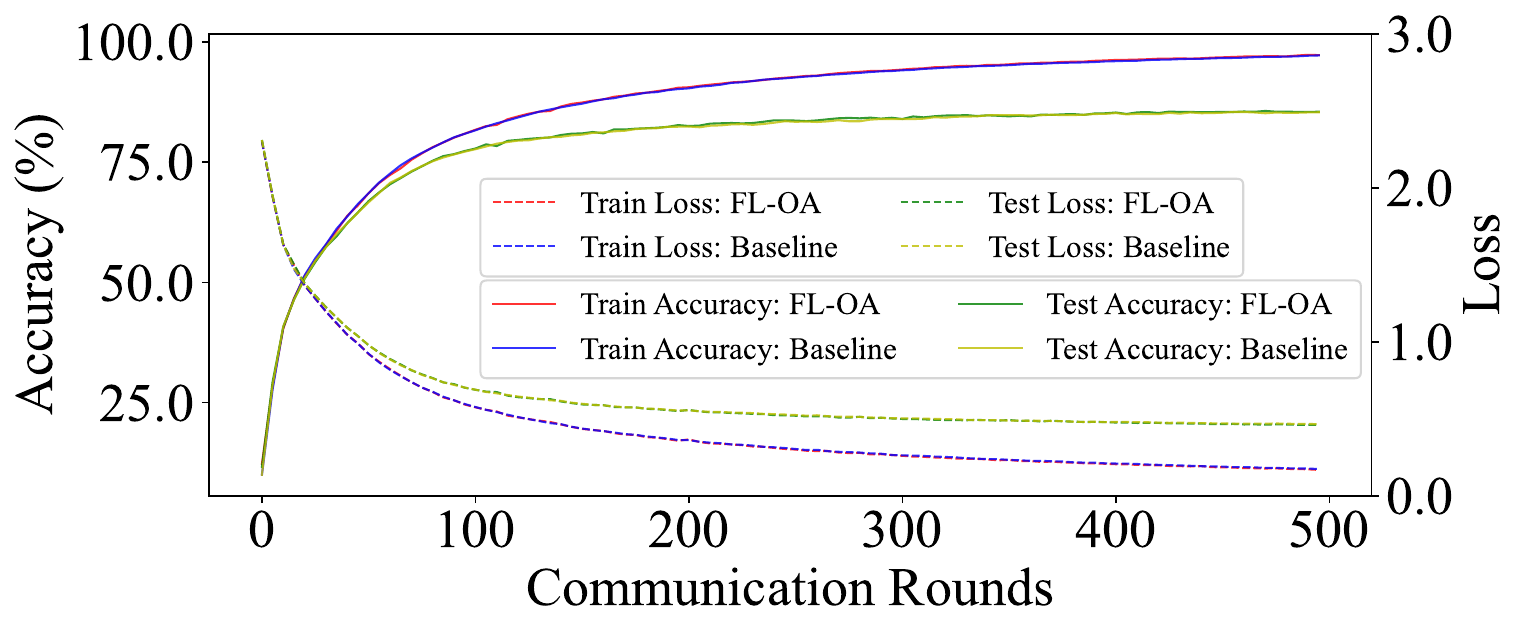}
  }
    \subfloat[CIFAR10,  DIR(0.6).] 
  {
      \label{6232157}  \includegraphics[width=0.24\linewidth]{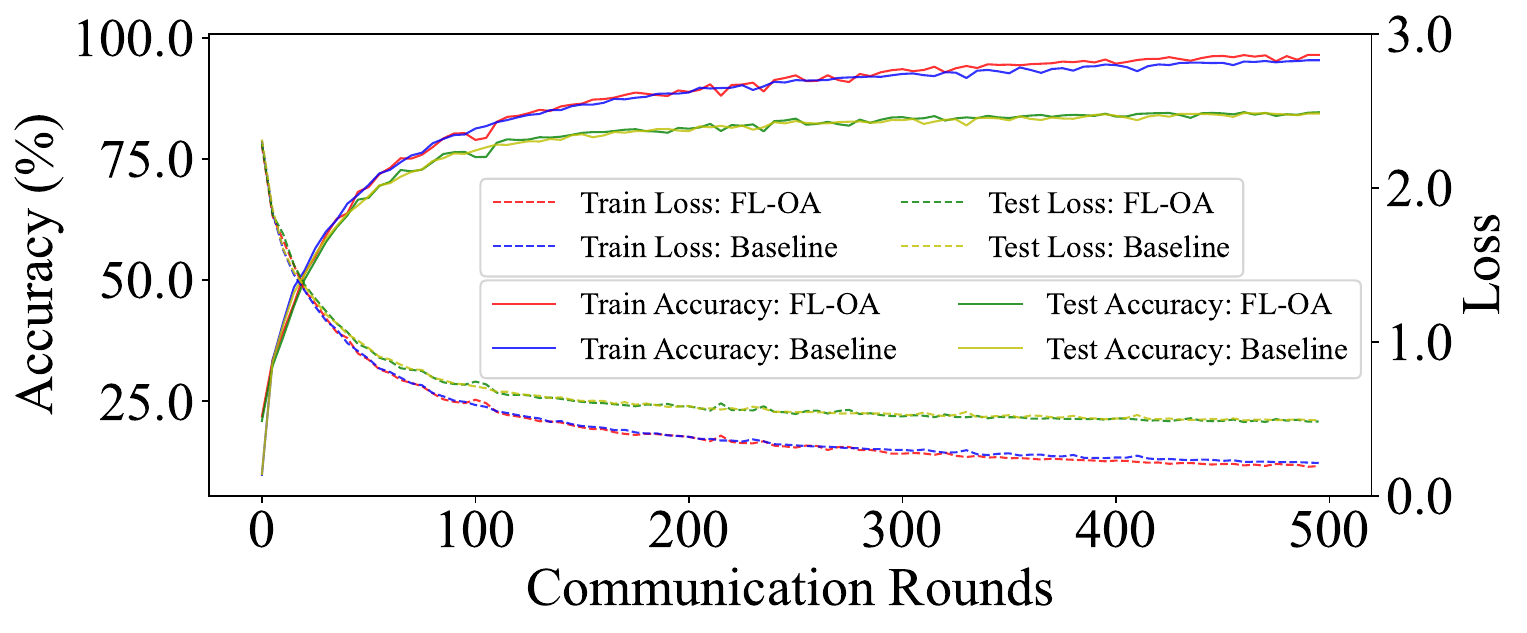}
  }
    \subfloat[CIFAR10,  DIR(0.3).] 
  {
      \label{6232157}  \includegraphics[width=0.24\linewidth]{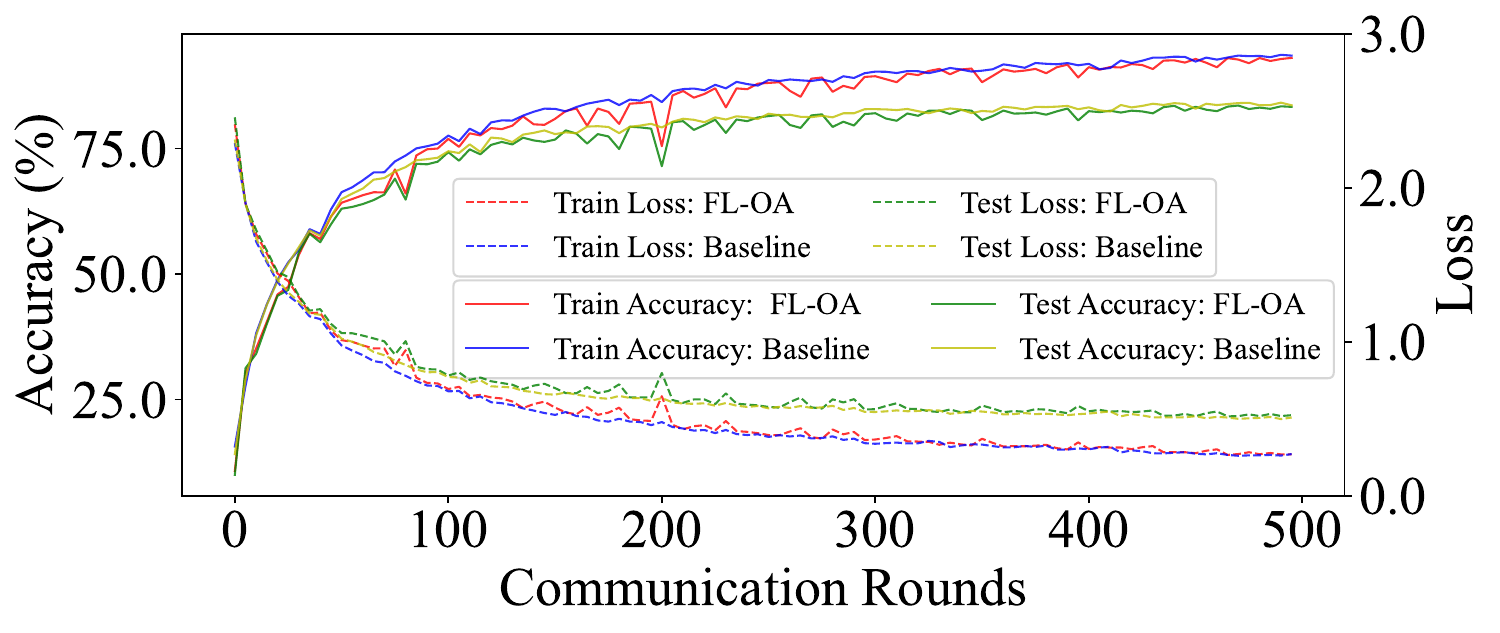}
  }
    \subfloat[CIFAR10,  DIR(0.1).] 
  {
      \label{6232157}  \includegraphics[width=0.24\linewidth]{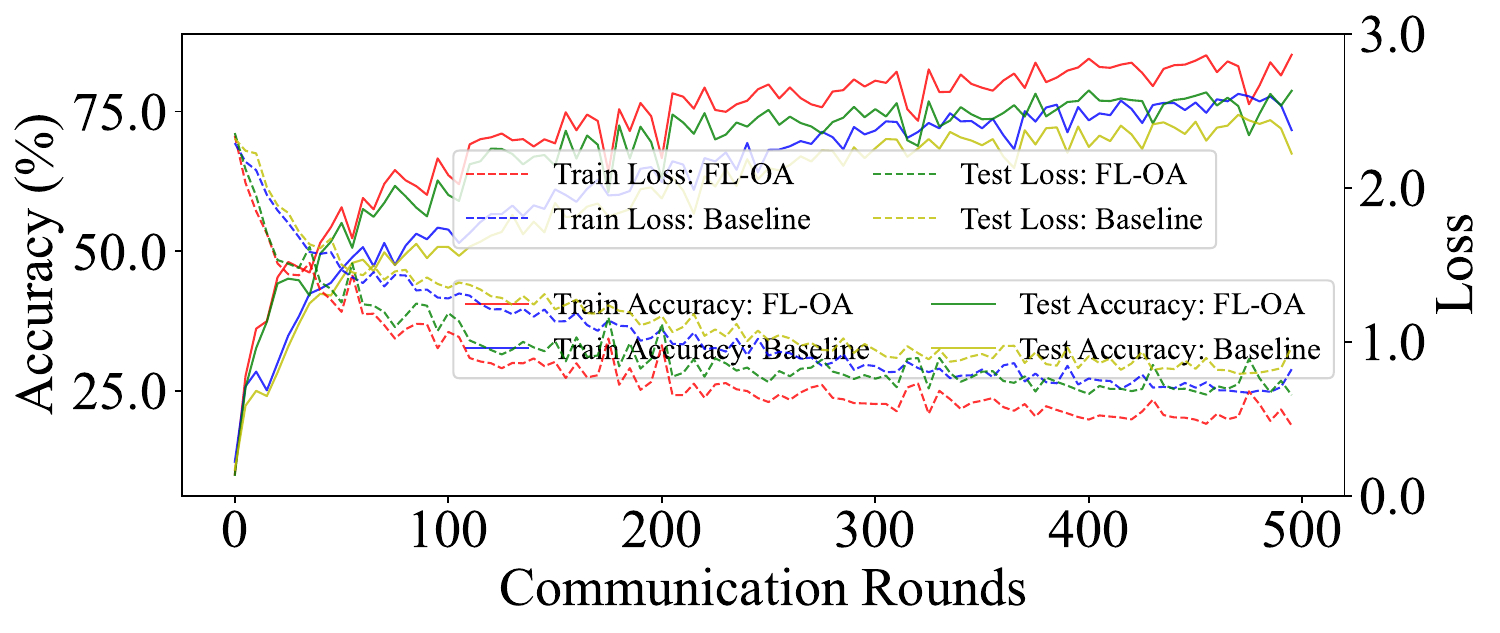}
  }

  \subfloat[CIFAR100, IID.] 
  {
      \label{6232157}  \includegraphics[width=0.24\linewidth]{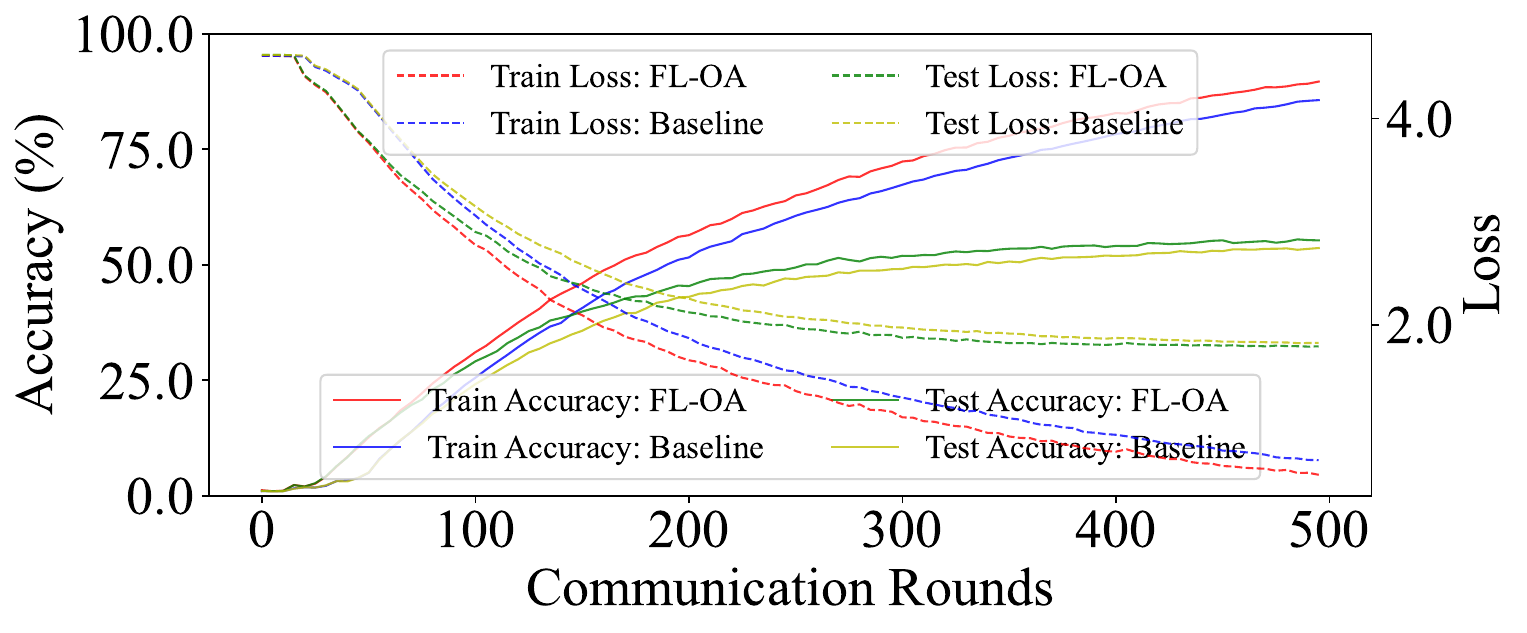}
  }
  \subfloat[CIFAR100,  DIR(0.6).] 
  {
      \label{6232157}  \includegraphics[width=0.24\linewidth]{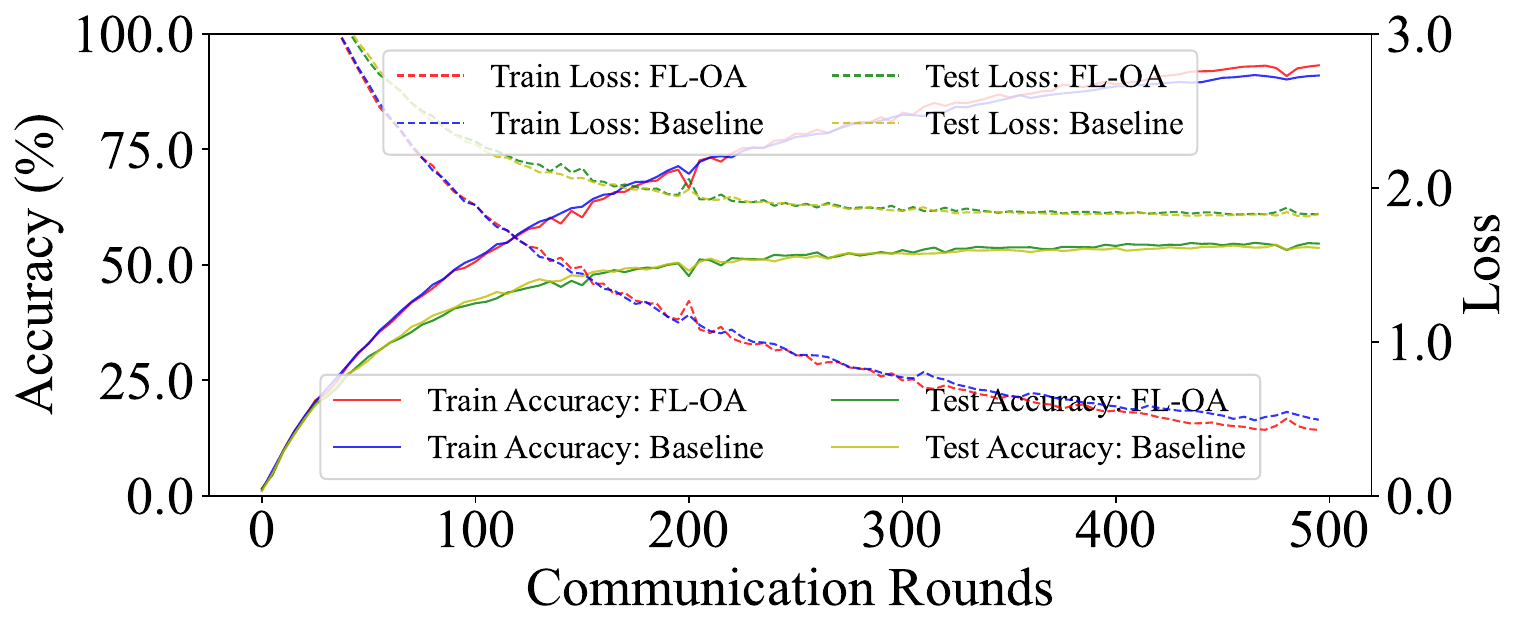}
  }
  \subfloat[CIFAR100,  DIR(0.3).] 
  {
      \label{6232157}  \includegraphics[width=0.24\linewidth]{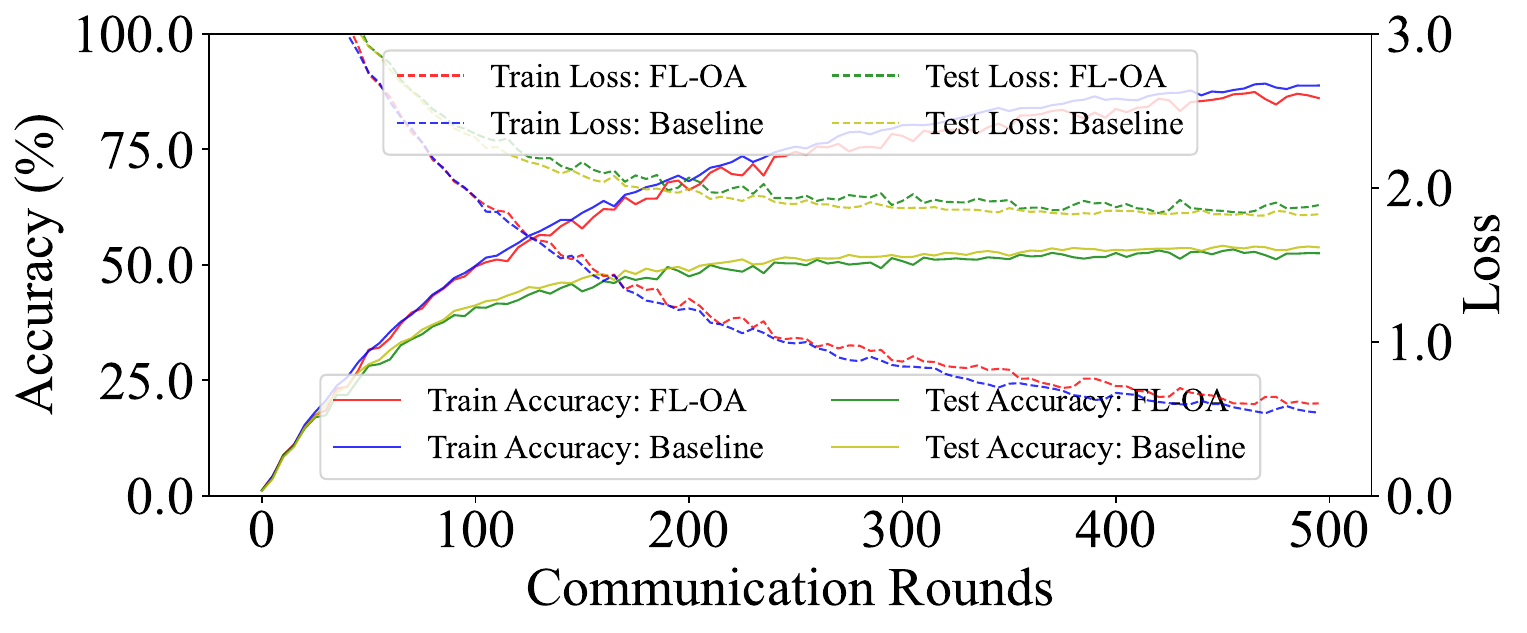}
  }
  \subfloat[CIFAR100,  DIR(0.1).] 
  {
      \label{6232157}  \includegraphics[width=0.24\linewidth]{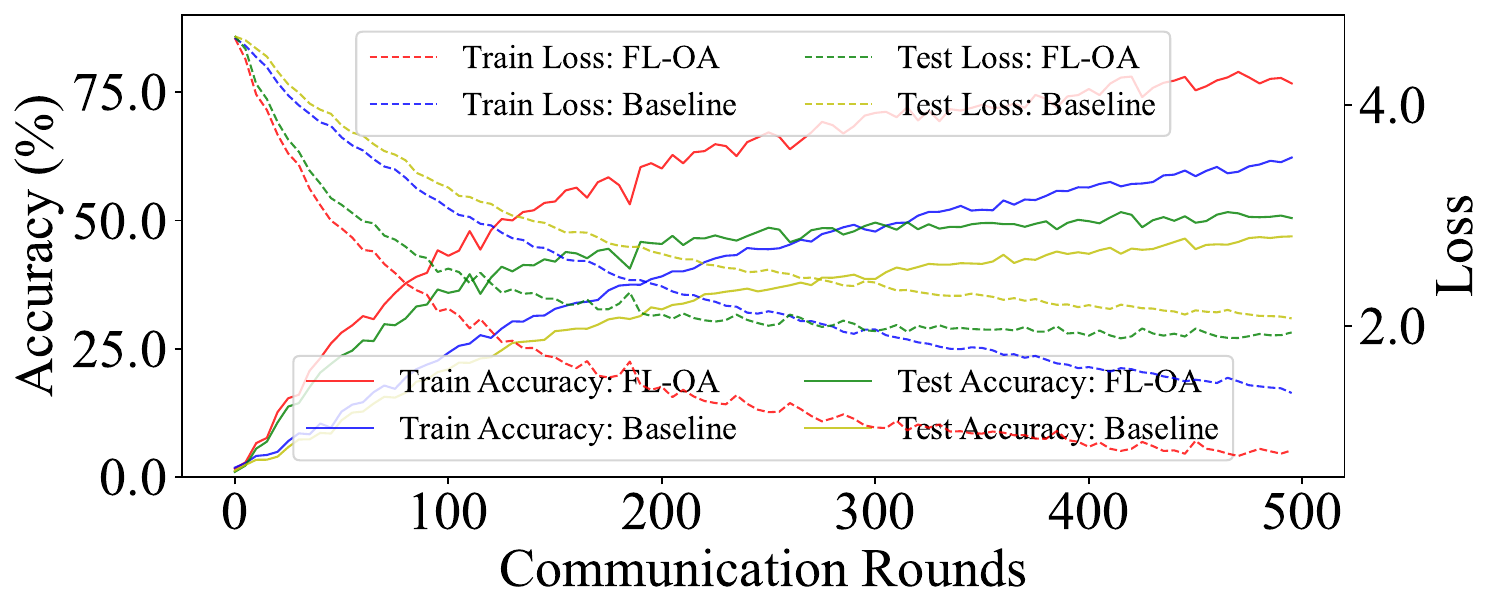}
  }
\caption{Convergence of the proposed FL-OA  over CIFAR10 and CIFAR100  in both IID and Non-IID settings.}
\label{62321571234}
\vspace{-15pt}
\end{figure*}

Fig. \ref{5231027} illustrates the divergence results of different schemes on CIFAR10 and CIFAR100 under both IID and Non-IID settings. 
It can be observed that, for all schemes, the divergence  among model updates still exists   under IID data. 
This is because, although the class distributions across devices are identical, the feature distributions of  samples held by  devices may still differ \cite{10286887}. 
In other words, even samples belonging to the same class may exhibit differences in their fine-grained features.
Furthermore, we observe that FL-OA yields lower divergence than the other schemes across various data distributions on both CIFAR10 and CIFAR100.
This indicates that the gradient ascent step  and the correction term in FL-OA effectively mitigate the divergence among model updates.
Notably, FedAvg exhibits the largest divergence.
In contrast, the other non-robust FL schemes achieve lower divergence than FedAvg because their regularization terms help improve the consistency of model updates.
In addition, since AlignIns and FL-Auditor  do  not enhance the update consistency during local training, their $\operatorname{Div}$ values are similar to those of FedAvg.
Based on these results, FL-OA is better positioned to defend against Byzantine attacks,  as further demonstrated in the following experiments.

In addition, as shown in Fig. \ref{5231027}, we observe that the divergence on CIFAR10 and CIFAR100 first grow and then fall during federated  training.
At the beginning of training, because of differences in local data distributions, each device optimizes its model  toward its own optimum, leading to significant differences among  model updates in terms of direction and magnitude.
Consequently, the divergence among device' model updates to increase rapidly.
As the global model gradually approaches the local optima, the direction and magnitude of each device's model updates gradually decrease, thereby reducing the divergence among them.
Thus, the divergence exhibits a trend of first increasing and then decreasing during the training process.
Furthermore, we observe that the maximum divergence decreases as data heterogeneity increases. 
This is because, under the IID setting, each device is trained to fit all classes, so most model parameters are involved in the update process.
In contrast, under the Non-IID settings, each device usually focuses on the dominant classes in its local data, while parameters associated with the missing classes are rarely updated.
Thus, only a smaller subset of parameters is  actively involved in the updates. 
Therefore, greater data heterogeneity leads to a lower maximum divergence among devices' model updates.

\subsubsection{Fidelity of FL-OA}
To highlight the fidelity of FL-OA,  we construct a version of FL-OA without the defense mechanism as a reference (i.e., FL-OA with only standard average aggregation of model updates).  
For the sake of clarity, we simplify the ``FL-OA without defense'' to the ``Baseline'' below.
To this end, we evaluate the performance of both FL-OA and Baseline in the absence of Byzantine attacks.
If FL-OA's performance is close to that of the Baseline,  FL-OA is considered to meet the fidelity requirement.
Fig. \ref{62321571234} presents the accuracy and loss of the two schemes on the training and test sets, respectively.
We observe that, overall, FL-OA and Baseline exhibit similar accuracy and loss across different data distributions, indicating that FL-OA preserves the model learning performance well.
Thus, Fig. \ref{62321571234} demonstrates the fidelity of FL-OA.
Notably, the performance of FL-OA is significantly higher than that of the Baseline under the DIR(0.1) setting on both CIFAR10 and CIFAR100, which still shows that FL-OA satisfies the fidelity.
This is because the  DIR (0.1) setting exacerbates the inconsistency of model updates, and some devices  whose updates deviate from the majority  are removed during the auditing.
In other words, these removed devices  are considered anomalous.
Therefore, model updates with large deviations  do not participate in the aggregation, which causes the performance of FL-OA to be higher than Baseline.
Furthermore, we notice that the test  accuracy of FL-OA and Baseline on CIFAR100 is significantly lower than the training accuracy, which is called over-fitting \cite{yu2020salvaging}.
This is because the model learns the noise  in the training samples, rather than  the essential pattern of the  samples.

\subsubsection{Robustness of FL-OA}
\begin{figure}[!t]
  \centering   
  \subfloat[CIFAR10, Gaussian attacks.] 
  {
      \label{7191143}  \includegraphics[width=0.45\linewidth]{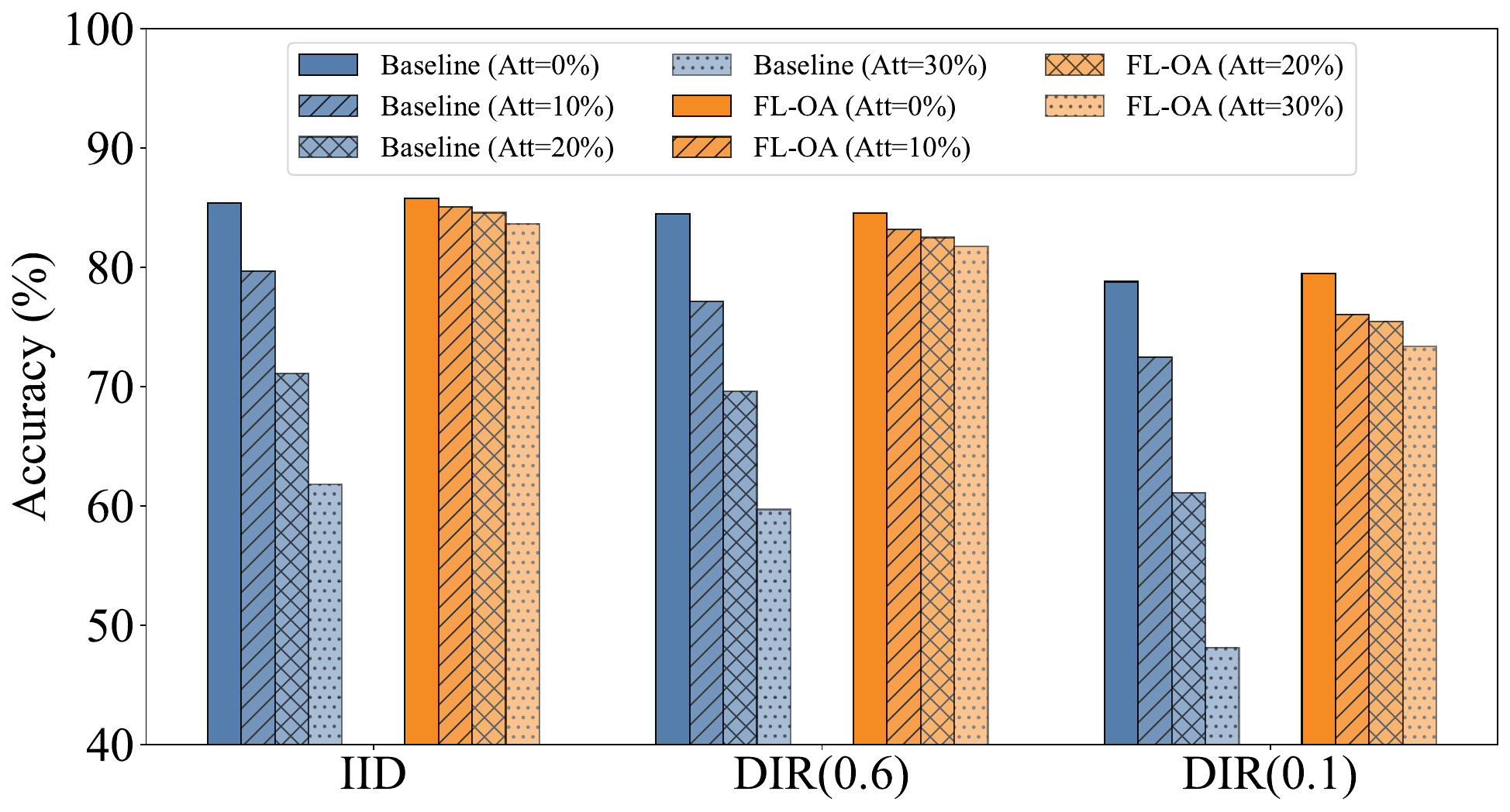}
  }
    \subfloat[CIFAR10, Neurotoxin  attacks.] 
  {
      \label{7191146}  \includegraphics[width=0.45\linewidth]{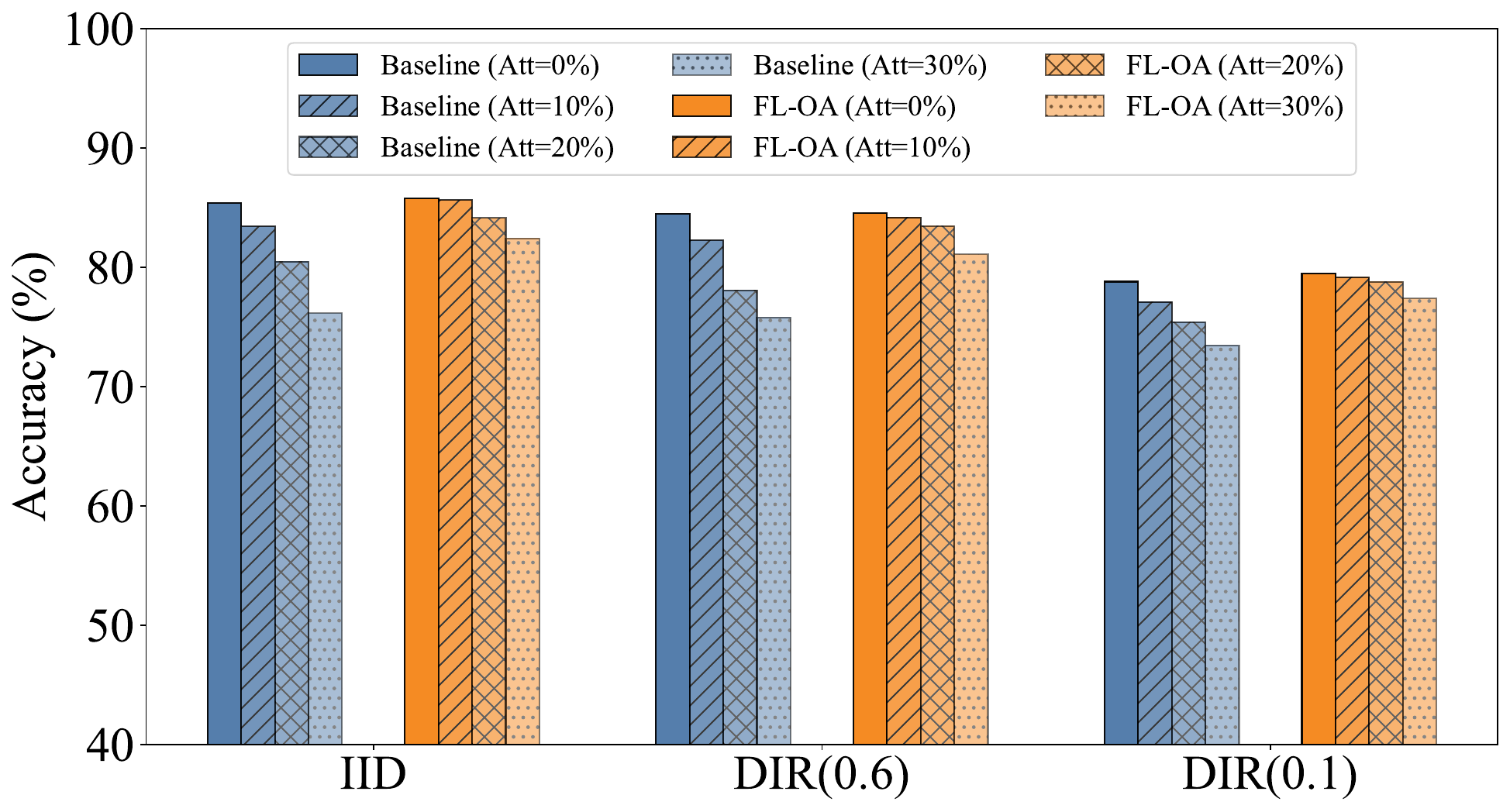}
  }
  
  \subfloat[CIFAR100, Gaussian attacks.] 
  {
      \label{7191144}  \includegraphics[width=0.45\linewidth]{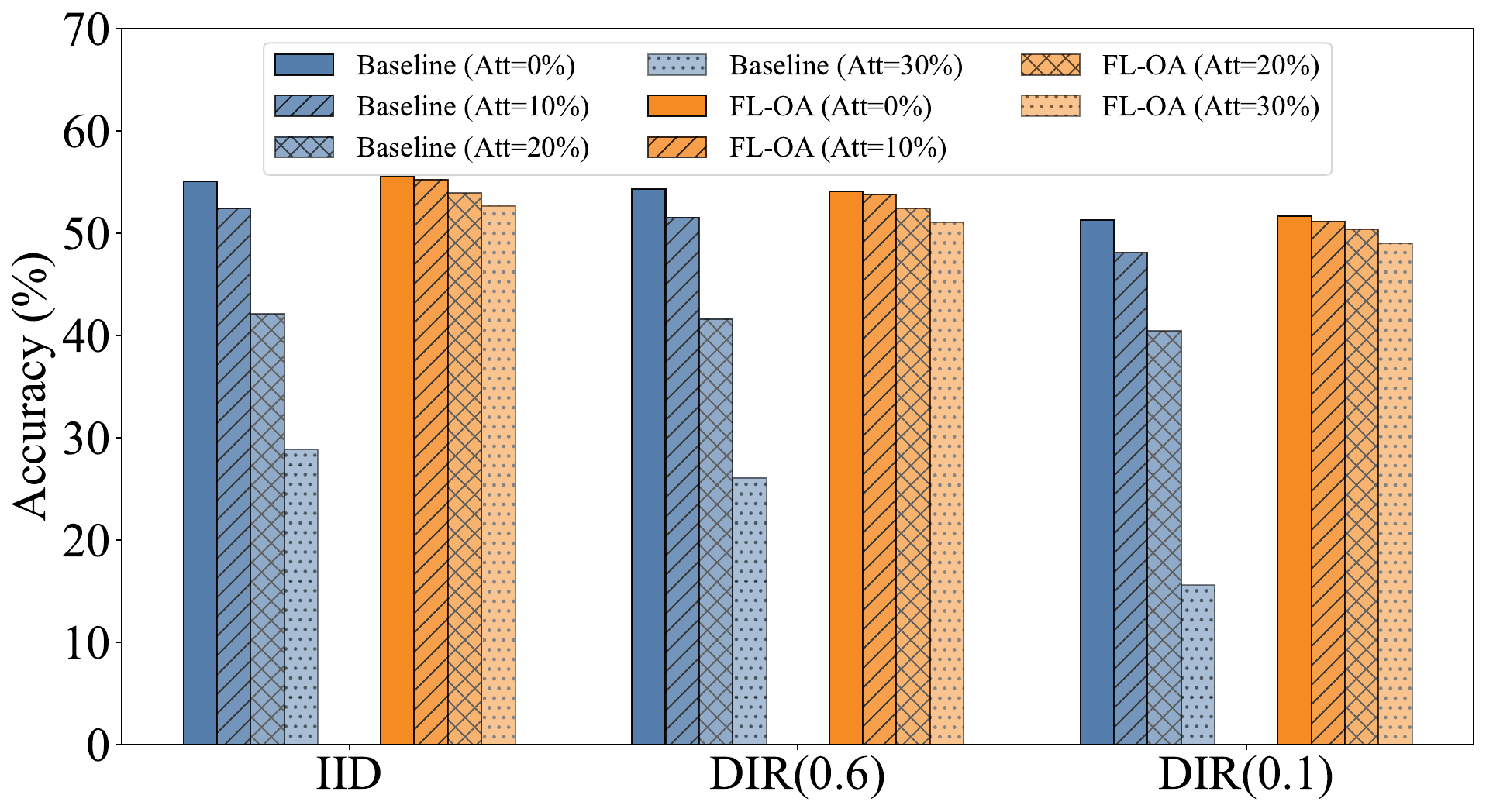}
  }
  \subfloat[CIFAR100, Neurotoxin  attacks.] 
  {
      \label{7191147}  \includegraphics[width=0.45\linewidth]{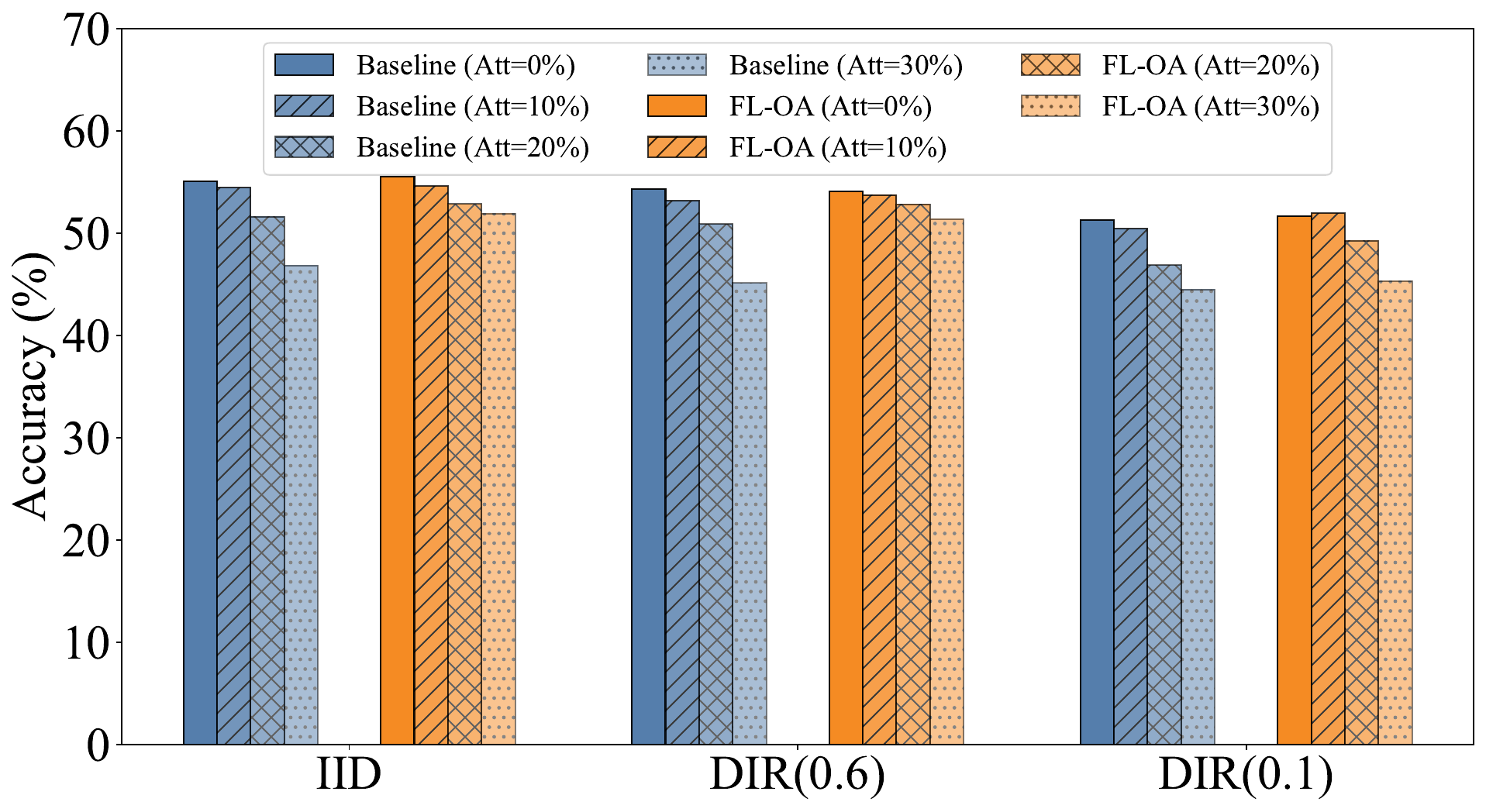}
  }
\caption{Test accuracy of FL-OA for different numbers of malicious devices over  CIFAR10 and CIFAR100  (IID and Non-IID).}
\vspace{-10pt}
\label{7191207}
\end{figure}

To evaluate the performance of FL-OA under Byzantine attacks, we  compare its  performance under Gaussian and Neurotoxin attacks across different proportions of malicious  devices. 
Specifically, we conduct experiments on the CIFAR10 and CIFAR100   with malicious device proportions of 0\%, 10\%, 30\%, and 50\%.
FL-OA is compared with a baseline variant, i.e., FL-OA using only average aggregation of model updates.
The experimental results are shown in Fig. \ref{7191207}.
We observe that under both Gaussian and neurotoxin attacks, the performance of the baseline  declines significantly, whereas FL-OA maintains the relatively stable performance, demonstrating its strong robustness.
In particular, under neurotoxin attacks, the accuracy of the global model learned by FL-OA is nearly identical to that achieved without attacks.
This is because TS can effectively mitigate the impact of malicious model updates during the aggregation process.
In contrast, the Gaussian attack has a more significant impact on the performance of FL-OA.
This is because Gaussian attacks perturb all parameters in the model updates, thereby causing more comprehensive damage to the global model, whereas Neurotoxin mainly poisons only a subset of critical parameters, making its impact relatively limited.

\begin{table*}[]
\centering
\caption{Test accuracy \% (Top-1) comparison between   FL-OA and existing schemes under Gaussian Attacks. }
\vspace{-10pt}
    \tabcolsep= 0.065cm
      \renewcommand{\arraystretch}{0.4}
\scalebox{0.8}{
\begin{tabular}{c|c|ccc|ccc|ccc|ccc|ccc|ccc}
\midrule
\multirow{2}{*}{ \makecell[c]{Dataset}} & \multirow{2}{*}{ \makecell[c]{$\operatorname{Att}$}} & \multicolumn{3}{c|}{Krum} & \multicolumn{3}{c|}{FLTrust} & \multicolumn{3}{c|}{FL-Auditor}& \multicolumn{3}{c|}{AlignIns} & \multicolumn{3}{c|}{FLgym}& \multicolumn{3}{c}{FL-OA} \\ 
&& \textcolor{white}{---}IID\textcolor{white}{---}  &DIR(0.6)&DIR(0.1)&\textcolor{white}{---}IID\textcolor{white}{---} &DIR(0.6)&DIR(0.1)&\textcolor{white}{---}IID\textcolor{white}{---}  
&DIR(0.6)&DIR(0.1)&\textcolor{white}{---}IID\textcolor{white}{---}  &DIR(0.6)
&DIR(0.1)&\textcolor{white}{---}IID\textcolor{white}{---}  &DIR(0.6)&DIR(0.1)&\textcolor{white}{---}IID\textcolor{white}{---}  &DIR(0.6)&DIR(0.1)    \\ \cmidrule{1-20}       
\multirow{5}{*}{ \makecell[c]{CIFAR10}} &0\%&80.94&79.58& 74.75&81.51&79.45&73.97&81.74&79.67&73.99&82.86&80.64&76.05&81.20&81.77&78.84&\textbf{85.75}&\textbf{84.55}&\textbf{79.45} \\ 
&10\%&75.74&68.25&50.83&76.14&69.31&51.66&76.96&70.14&54.61&82.29&80.63&75.78&81.44&80.00&\textbf{76.93}&\textbf{85.02}&\textbf{83.15}&76.04 \\ 
&30\%&65.57&48.46&42.05&65.15&50.82&42.70&53.91&55.62&44.89&81.13&80.32&72.32&79.64&77.95&\textbf{74.27}&\textbf{83.64}&\textbf{81.74}&73.35 \\ 
&50\%&46.86&41.44&37.76&49.92&46.25&34.40&44.60&42.64&33.81&70.88&68.06&60.16&66.10&66.84&\textbf{63.46}&\textbf{73.61}&\textbf{71.19}&62.97\\ \cmidrule{2-20} 
\multicolumn{2}{c|}{\shadecell{$\operatorname{Acc}_{50\%-0\%}\downarrow$}}
& \shadecell{\textcolor{ForestGreen}{34.08}}
& \shadecell{\textcolor{ForestGreen}{38.47}}
& \shadecell{\textcolor{ForestGreen}{36.99}}
& \shadecell{\textcolor{ForestGreen}{31.59}}
& \shadecell{\textcolor{ForestGreen}{33.20}}
& \shadecell{\textcolor{ForestGreen}{39.57}}
& \shadecell{\textcolor{ForestGreen}{37.14}}
& \shadecell{\textcolor{ForestGreen}{37.03}}
& \shadecell{\textcolor{ForestGreen}{40.18}}
& \shadecell{\textcolor{ForestGreen}{11.98}}
& \shadecell{\textcolor{ForestGreen}{12.58}}
& \shadecell{\textcolor{ForestGreen}{15.89}}
& \shadecell{\textcolor{ForestGreen}{15.10}}
& \shadecell{\textcolor{ForestGreen}{14.93}}
& \shadecell{\textcolor{ForestGreen}{15.38}}
& \shadecell{\textcolor{ForestGreen}{12.14}}
& \shadecell{\textcolor{ForestGreen}{13.36}}
& \shadecell{\textcolor{ForestGreen}{16.48}}\\
\midrule
\multirow{5}{*}{ \makecell[c]{CIFAR100}} &0\%&38.94&38.76&37.07&38.06&38.58&37.73&38.39&38.93&37.49&41.39&42.33&40.17&43.96&42.64&40.29&\textbf{55.50}&\textbf{54.09}&\textbf{51.61}\\
&10\%&35.24&34.14&32.20&34.24&34.14&28.65&32.15&32.41&29.76&35.87&36.53&35.26&42.14&41.52&40.45&\textbf{55.23}&\textbf{53.74}&\textbf{51.13} \\ 
&30\%&31.40&23.87&19.27&27.62&21.53&14.05&27.46&22.05&14.05&30.78&31.84&32.47&38.11&35.23&33.27&\textbf{52.64}&\textbf{51.06}&\textbf{49.02} \\ 
&50\%&27.16&18.47&13.74&16.16&14.16&11.45&23.15&13.98&12.14&26.07&26.59&26.39&30.15&28.75&25.88&\textbf{42.89}&\textbf{42.40}&\textbf{39.39} \\ \cmidrule{2-20}  
\multicolumn{2}{c|}{\shadecell{$\operatorname{Acc}_{50\%-0\%}\downarrow$}}
& \shadecell{\textcolor{ForestGreen}{11.78}}
& \shadecell{\textcolor{ForestGreen}{20.02}}
& \shadecell{\textcolor{ForestGreen}{23.33}}
& \shadecell{\textcolor{ForestGreen}{21.90}}
& \shadecell{\textcolor{ForestGreen}{24.42}}
& \shadecell{\textcolor{ForestGreen}{26.28}}
& \shadecell{\textcolor{ForestGreen}{15.24}}
& \shadecell{\textcolor{ForestGreen}{24.95}}
& \shadecell{\textcolor{ForestGreen}{25.35}}
& \shadecell{\textcolor{ForestGreen}{15.32}}
& \shadecell{\textcolor{ForestGreen}{15.74}}
& \shadecell{\textcolor{ForestGreen}{13.78}}
& \shadecell{\textcolor{ForestGreen}{13.81}}
& \shadecell{\textcolor{ForestGreen}{13.89}}
& \shadecell{\textcolor{ForestGreen}{14.41}}
& \shadecell{\textcolor{ForestGreen}{12.61}}
& \shadecell{\textcolor{ForestGreen}{11.69}}
& \shadecell{\textcolor{ForestGreen}{12.22}}\\
\midrule
\end{tabular}}
\label{6271001} 
\vspace{-10pt}
\end{table*}

\begin{table*}[]
\centering
\caption{Test accuracy \% (Top-1) comparison between   FL-OA and existing schemes under Neurotoxin Attacks.}
\vspace{-10pt}
    \tabcolsep= 0.065cm
      \renewcommand{\arraystretch}{0.4}
\scalebox{0.8}{
\begin{tabular}{c|c|ccc|ccc|ccc|ccc|ccc|ccc}
\midrule
\multirow{2}{*}{ \makecell[c]{Dataset}} & \multirow{2}{*}{ \makecell[c]{$\operatorname{Att}$}} & \multicolumn{3}{c|}{Krum} & \multicolumn{3}{c|}{FLTrust} & \multicolumn{3}{c|}{FL-Auditor}& \multicolumn{3}{c|}{AlignIns} & \multicolumn{3}{c|}{FLgym}& \multicolumn{3}{c}{FL-OA} \\ 
&&\textcolor{white}{---}IID\textcolor{white}{---} &DIR(0.6)&DIR(0.1)&\textcolor{white}{---}IID\textcolor{white}{---} &DIR(0.6)&DIR(0.1)&\textcolor{white}{---}IID\textcolor{white}{---} 
&DIR(0.6)&DIR(0.1)&\textcolor{white}{---}IID\textcolor{white}{---} &DIR(0.6)
&DIR(0.1)&\textcolor{white}{---}IID\textcolor{white}{---} &DIR(0.6)&DIR(0.1)&\textcolor{white}{---}IID\textcolor{white}{---} &DIR(0.6)&DIR(0.1)    \\ \cmidrule{1-20}     
\multirow{4}{*}{ \makecell[c]{CIFAR10}} &0\%&80.94&79.58&74.75&81.51&79.45&73.97&81.74&79.67&73.99&82.86&80.64&76.05&81.20&81.77&78.64&\textbf{85.75}&\textbf{84.55}&\textbf{79.45} \\ 
&10\%&80.41&79.57&74.91&81.33&79.88&73.91&81.47&79.69&73.16&79.34&78.12&73.20&81.92&80.90&78.98&\textbf{85.62}&\textbf{84.15}&\textbf{79.15}\\ 
&30\%&78.18&76.27&73.14&79.11&76.70&70.40&78.10&75.33&71.90&73.33&72.74&65.59&75.58&73.49&70.45&\textbf{82.41}&\textbf{81.06}&\textbf{77.38} \\ 
&50\%&68.09&64.46&59.79&73.17&61.58&63.98&72.00&65.57&63.13&66.37&65.69&61.73&68.63&68.02&61.50&\textbf{76.30}&\textbf{75.02}&\textbf{72.09}  \\ \cmidrule{2-20}
\multicolumn{2}{c|}{\shadecell{$\operatorname{Acc}_{50\%-0\%}\downarrow$}}
& \shadecell{\textcolor{ForestGreen}{12.85}}
& \shadecell{\textcolor{ForestGreen}{15.12}}
& \shadecell{\textcolor{ForestGreen}{14.96}}
& \shadecell{\textcolor{ForestGreen}{8.34}}
& \shadecell{\textcolor{ForestGreen}{17.87}}
& \shadecell{\textcolor{ForestGreen}{9.99}}
& \shadecell{\textcolor{ForestGreen}{9.74}}
& \shadecell{\textcolor{ForestGreen}{14.10}}
& \shadecell{\textcolor{ForestGreen}{10.86}}
& \shadecell{\textcolor{ForestGreen}{16.49}}
& \shadecell{\textcolor{ForestGreen}{14.95}}
& \shadecell{\textcolor{ForestGreen}{14.32}}
& \shadecell{\textcolor{ForestGreen}{12.57}}
& \shadecell{\textcolor{ForestGreen}{13.75}}
& \shadecell{\textcolor{ForestGreen}{17.14}}
& \shadecell{\textcolor{ForestGreen}{9.45}}
& \shadecell{\textcolor{ForestGreen}{9.53}}
& \shadecell{\textcolor{ForestGreen}{7.36}}\\
\midrule
\multirow{4}{*}{ \makecell[c]{CIFAR100}} &0\%&38.94&38.76&37.07&38.06&38.58&37.73&38.39&38.93&37.49&41.39&42.33&40.17&43.96&42.64&40.29&\textbf{55.50}&\textbf{54.09}&\textbf{51.61}\\
&10\%&38.78&38.31&36.98&38.57&38.60&37.71&38.44&38.96&37.17&36.50&36.91&35.24&43.65&41.60&40.04&\textbf{54.57}&\textbf{53.72}&\textbf{51.93} \\ 
&30\%&31.51&32.34&30.54&35.43&35.43&32.26&36.46&33.41&32.49&33.48&31.87&34.25&33.74&34.59&35.03&\textbf{51.90}&\textbf{51.36}&\textbf{45.30} \\ 
&50\%&25.74&24.01&22.26&26.26&24.46&20.41&28.14&26.01&24.57&25.88&23.29&22.94&30.76&27.98&26.75&\textbf{47.84}&\textbf{44.81}&\textbf{41.24} \\ \cmidrule{2-20}  
\multicolumn{2}{c|}{ \shadecell{$\operatorname{Acc}_{50\%-0\%}\downarrow$}}
& \shadecell{\textcolor{ForestGreen}{13.20}}
& \shadecell{\textcolor{ForestGreen}{14.75}}
& \shadecell{\textcolor{ForestGreen}{14.81}}
& \shadecell{\textcolor{ForestGreen}{11.80}}
& \shadecell{\textcolor{ForestGreen}{14.12}}
& \shadecell{\textcolor{ForestGreen}{17.32}}
& \shadecell{\textcolor{ForestGreen}{10.25}}
& \shadecell{\textcolor{ForestGreen}{12.92}}
& \shadecell{\textcolor{ForestGreen}{12.92}}
& \shadecell{\textcolor{ForestGreen}{15.51}}
& \shadecell{\textcolor{ForestGreen}{19.04}}
& \shadecell{\textcolor{ForestGreen}{17.23}}
& \shadecell{\textcolor{ForestGreen}{13.20}}
& \shadecell{\textcolor{ForestGreen}{14.66}}
& \shadecell{\textcolor{ForestGreen}{13.54}}
& \shadecell{\textcolor{ForestGreen}{7.66}}
& \shadecell{\textcolor{ForestGreen}{9.28}}
& \shadecell{\textcolor{ForestGreen}{10.37}}\\
\midrule
\end{tabular}}
\label{6271002} 
\vspace{-10pt}
\end{table*}

\begin{table*}[]
\centering
\caption{Test accuracy \% (Top-1) comparison between   FL-OA and existing schemes under Focused-Flip Attacks.}
\vspace{-10pt}
    \tabcolsep= 0.065cm
      \renewcommand{\arraystretch}{0.4}
\scalebox{0.8}{
\begin{tabular}{c|c|ccc|ccc|ccc|ccc|ccc|ccc}
\midrule
\multirow{2}{*}{ \makecell[c]{Dataset}} & \multirow{2}{*}{ \makecell[c]{$\operatorname{Att}$}} & \multicolumn{3}{c|}{Krum} & \multicolumn{3}{c|}{FLTrust} & \multicolumn{3}{c|}{FL-Auditor}& \multicolumn{3}{c|}{AlignIns} & \multicolumn{3}{c|}{FLgym}& \multicolumn{3}{c}{FL-OA} \\ 
&&\textcolor{white}{---}IID\textcolor{white}{---} &DIR(0.6)&DIR(0.1)&\textcolor{white}{---}IID\textcolor{white}{---} &DIR(0.6)&DIR(0.1)&\textcolor{white}{---}IID\textcolor{white}{---} 
&DIR(0.6)&DIR(0.1)&\textcolor{white}{---}IID\textcolor{white}{---} &DIR(0.6)
&DIR(0.1)&\textcolor{white}{---}IID\textcolor{white}{---} &DIR(0.6)&DIR(0.1)&\textcolor{white}{---}IID\textcolor{white}{---} &DIR(0.6)&DIR(0.1)    \\ \cmidrule{1-20}  
\multirow{4}{*}{ \makecell[c]{CIFAR10}} &0\%&80.94&79.58&74.75&81.51&79.45&73.97&81.74&79.67&73.99&82.86&80.64&76.05&81.20&81.77&78.64&\textbf{85.75}&\textbf{84.55}&\textbf{79.45} \\ 
&10\%&78.90&77.93&72.68& 79.82& 78.84& 73.14&81.27 &79.05 & 73.45& 79.51& 77.32 &72.64 &80.74&81.22&76.47&\textbf{85.48}&\textbf{84.46}&\textbf{79.29}\\ 
&30\%&71.34&70.48&68.54& 72.96& 72.10& 69.65& 75.44& 72.31& 70.47&74.37  &72.04  &65.55 &76.49&74.25&70.95&\textbf{81.26}&\textbf{80.15}&\textbf{76.27} \\ 
&50\%&63.41&62.57&57.86& 67.50& 65.78& 59.83& 69.20&67.56  &61.50 &66.66 &64.46 & 60.64&68.06&63.94&60.77&\textbf{73.92}&\textbf{72.73}&\textbf{68.04}  \\ \cmidrule{2-20}  
\multicolumn{2}{c|}{\shadecell{$\operatorname{Acc}_{50\%-0\%}\downarrow$}}
& \shadecell{\textcolor{ForestGreen}{17.53}}
& \shadecell{\textcolor{ForestGreen}{17.01}}
& \shadecell{\textcolor{ForestGreen}{16.89}}
& \shadecell{\textcolor{ForestGreen}{14.01}}
& \shadecell{\textcolor{ForestGreen}{13.67}}
& \shadecell{\textcolor{ForestGreen}{14.14}}
& \shadecell{\textcolor{ForestGreen}{12.54}}
& \shadecell{\textcolor{ForestGreen}{12.11}}
& \shadecell{\textcolor{ForestGreen}{12.49}}
& \shadecell{\textcolor{ForestGreen}{16.20}}
& \shadecell{\textcolor{ForestGreen}{16.18}}
& \shadecell{\textcolor{ForestGreen}{15.41}}
& \shadecell{\textcolor{ForestGreen}{13.14}}
& \shadecell{\textcolor{ForestGreen}{17.83}}
& \shadecell{\textcolor{ForestGreen}{17.87}}
& \shadecell{\textcolor{ForestGreen}{11.83}}
& \shadecell{\textcolor{ForestGreen}{11.82}}
& \shadecell{\textcolor{ForestGreen}{11.41}}\\
\midrule
\multirow{4}{*}{ \makecell[c]{CIFAR100}} &0\%&38.94&38.76&37.07&38.06&38.58&37.73&38.39&38.93&37.49&41.39&42.33&40.17&43.96&42.64&40.29&\textbf{55.50}&\textbf{54.09}&\textbf{51.61}\\
&10\%&34.50&34.29&33.11& 37.29& 37.07&32.46 & 37.94& 37.86& 32.98& 40.15& 40.07& 39.46&41.45&41.02&40.20&\textbf{55.36}&\textbf{54.17}&\textbf{51.44} \\ 
&30\%&30.93 &28.44 &25.04 & 28.63& 22.22&28.83 & 29.72 & 28.58&25.57 & 35.46& 35.15&34.38 &36.79&36.62&34.60&\textbf{51.95}&\textbf{50.62}&\textbf{48.05} \\ 
&50\%&25.72 & 21.37&16.66 & 21.80& 20.92 &19.17 & 24.87& 24.49&19.70 & 25.73&24.66 & 23.25&29.17&27.15&25.22&\textbf{45.24}&\textbf{43.05}&\textbf{42.30} \\ \cmidrule{2-20}  
\multicolumn{2}{c|}{\shadecell{$\operatorname{Acc}_{50\%-0\%}\downarrow$}}
& \shadecell{\textcolor{ForestGreen}{13.22}}
& \shadecell{\textcolor{ForestGreen}{17.39}}
& \shadecell{\textcolor{ForestGreen}{20.41}}
& \shadecell{\textcolor{ForestGreen}{16.26}}
& \shadecell{\textcolor{ForestGreen}{17.66}}
& \shadecell{\textcolor{ForestGreen}{18.56}}
& \shadecell{\textcolor{ForestGreen}{13.52}}
& \shadecell{\textcolor{ForestGreen}{14.44}}
& \shadecell{\textcolor{ForestGreen}{17.79}}
& \shadecell{\textcolor{ForestGreen}{15.66}}
& \shadecell{\textcolor{ForestGreen}{17.67}}
& \shadecell{\textcolor{ForestGreen}{16.92}}
& \shadecell{\textcolor{ForestGreen}{14.79}}
& \shadecell{\textcolor{ForestGreen}{15.49}}
& \shadecell{\textcolor{ForestGreen}{15.07}}
& \shadecell{\textcolor{ForestGreen}{10.26}}
& \shadecell{\textcolor{ForestGreen}{11.04}}
& \shadecell{\textcolor{ForestGreen}{9.31}}\\
\midrule
\end{tabular}}
\label{6271003} 
\vspace{-10pt}
\end{table*}

In addition, we compare the performance of FL-OA with that of five existing schemes (i.e., Krum, FLTrust, FL-Auditor, AlignIns, and FLgym).
Specifically, we evaluate their test accuracy  on  CIFAR10 and CIFAR100  when subjected to Gaussian, Neurotoxin, and Focused-Flip attacks across different data distributions.
 The proportion of malicious devices is set to 0\%, 10\%, 30\%, and 50\%, respectively.
The   results are reported in Tables \ref{6271001} to \ref{6271003}.
It can be observed that   FL-OA achieves the highest test accuracy in the absence of malicious devices.
 This indicates that the gradient ascent step and correction term introduced by FL-OA enhance the consistency of model updates and further improve the performance of the global model.
 As the proportion of malicious devices increases, the accuracy of FL-OA gradually declines, which is a reasonable phenomenon.
This is because fewer benign samples remain available for effective model training.
When the proportion of malicious devices reaches 50\%, the performance of Krum degrades  significantly. 
This is because Krum is a statistical knowledge-based approach whose efficacy   relies on the assumption that benign devices are in the majority.
Once this assumption no longer holds, the updates selected by Krum may derive from malicious devices, leading to a significant decline in global model performance.
Furthermore, under the DIR(0.1) setting, the test accuracy of both FLTrust and FL-Auditor also drops noticeably.
This is because both schemes are  root dataset-based approach, the data distribution of the root dataset differs  from that of the device's local dataset, which weakens their ability to distinguish malicious updates.
Moreover,   AlignIns and FLgym  achieve relatively high test accuracy on both Neurotoxin and focused-flip attacks.
This is because AlignIns adopts a multi-granularity evaluation that offers strong robustness, while FLgym integrates a sliding window-based weight recovery mechanism, allowing some benign updates that are mistakenly filtered out to still participate in aggregation.
Overall, FL-OA  outperforms the other five schemes across different data distributions and attack settings.
However, since FL-OA improves the local training process under Non-IID data, the test accuracy of the model cannot be directly used to compare its defense capability with that of the baseline schemes.

To provide a more intuitive evaluation of robustness,  we calculate the difference in test accuracy for each scheme when the proportion of malicious devices is 50\% and 0\%, denoted as $\operatorname{Acc}_{50\%-0\%}$.
A smaller value of $\operatorname{Acc}$ indicates stronger robustness.
As highlighted in green in Tables \ref{6271001} to \ref{6271003}, FL-OA generally achieves lower $\operatorname{Acc}$ values than the baseline methods, demonstrating its superior robustness against Byzantine attacks. 
In particular, under the DIR(0.1) setting, Krum shows  large $\operatorname{Acc}$ values across all attack types, because severe inconsistency among benign model updates weakens its defense effectiveness.
Furthermore, although the performance degradation of AlignIns and FLgym under Non-IID settings is similar to that under the IID setting,  their overall performance decline remains significant.

\subsubsection{Ablation Study}

\begin{table}[t]
\caption{Ablation study of FL-OA with   Gaussian and neurotoxin attacks on CIFAR100.}
\vspace{-10pt}
\tabcolsep= 0.07cm
\centering
\scalebox{0.9}{
\begin{tabular}{c|c|ccc|ccc}
\midrule\midrule
 \multicolumn{2}{c}{\multirow{2}{*}{Methods}}  &\multicolumn{3}{c}{Gaussian ($\operatorname{Att} = 50\%$)} &\multicolumn{3}{c}{Neurotoxin ($\operatorname{Att} = 50\%$)}  \\
\multicolumn{2}{c}{}&\textcolor{white}{---}IID\textcolor{white}{---}   &DIR(0.6) &DIR(0.1)&\textcolor{white}{---}IID\textcolor{white}{---}   &DIR(0.6) &DIR(0.1) \\\midrule
\multirow{6}{*}{\rotatebox{90}{RseNet18}}& \multirow{1}{*}{w/o  GAS} &  40.85 & 37.99 & 34.47 & 43.10 & 40.56 & 37.24 \\
& \multirow{1}{*}{w/o CT} & 40.57 & 40.36 & 38.38 & 42.50 & 42.25 & 40.68 \\
& \multirow{1}{*}{(w/o GAS \& CT)} & 33.34 &  32.20 & 29.27 & 35.29 &33.45& 31.77  \\
 & \multirow{1}{*}{FL-OA ($\kappa= 100\%$)} & 41.62 & 38.18  & 35.64 & 46.57 & 42.05 & 38.27 \\
 &   \multirow{1}{*}{FL-OA ($\kappa= 70\%$)} & 42.28 & 40.80  & 37.45 & \textbf{47.84} & 43.06 & 40.56 \\
 &       \multirow{1}{*}{FL-OA ($\kappa= 30\%$)} & \textbf{42.98}& \textbf{42.40} & \textbf{39.39} &47.82 & \textbf{44.81} & \textbf{41.24} \\\midrule
\multirow{6}{*}{\rotatebox{90}{ResNet50}} 
& \multirow{1}{*}{w/o GAS} 
& 42.28 & 42.19 & 38.52 & 45.92 & 42.34 & 40.25 \\

& \multirow{1}{*}{w/o CT} 
& 40.49 & 38.54 & 37.99 & 40.44 & 39.29 & 36.95 \\

& \multirow{1}{*}{(w/o GAS \& CT)} 
& 35.86 & 35.06 & 33.43 & 35.06 & 35.87 & 33.23 \\

& \multirow{1}{*}{FL-OA ($\kappa=100\%$)} 
& 44.84 & 43.60 & 41.49 & 46.82 & 45.63 & 43.94 \\

& \multirow{1}{*}{FL-OA ($\kappa=70\%$)} 
& 46.10   &  \textbf{44.76} & 42.45  & 47.59 &  \textbf{46.26} &  \textbf{44.55} \\

& \multirow{1}{*}{FL-OA ($\kappa=30\%$)} 
&  \textbf{46.31} & 45.58 &   \textbf{43.92}  &  \textbf{47.85} & 45.70 & 43.02 \\
\midrule
\midrule
\end{tabular}}
\justifying
Note: ``w/o GAS''  and ``w/o CT'' denote FL-OA without the gradient ascent step and without the correction term in local training, respectively.
FL-OA ($\kappa$) indicates that    the top-$\kappa\%$   parameters are selected  for auditing.
\label{3159191351}
\vspace{-5pt}
\end{table}

To further evaluate the contribution of each component in FL-OA, we conduct ablation studies on  CIFAR100 using ResNet18 and ResNet50  under two attack scenarios, where 50\% of the devices launch Gaussian attacks and Neurotoxin attacks, respectively. 
The results are reported in Table \ref{3159191351}.
(\textit{i}) \textbf{Ablation on local training.} 
Specifically, we remove the gradient ascent step (GAS) and the correction term (CT) from the local training  of FL-OA to evaluate their effects on model performance. 
The results show that ``w/o GAS \& CT'' achieves the worst performance. 
This is because, without the gradient ascent step and the correction term, local training cannot enhance the consistency among benign updates, thereby weakening the defense capability of FL-OA. 
In contrast, when GAS and CT are introduced separately, FL-OA consistently achieves better performance, indicating that each of them is beneficial to the robustness of FL-OA.
(\textit{ii}) \textbf{Ablation on critical parameters.} To evaluate the role of critical parameters, we  compare the performance of FL-OA when using all parameters ($\kappa = 100\%$) and when using only critical parameters ($\kappa = 70\%$ and $\kappa = 30\%$). The results show that FL-OA performs better when only the  70\% or  30\% critical parameters are used than when all parameters are used. This indicates that the  parameter importance indicator $\operatorname{PII}$ can  identify the critical parameters in  updates, thereby improving the effectiveness of FL-OA in defending against malicious updates.
Moreover, the ablation studies conducted with ResNet18 and ResNet50 yield consistent   conclusions with those presented above, indicating that the proposed components generalize well across different model architectures.
 
\subsubsection{Performance of FL-OA under High Percentages of  Malicious Devices}

In FL-OA, the root model update $\textit{\textbf{u}}_{\textit{root}}^{t}$ is employed as a reference for evaluating each model updates submitted by each device.
Thus, different from defense methods that rely on the assumption that benign devices constitute the majority,  the effectiveness of FL-OA is not constrained by the percentage of malicious devices.
To evaluate the performance of FL-OA under high percentage of malicious devices, we conduct experiments under   focused-flip attacks by setting the   percentage of malicious devices   to 50\%, 60\%, and 70\%, respectively, using a setting without any malicious devices (0\%) as a reference.
In addition, we construct a version of FL-OA that aggregates model updates using only the standard averaging method. 
For simplicity, this variant is  referred to as ``Baseline.''
As shown in Table \ref{0121021},   FL-OA exhibits   stronger defensive capability than Baseline when the percentage of malicious devices is high.
For example, under the IID setting on CIFAR10, when the percentage of malicious devices reaches 50\%, the model accuracy  of FL-OA decreases by only approximately 12\% relative to the attack-free setting, whereas that of Baseline decreases by approximately 31\%. 
When the percentage further increases to 70\%, the accuracy of FL-OA decreases by approximately 19\%, while that of Baseline decreases by approximately 33\%. 
These results demonstrate that although increasing the percentage of malicious devices degrades the model accuracy of FL-OA, it remains effective in defending against malicious devices even when malicious devices constitute the majority.
The accuracy degradation of FL-OA is attributed to the decrease in the number of benign samples available for model training as the percentages of malicious devices increases.

\begin{table}[t]
\centering
\caption{Performance of FL-OA under high Percentages of malicious devices.}
\scalebox{0.8}{
\renewcommand{\arraystretch}{0.8}
\tabcolsep=0.17cm
\begin{tabular}{cc|ccc|ccc}
\midrule\midrule
\multirow{3}{*}{\textbf{Dataset}}
& \multirow{3}{*}{$\operatorname{Att}$}
& \multicolumn{6}{c}{\textbf{Test Accuracy (\%)}}
  \\

& & \multicolumn{3}{c|}{Baseline}
  & \multicolumn{3}{c}{FL-OA}\\

& & IID
  & DIR(0.6)
  & DIR(0.1)
& IID
  & DIR(0.6)
  & DIR(0.1) \\
\cmidrule{1-8}

\multirow{4}{*}{CIFAR10}
& 0\% & 85.42 & 85.77 & 79.58 & 85.75 & 84.57 & 79.45   \\
& 50\% & 54.23 & 61.33 & 55.43 & 73.92 & 72.73 & 68.04 \\
& 60\% &47.10 & 42.54 & 36.40 & 71.18 & 71.44 & 65.67  \\
& 70\% &19.68 & 23.82 & 15.74 & 66.74 & 64.23 & 61.03  \\
\cmidrule{1-8}

\multirow{4}{*}{CIFAR100}
& 0\% & 55.11 & 54.80 & 52.09 & 55.50 & 54.09 & 51.61  \\
& 50\% & 34.72 & 34.32 & 30.22 & 45.36 & 43.05 & 42.30   \\
& 60\% & 12.07 & 14.53 & 12.25 & 37.76 & 36.48 & 28.71 \\
& 70\% &7.74 & 8.74 & 7.99 & 26.74 & 24.94 & 19.52 \\
\midrule\midrule
\end{tabular}}
\label{0121021}
\end{table}

\subsubsection{Visualization of the Distribution of Critical Parameters}
\begin{figure}[t]
  \centering
  
  \subfloat[]{
      \label{112416531}\includegraphics[width=0.3\linewidth]{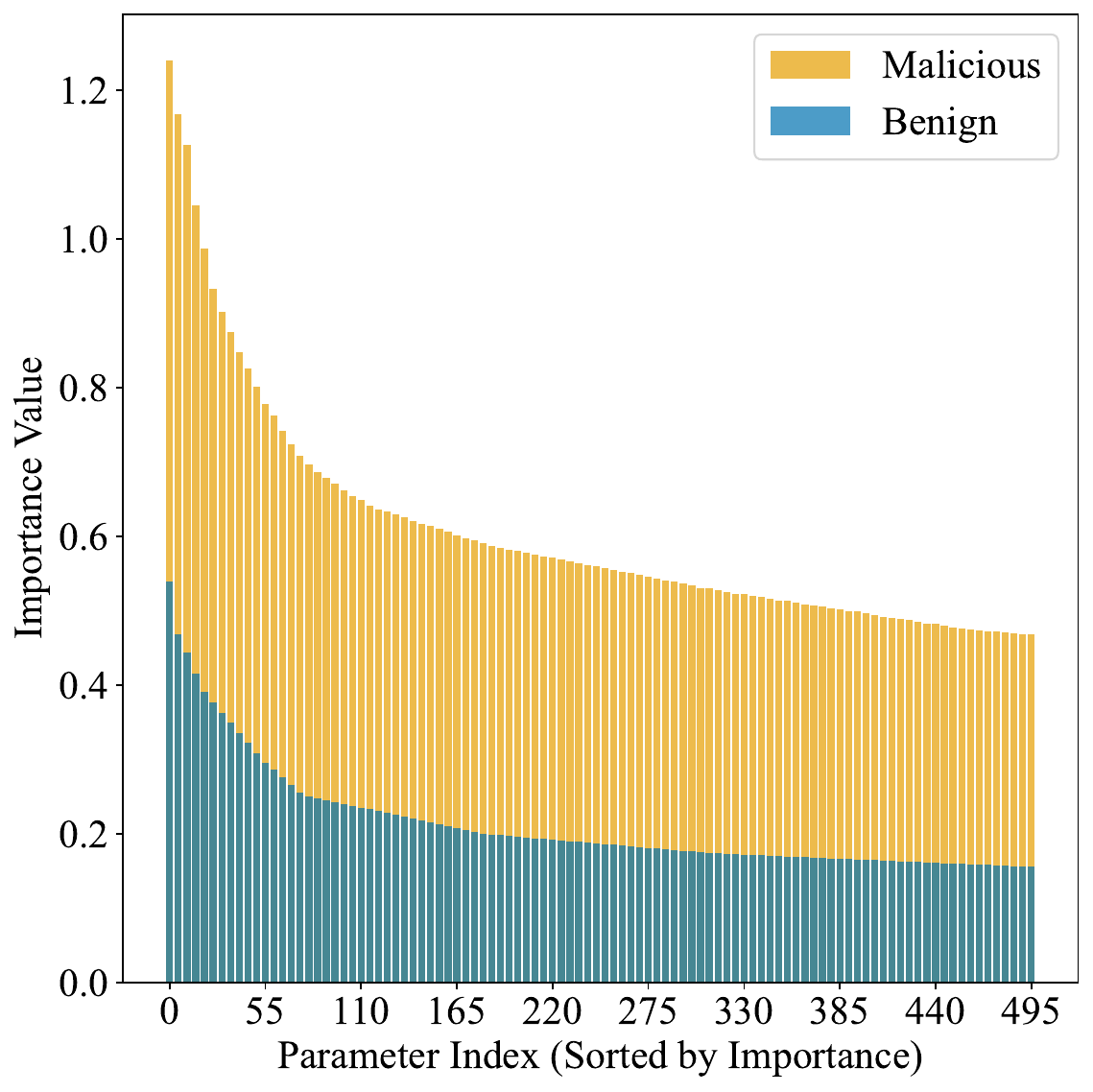}
  }
  \subfloat[]{
      \label{112416532}\includegraphics[width=0.3\linewidth]{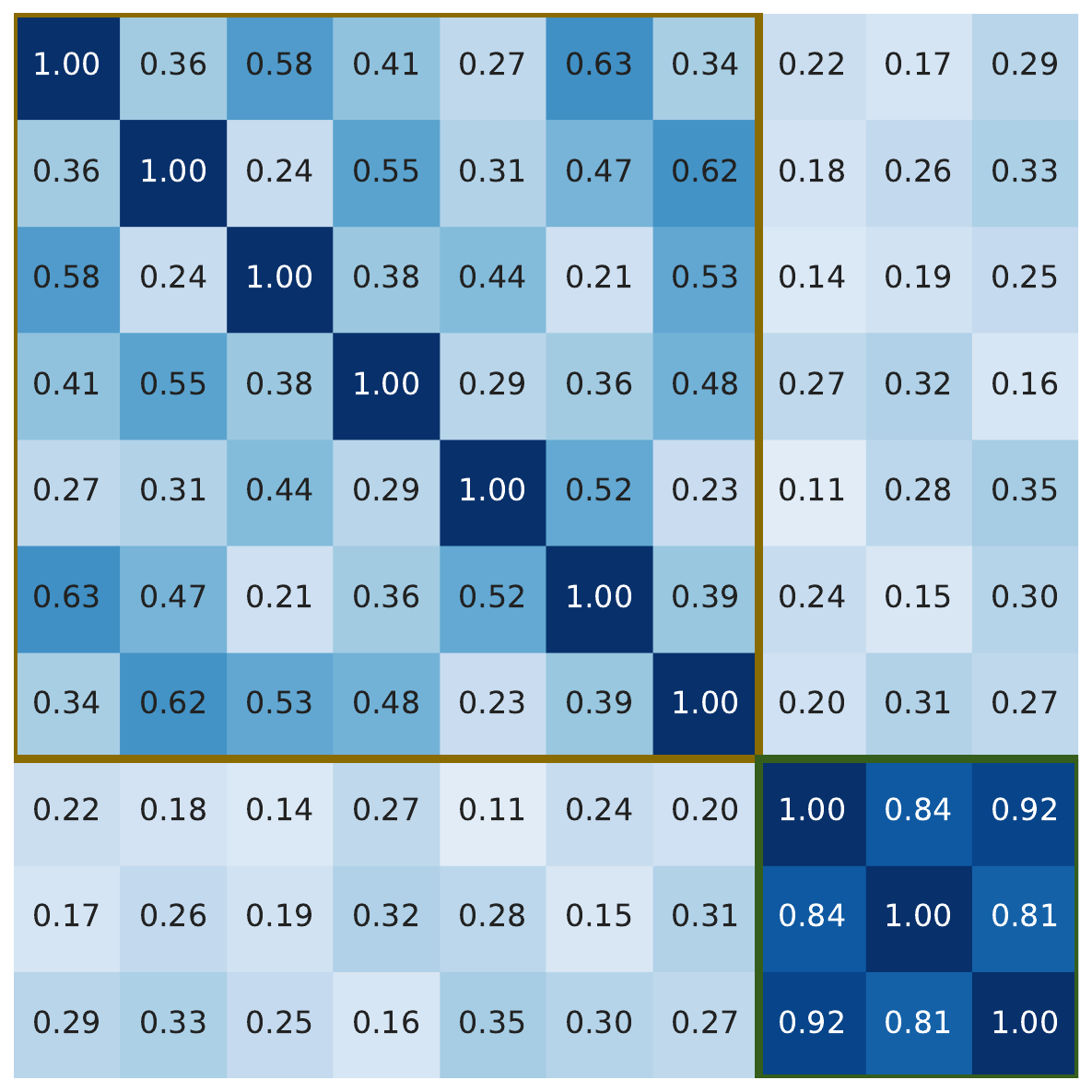}
  }
    \subfloat[]{
      \label{112416533}\includegraphics[width=0.3\linewidth]{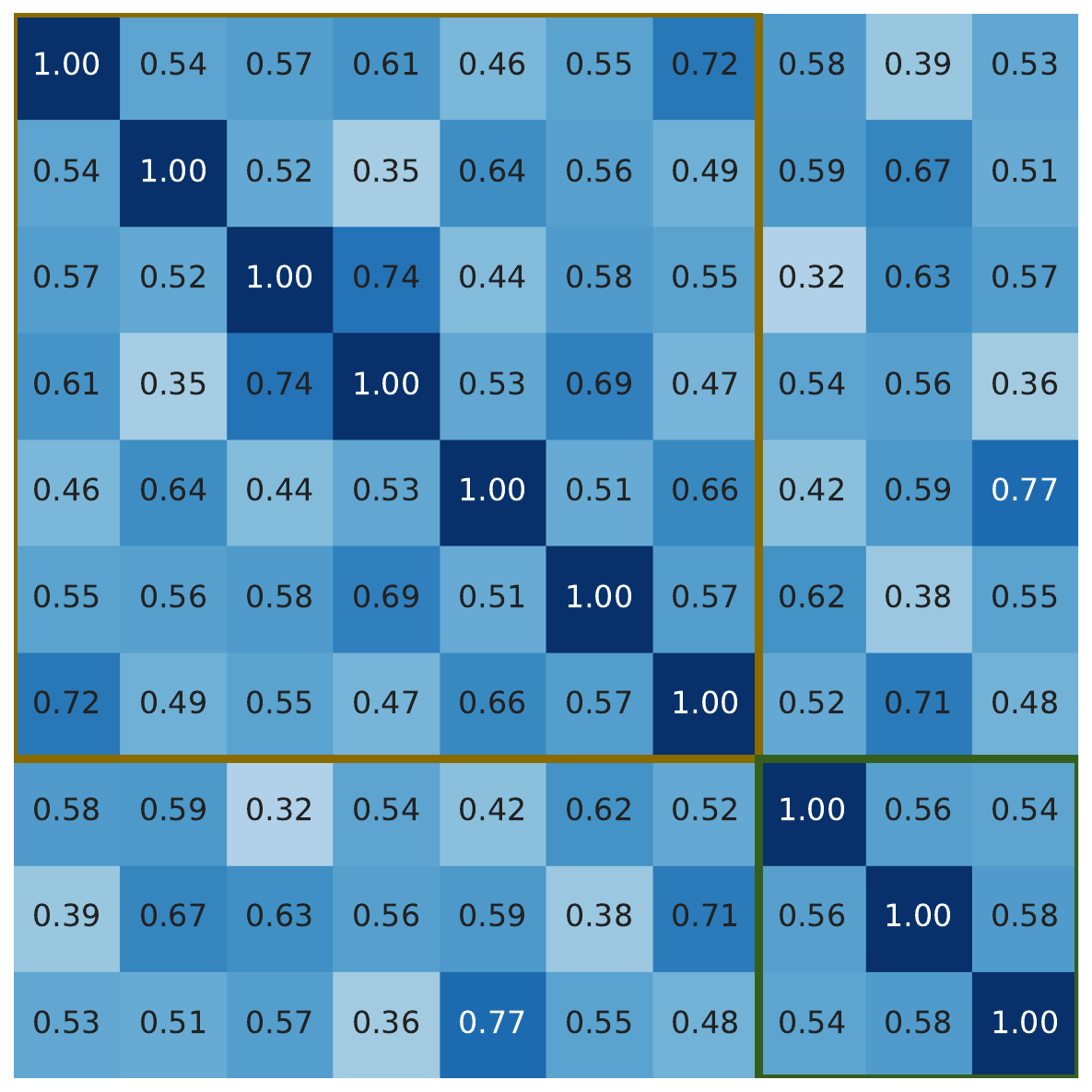}
  }
\caption{(a) The distribution of parameter importance values. (b) Pairwise similarity of critical parameters in model updates between benign and malicious groups. (c) Pairwise similarity of all parameters in model updates between benign and malicious groups.}
\label{11241653}
\end{figure}

To further validate the rationality of critical parameter extraction and the effectiveness of the Parameter Importance Indicator, we conduct experiments on CIFAR100 for analysis, involving three malicious devices and seven benign devices.
 Specifically, we visualize the top 500 parameters in benign and malicious updates sorted  by importance, as shown in Fig. \ref{112416531}. 
 It can be observed that the  importance values of malicious updates are  higher than those of benign updates. 
This is because malicious updates tend to exhibit larger magnitudes on certain parameters and deviate more significantly from the majority of model updates, thereby resulting in higher $\operatorname{PII}$ values.
In addition, we compute the cosine similarity between benign and malicious updates based on the critical parameters, as well as that based on all parameters.
Fig. \ref{112416532} shows that the similarity based on the critical parameters differs significantly between benign and malicious updates. In contrast, Fig. \ref{112416533} shows that the similarity based on all parameters fails to reveal a clear distinction between the two.
These results indicate that critical parameter extraction can alleviate the curse of dimensionality and  enhance the separability between benign and malicious updates.

\subsubsection{The Impact of  $\alpha$ and $\beta$ on FL-OA}
\begin{figure}[!t]
  \centering
  \subfloat[Gaussian Attacks]
  {
      \label{2063141}  \includegraphics[width=0.3\linewidth]{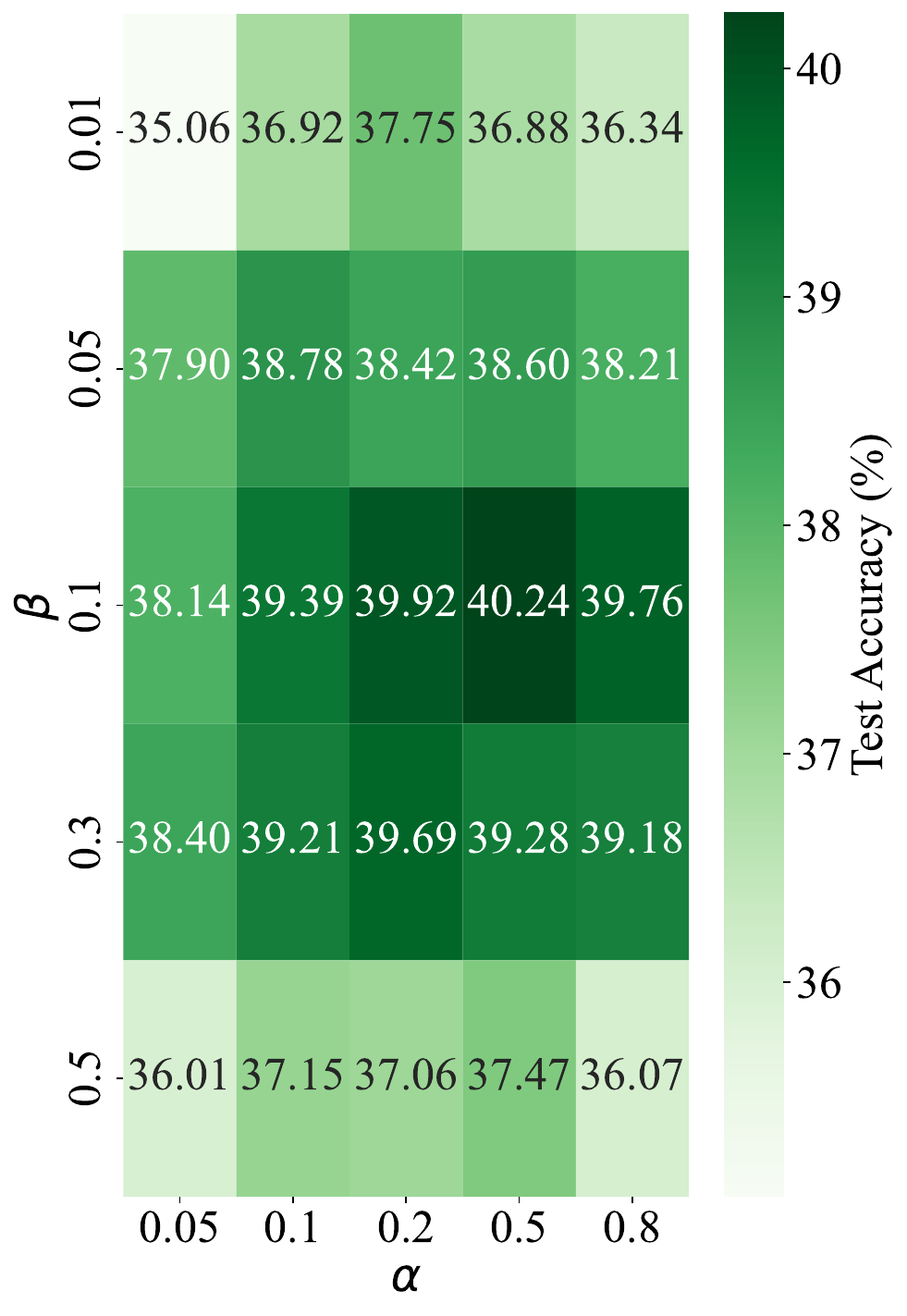}
  }
  \subfloat[Neurotoxin Attacks]
  {
      \label{20631412}  \includegraphics[width=0.3\linewidth]{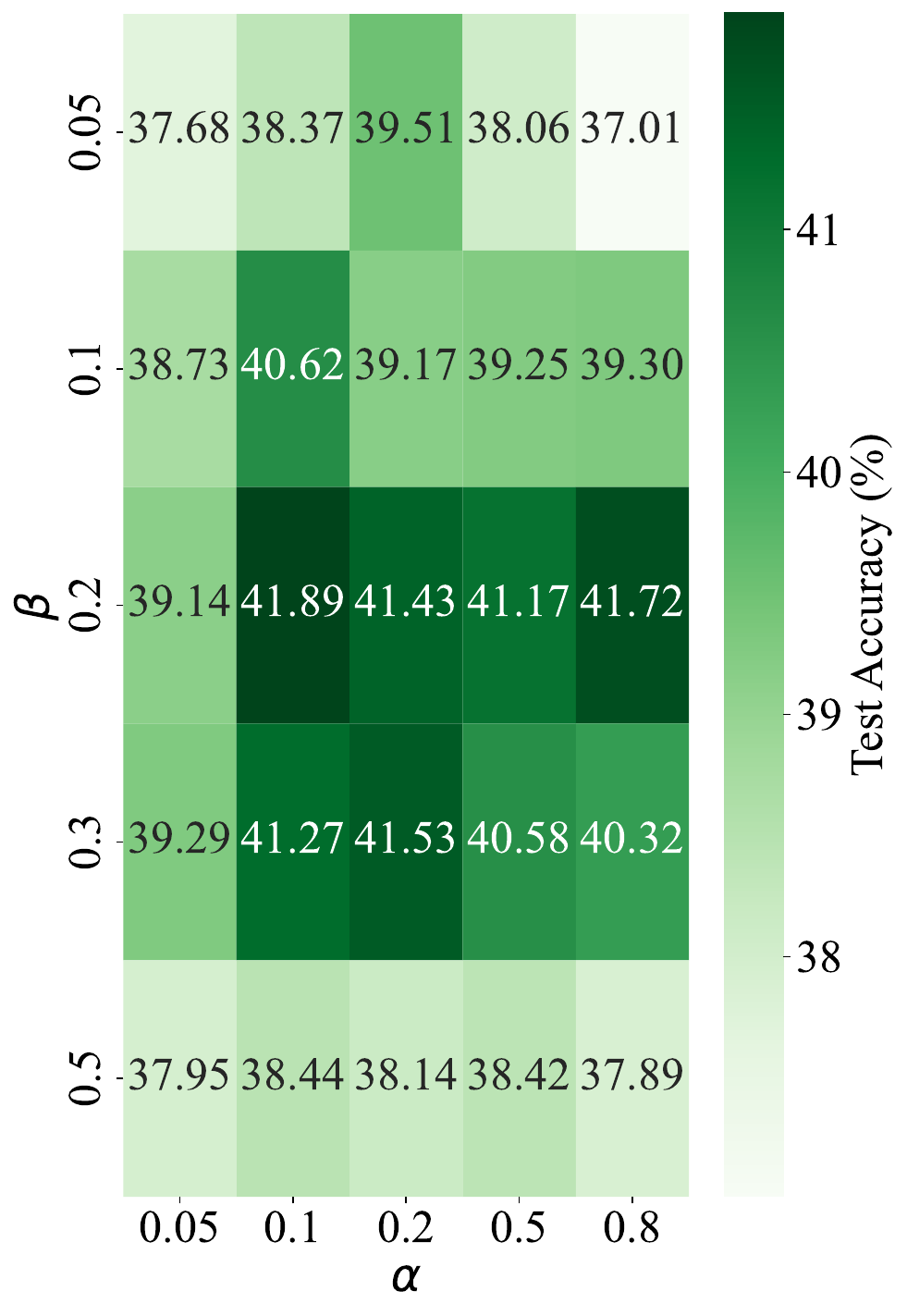}
  }
    \subfloat[Focused-Flip  Attacks]
  {
      \label{20631413}  \includegraphics[width=0.3\linewidth]{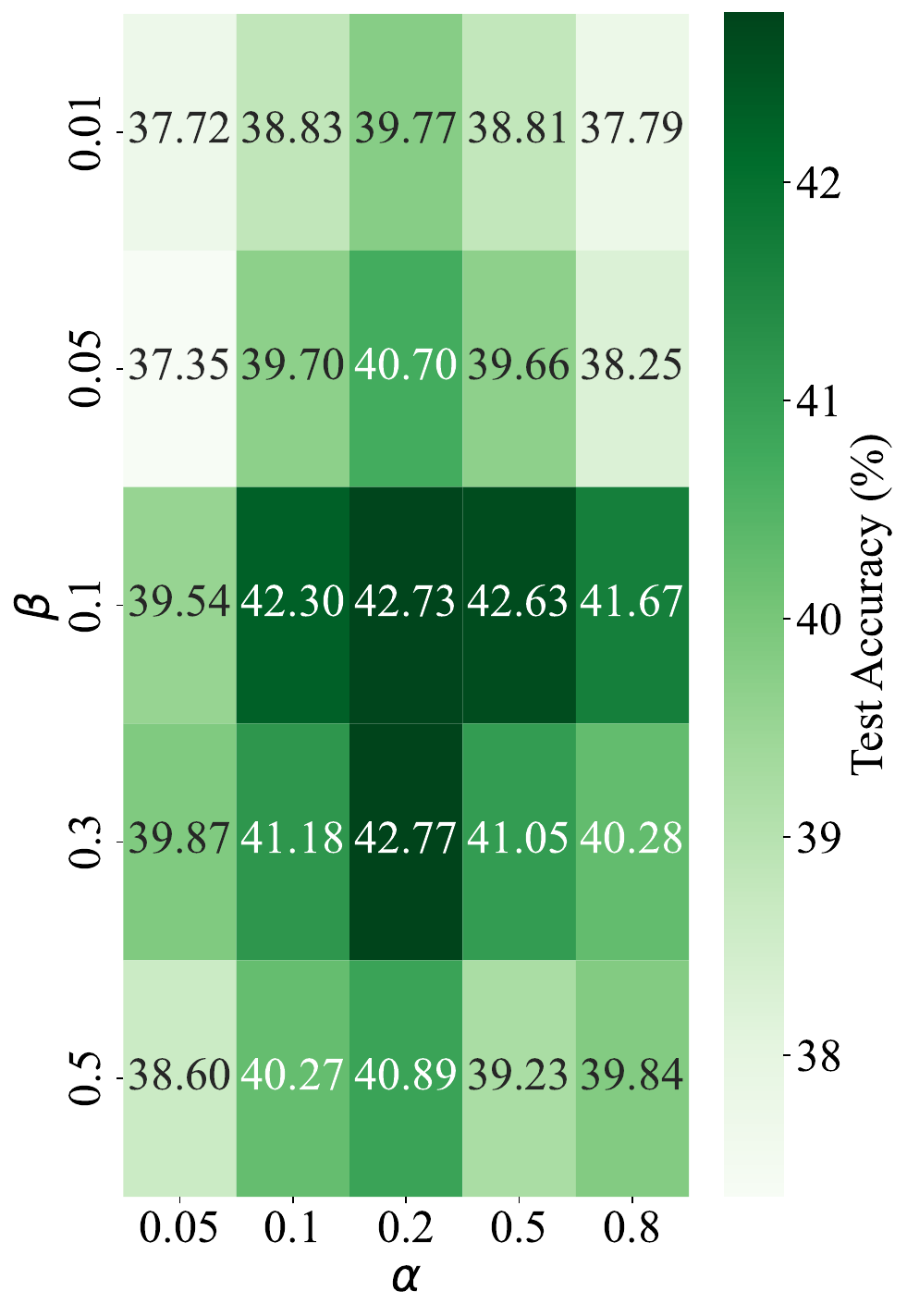}
  }
\caption{Test accuracy (\%) of FL-OA under   the coefficients $\alpha$ and $\beta$.}
\label{2063141300}
\vspace{-5pt}
\end{figure}

FL-OA introduces a gradient ascent step and a correction term into local training.
The effect of the correction term is controlled by the coefficient $\alpha$, while that of the gradient ascent step is scaled by $\beta$. 
Thus, it is necessary to analyze the sensitivity of FL-OA to the coefficients $\alpha$ and $\beta$.
To this end, we evaluate the performance of FL-OA under three attack scenarios, including Gaussian attacks, Neurotoxin attacks, and Focused-Flip attacks, where 50\% of the devices are malicious.
The experiments are conducted on  CIFAR100  under the DIR(0.1) data distribution. 
Specifically, the  coefficient $\alpha$ is selected from $\{0.05, 0.1, 0.2, 0.5, 0.8\}$, and $\beta$ is selected from $\{0.01, 0.05, 0.1, 0.3, 0.5\}$. 
All other experimental settings are kept at their default values.
As shown in Fig. \ref{2063141300}, the results indicate that when $\alpha= 0.8$, the effect of the correction term is excessively amplified, leading to a decline in the test accuracy of FL-OA. 
This is because an overly  large $\alpha$ emphasizes the consistency between the local model and the global model,  weakening the classification capability of the  model. 
This observation suggests that, defense schemes should not focus solely on the consistency among benign updates while neglecting the classification performance of the model itself when defending against Byzantine attacks.
 Notably, the performance of FL-OA varies only slightly across different values of  $\alpha$, indicating that FL-OA is relatively insensitive to $\alpha$.
Furthermore, we observe that when   $\beta$ is too small, local training tends to overemphasize the gradient ascent step, making the local model more prone to under-fitting and causing a noticeable drop in accuracy.
Overall, FL-OA is  sensitive to smaller values of $\beta$. 
Therefore, $\beta$ should be set to a relatively larger value to enhance the consistency among model updates while maintaining training convergence.

\subsubsection{Performance of FL-OA on Two-Class Non-IID Data}

In our default experimental setup, we employ the Dirichlet distribution for data partitioning, which ensures overlapping classes between devices and the OS.
 However, in real-world scenarios, a significant distribution shift may exist between the root dataset and the devices' data. 
In extreme cases, their class sets may be completely disjoint.
 To evaluate the robustness of FL-OA under such extreme  distribution shift, we conducted experiments on two-class Non-IID Data \cite{zhao2018federated}.
 Specifically, we sort  the data by class to create highly heterogeneous distributions on CIFAR10.
The train  set is evenly divided among the 100 devices and OS. 
We further design two root dataset configurations, denoted as Root2  and Root10, respectively.
\begin{itemize}
  \item Root2: The sorted data is divided into 202 partitions. Each device and OS are randomly assigned two partitions from two different classes.

  \item  Root10: The sorted data is partitioned into 210 partitions. Each device receives 2 partitions from 2 different classes, while the OS allocates 10 partitions covering all 10 classes, thereby providing the server with a more comprehensive view.
\end{itemize}

We conduct  experiments under the Root2 and Root10 settings, considering both Gaussian attacks and Neurotoxin attacks, with the proportion of malicious devices  $\operatorname{Att}$  $\in \{0,10\%,20\%,30\%\}$.
For comparison, we  performed experiments with 100 devices under the DIR(0.1)  distribution. 
The experiment results are reported in Table \ref{6271002123}.
The results show that the performance of FL-OA declines significantly under the two-class Non-IID  setting.
Even in the absence of  attacks, the global model accuracy on CIFAR10 dropped below 60\%, indicating that the correction term is ineffective when the participant class distributions are completely disjoint.
Furthermore, Gaussian and Neurotoxin attacks further exacerbate the performance degradation of FL-OA in both Root2 and Root10 settings, highlighting the limitations of our framework in such  distribution shift.
In contrast, FL-OA maintains stable performance under the Dirichlet distribution with DIR(0.1).
These results indicate that our framework is  well-suited for scenarios where  data classes among devices overlap significantly.

\begin{table}[t]
\centering
\caption{Test Accuracy (Top-1) of   FL-OA on two-class Non-IID data and DIR(0.1) under CIFAR10.}
\tabcolsep= 0cm
\scalebox{0.7}{
\begin{tabular}{c|c|c|c|c|c|c|c|c|c|c|c|c}
\midrule\midrule
\multirow{1}{*}{ \makecell[c]{Root dataset}} & \multicolumn{4}{c|}{Root2}& \multicolumn{4}{c|}{Root10} & \multicolumn{4}{c}{DIR(0.1)}\\ \midrule
 $\operatorname{Att}$ &0\%&10\%&20\%&30\%&0\%&10\%&20\%&30\%&0\%&10\%&20\%&30\%\\\midrule
Gaussian attacks &\multirow{2}{*}{53.17\%}&50.04\%&41.85\%&26.27\%&\multirow{2}{*}{58.64\%}&58.19\%&52.93\%&47.44\%&\multirow{2}{*}{79.45\%}&76.04\%&75.46\%&73.35\%\\
Neurotoxin attacks & &51.60\%&48.19\%&43.33\%& &58.36\%&58.12\%&57.65\%& &79.15\%&78.69\%&77.38\%\\
\midrule\midrule
\end{tabular}}
\label{6271002123} 
\end{table}

\subsubsection{Effectiveness on More Datasets}
To further validate the effectiveness of FL-OA on more datasets, we conduct experiments on SVHN \cite{netzer2011reading}, CINIC \cite{darlow2018cinic}, and Tiny-ImageNet \cite{le2015tiny},  using the ResNet18 architecture. 
In these experiments, 50\% of the devices are set to be malicious and launch Gaussian attacks, while  other hyper-parameters are kept at  default settings.
We  compare the test accuracy of FL-OA with that of existing schemes, as reported in Table \ref{3271258}. 
The results show that FL-OA achieves the highest test accuracy under both IID and DIR(0.1) settings. For example, under the DIR(0.1) setting, FL-OA attains the test accuracies of 72.22\%, 43.43\%, and 22.47\% on the SVHN, CINIC, and Tiny-ImageNet datasets, respectively.
In addition, we compare FL-OA with existing schemes in terms of the $\operatorname{Acc}_{50\%-0\%}$ metric. 
The results show that  FL-OA demonstrates superior defensive performance   compared to existing schemes.
Specifically, under the DIR(0.1) setting on the three datasets, FL-OA achieves the lowest $\operatorname{Acc}_{50\%-0\%}$ values of 8.62, 8.82, and  3.16, respectively. 
Compared with AlignIns, these $\operatorname{Acc}$ values are reduced by 21.42\%, 0.03\%, and 57.62\%, respectively. 
These results further verify the robustness of FL-OA across different datasets.

\begin{table}[t]
\caption{Test accuracy (\% $\uparrow$) of FL-OA and existing schemes on more datasets.}
\vspace{-10pt}
\tabcolsep= 0.18cm
\centering
\scalebox{0.9}{
\begin{tabular}{c|cc|cc|cc}
\hline\hline
 \multirow{2}{*}{Methods}  &\multicolumn{2}{c}{SVHN} &\multicolumn{2}{c}{CINIC}&\multicolumn{2}{c}{Tiny-ImageNet}  \\
&\textcolor{white}{---}IID\textcolor{white}{---}   &DIR(0.1) &\textcolor{white}{---}IID\textcolor{white}{---}   &DIR(0.1) &\textcolor{white}{---}IID\textcolor{white}{---}   &DIR(0.1)  \\\midrule
 \multirow{1}{*}{Krum} &  57.56 & 51.05 & 36.83 & 30.38 & 18.84 & 15.08 \\
 \multirow{1}{*}{FLTrust} & 62.50 & 58.31 & 41.43 & 32.97 & 17.12 & 15.44 \\
 \multirow{1}{*}{FL-Auditor} & 59.88 &  56.75 & 43.62 & 38.52 &17.17& 14.10  \\
  \multirow{1}{*}{AlignIns} &75.13 &62.89  & 42.98 & 40.43& 20.30 & 18.84 \\
    \multirow{1}{*}{FLgym} & 75.54 & 65.48  & 44.27 &39.68 & 23.63& 20.29\\
        \multirow{1}{*}{FL-OA} & \textbf{78.41}& \textbf{72.22} & \textbf{48.72} &\textbf{43.43} & \textbf{26.11} & \textbf{22.47} \\\hline 
\midrule
\end{tabular}}
\justifying
\label{3271258}
\vspace{-5pt}
\end{table}

\begin{table}[t]
\caption{$\operatorname{Acc}_{50\%-0\%}$ (\% $\downarrow$) of FL-OA and existing schemes on more datasets.}
\vspace{-10pt}
\tabcolsep= 0.18cm
\centering
\scalebox{0.9}{
\begin{tabular}{c|cc|cc|cc}
\hline\hline
 \multirow{2}{*}{Methods}  &\multicolumn{2}{c}{SVHN} &\multicolumn{2}{c}{CINIC}&\multicolumn{2}{c}{Tiny-ImageNet}  \\
&\textcolor{white}{---}IID\textcolor{white}{---}   &DIR(0.1) &\textcolor{white}{---}IID\textcolor{white}{---}   &DIR(0.1) &\textcolor{white}{---}IID\textcolor{white}{---}   &DIR(0.1)  \\\midrule
 \multirow{1}{*}{Krum} & 24.06 & 26.20 &12.20 & 14.38 & 7.62 & 8.89 \\
 \multirow{1}{*}{FLTrust} & 18.95& 20.52 & 9.25 & 12.78 & 6.13 & 8.21 \\
 \multirow{1}{*}{FL-Auditor} & 21.67 &  18.11& 8.48& 8.97 &7.97& 8.59  \\
  \multirow{1}{*}{AlignIns} &7.40 &12.81 & 8.20 & 7.22&5.81 & 4.25 \\
    \multirow{1}{*}{FLgym} & 8.29 & 1.35  & 6.24 &5.70& 4.86& 5.04\\
        \multirow{1}{*}{FL-OA} & \textbf{5.35}& \textbf{8.62} & \textbf{5.86} &\textbf{8.82} & \textbf{2.11} & \textbf{3.16} \\\hline 
\midrule
\end{tabular}}
\justifying
\label{3271258}
\vspace{-5pt}
\end{table}

\subsubsection{Scalability  of FL-OA}
\begin{figure}[!t]
  \centering   
  \subfloat[CIFAR10.] 
  {
      \label{3116062}  \includegraphics[width=0.45\linewidth]{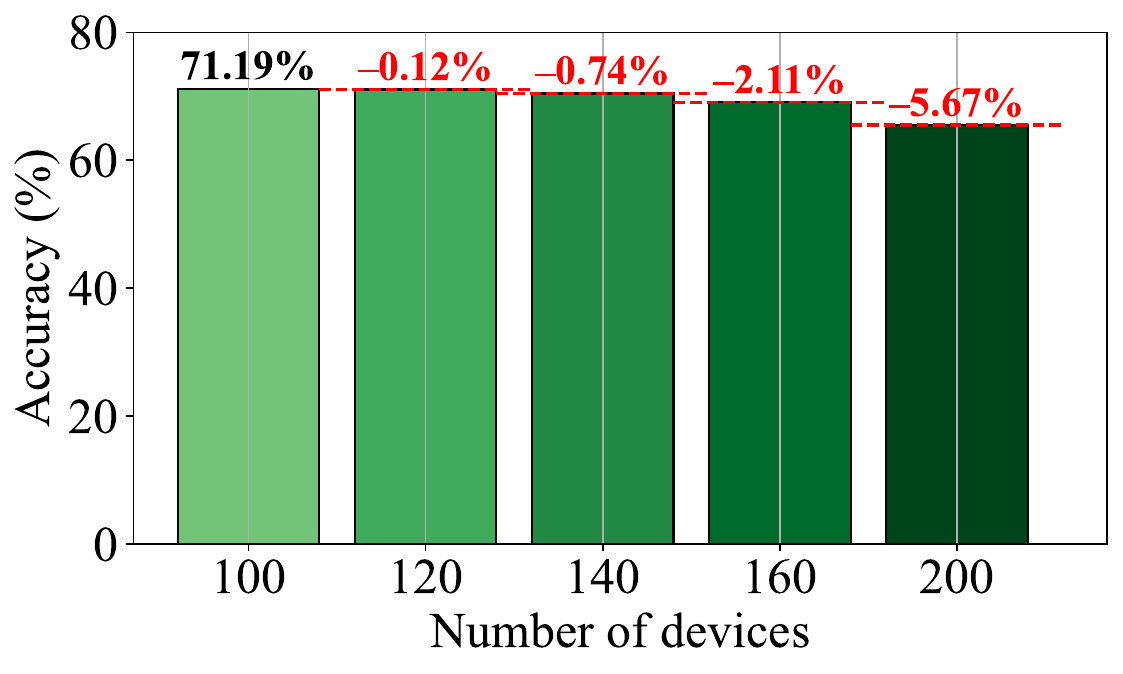}
  }
  \subfloat[CIFAR100.] 
  {
      \label{3116063}  \includegraphics[width=0.45\linewidth]{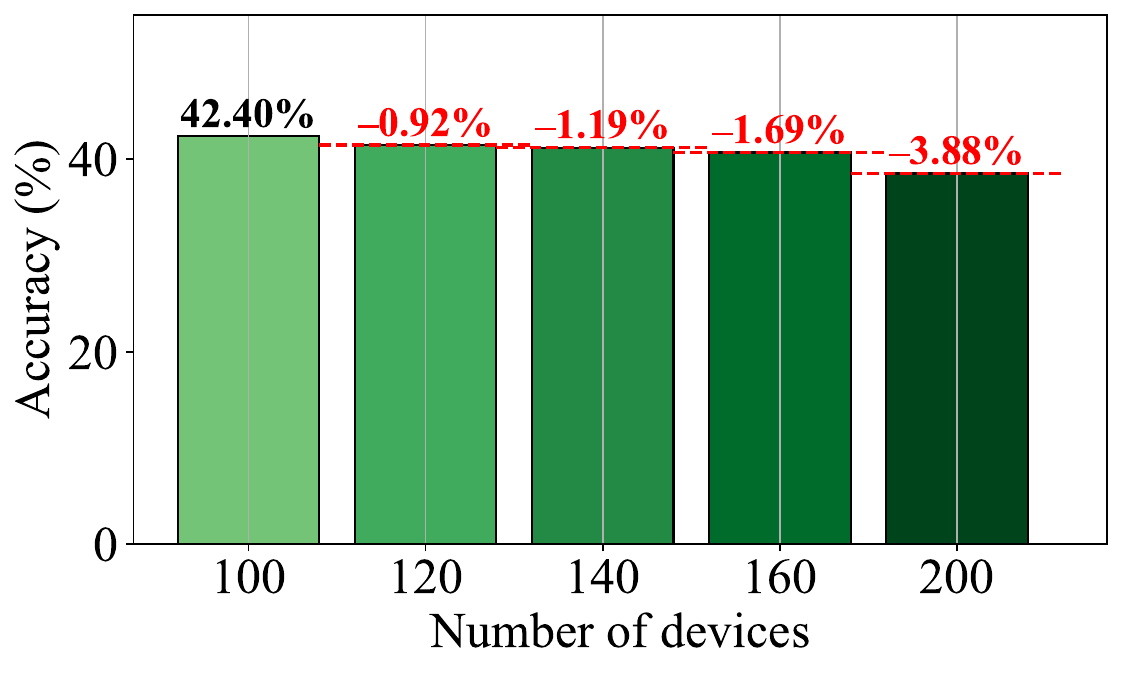}
  }

\caption{
Scalability of FL-OA against  Gaussian   attacks over CIFAR10 and CIFAR100 with data distribution DIR(0.1).}
\vspace{-10pt}
\label{311606}
\end{figure}

In practical applications, FL is often deployed in environments with varying numbers of devices.
Thus,    scalability is crucial for determining whether FL-OA can maintain stable performance as the participant population changes.
 To evaluate the scalability of FL-OA, we conduct experiments on  CIFAR10 and CIFAR100  under the DIR(0.6) setting, where 50\% of the devices are malicious. 
The number of devices is set to 100, 120, 140, 160, and 200, respectively.
Fig. \ref{311606} illustrates the scalability of FL-OA under Gaussian attacks. 
The results show that, as the number of devices increases from 100 to 200, the performance of FL-OA decreases by 5.67\% on CIFAR10 and by 3.88\% on CIFAR100. 
This is mainly because a larger number of devices leads to a more dispersed data distribution and less data available to each device, making it more difficult for local models to fit their data.
Notably, when the number of devices increases to 160, the performance degradation of FL-OA remains relatively small, indicating that FL-OA exhibits good scalability.

 \subsubsection{Stability of FL-OA}

\begin{figure}[t]
  \centering
\begin{scriptsize}
  \subfloat[Gaussian attacks]
  {
      \label{22523161}  \includegraphics[width=0.44\linewidth]{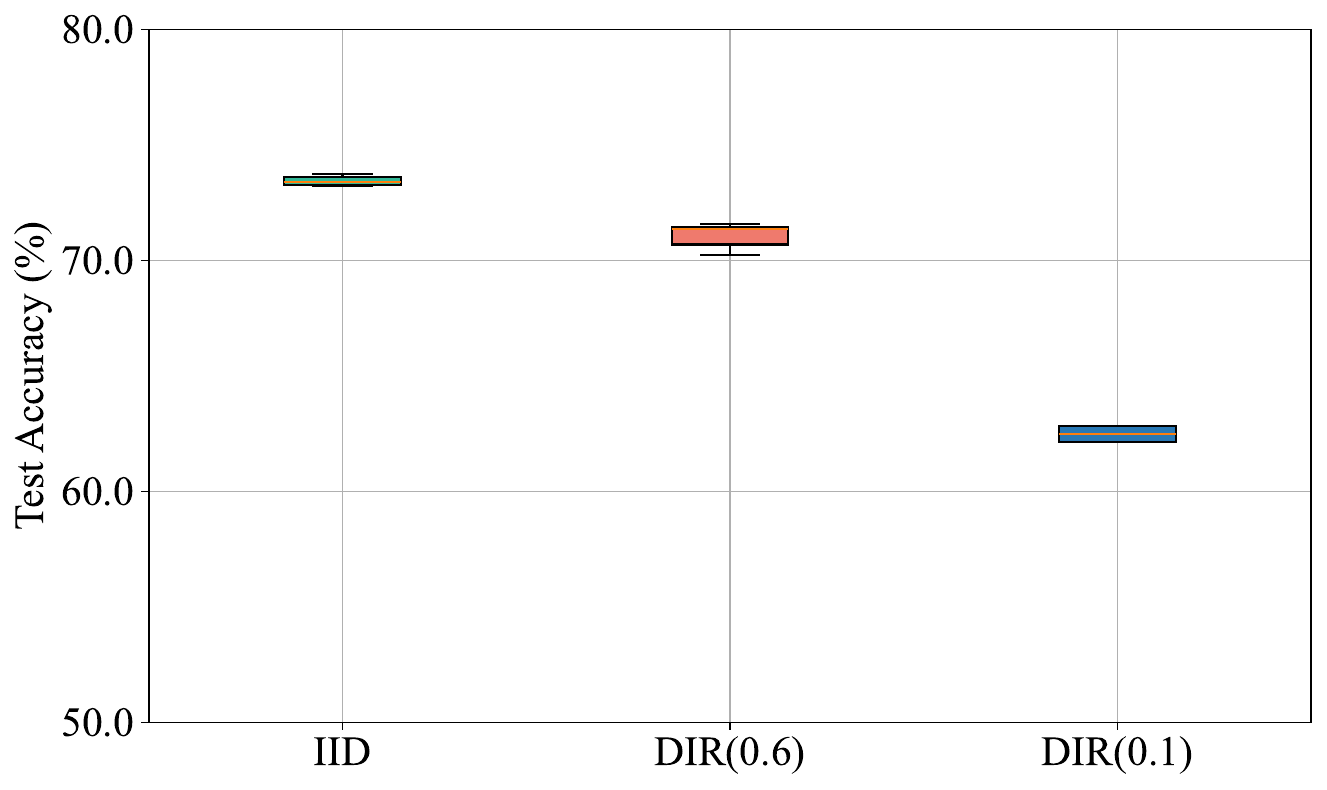}

  }
  \subfloat[Neurotoxin attacks]
  {
      \label{22523162}  \includegraphics[width=0.44\linewidth]{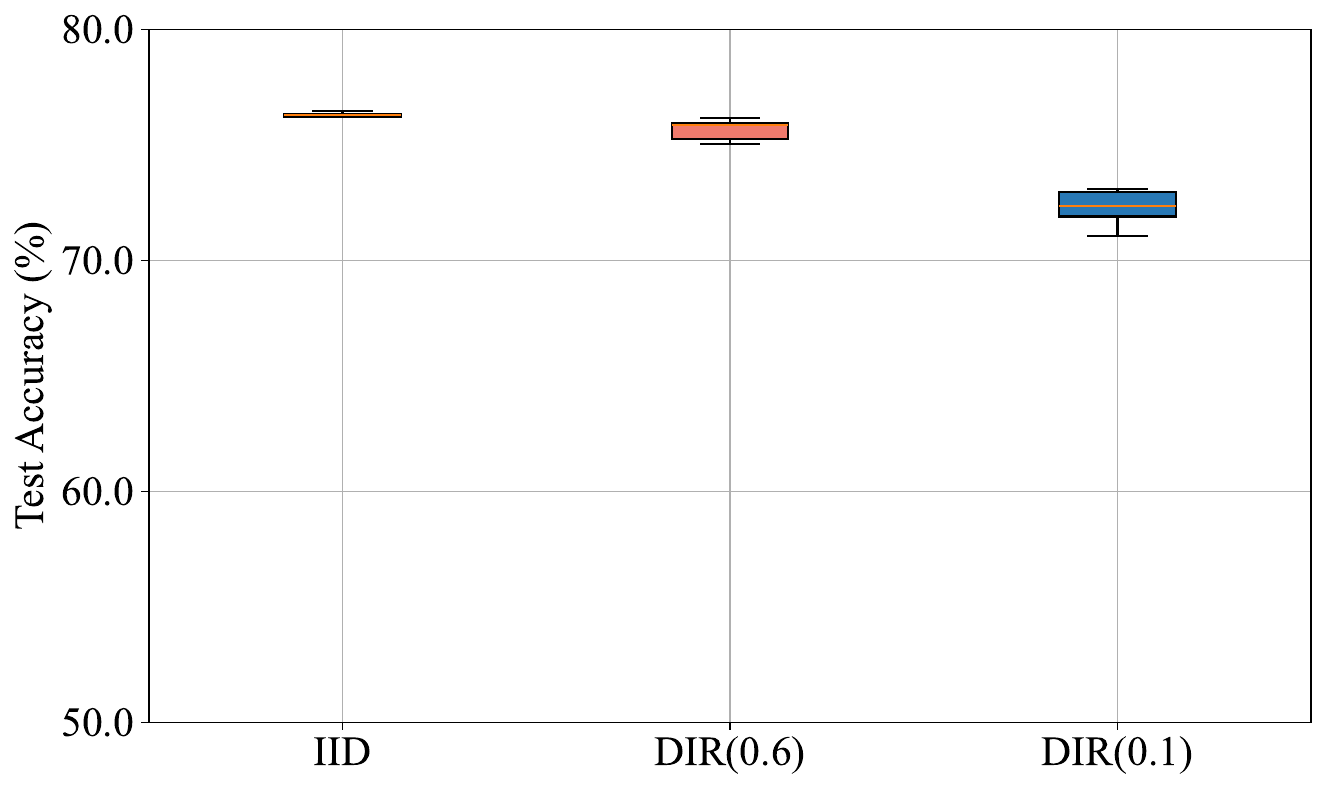}
  }
\caption{Stability of FL-OA under both Gaussian and Neurotoxin attacks on CIFAR10.}
\label{2252316}
\end{scriptsize}
\vspace{-10pt}
\end{figure}

As observed from Fig. \ref{2252316}, the performance of FL-OA exhibits some fluctuations. 
Thus, it is necessary to  evaluate its stability through multiple randomized experiments. 
To this end, we conduct experiments on   CIFAR10, where 50\% of the device are malicious, under three data distribution settings: IID, DIR(0.6), and DIR(0.1).
 Each experiment is repeated 10 times.
The boxplots in Fig. \ref{2252316} visually demonstrate the stability of FL-OA.
It can be observed that FL-OA exhibits only slight performance fluctuations under attack scenarios, and no outliers are observed in the these results, indicating that FL-OA has good stability.

\subsubsection{Complexity Analysis}
\begin{table}[t]
  \caption{Time and communication complexity in FL-OA.} 
  \vspace{-10pt}
\centering
\scalebox{0.74}{
\begin{tabular}{c|ccc}
\hline Notation & Task Server&Outsourced Server&Device \\
\hline Communication complexity &$\mathcal{O}(3nT \cdot |\mathbb{X}|+ |\mathbb{Y}|)$&$\mathcal{O}(nT \cdot |\mathbb{X}| +|\mathbb{Y}|)$&$\mathcal{O}(2T\cdot |\mathbb{X}|)$\\
 Time complexity & $\mathcal{O}(Tn \cdot \textbf{T}_{\textit{agg}})$&$\mathcal{O}(Tn \cdot \textbf{T}_{\textit{aud}} +T\cdot \textbf{T}_{\textit{train}})$&$\mathcal{O}(T \cdot \textbf{T}_{\textit{train}})$\\
\hline
\end{tabular}}
\vspace{-5pt}
\label{841651}
\end{table}

To evaluate the computational overhead of FL-OA, we analyze the time  and communication complexity of each entity, as shown in Table \ref{841651}.
Specifically, for TS, the time complexity is $\mathcal{O}(Tn \cdot \textbf{T}_{\textit{agg}})$,
where  $n \cdot \textbf{T}_{\textit{agg}}$ is the time required to aggregate the updates from $n$ devices, and $T$ denotes the number of communication rounds.
In each round, TS receives model updates from $n$ selected devices and sends the global model back to them, resulting in a communication complexity  of $\mathcal{O}(2nT \cdot |\mathbb{X}|)$, where $|\mathbb{X}|$ denotes  the amount of parameters uploaded by a device in one  round.
In addition, OS sends trust scores and normalized model updates  to  TS, which incurs an extra communication cost of $\mathcal{O}(nT \cdot |\mathbb{X}| +|\mathbb{Y}|)$, where $|\mathbb{Y}|$ denotes   the number of trust scores. 
Thus, the communication complexity of TS is $\mathcal{O}(3nT \cdot |\mathbb{X}|+ |\mathbb{Y}|)$.
For OS, the time complexity is $\mathcal{O}(Tn \cdot \textbf{T}_{\textit{aud}} +T\cdot \textbf{T}_{\textit{train}})$, where $n \cdot \textbf{T}_{\textit{aud}}$ is the time required for OS to audit $n$ devices, and $\textbf{T}_{\textit{train}}$ denotes  time spent by the model training  in one communication round.
Thus, the communication complexity of OS is represented as $\mathcal{O}(nT \cdot |\mathbb{X}| +|\mathbb{Y}|)$.
For each device, the time and communication complexity are $\mathcal{O}(T \cdot \textbf{T}_{\textit{train}})$ and $\mathcal{O}(2T\cdot |\mathbb{X}|)$, respectively.
We can observe that the time complexity of  TS is significantly reduced due to  OS helping the TS to audit model updates.
Although the introduction of OS increases the communication overhead of TS by $\mathcal{O}(nT \cdot |\mathbb{X}| +|\mathbb{Y}|)$, this overhead is affordable in practice.
Specifically, $|\mathbb{Y}|$ denotes the trust scores transmitted between OS and TS, which are negligible compared to the parameters of model update, i.e., $|\mathbb{Y}| \ll |\mathbb{X}|$.
Furthermore, transmitting model updates between TS and OS is feasible in the real world.
Therefore, the additional communication overhead introduced by OS is acceptable for FL-OA.

To further evaluate the  transmission delay introduced by FL-OA between the TS and the OS, we conduct measurements in  practical Wide-Area Network (WAN) scenarios.
 In the experiments, the TS and OS are deployed at two   locations with a physical distance of approximately 12.51 km. 
 \sloppy
\begin{wrapfigure}{t}{0.19\textwidth}
     \vspace{-11pt}
  \centering
 \includegraphics[width=0.4\columnwidth]{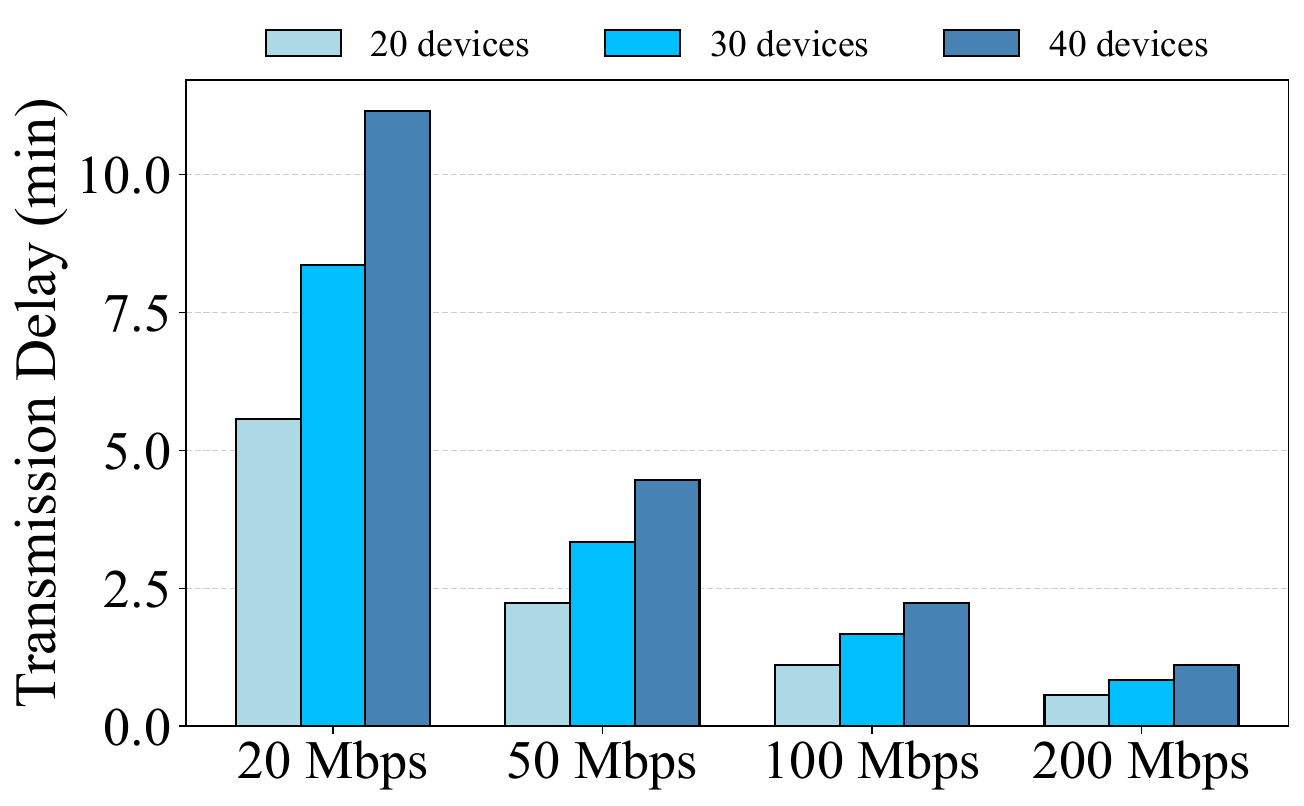}
     \vspace{-25pt}
  \caption{The transmission delay between the TS and the OS in  FL-OA.}\label{61117271}
   \vspace{-10pt}
\end{wrapfigure} 
 To emulate WAN scenarios under different  bandwidth conditions, we limit both the uplink and downlink transmission bandwidths  between the TS and the OS to 20 Mbps, 50 Mbps, 100 Mbps, and 200 Mbps, respectively.
Additionally, the size of each model update is approximately 41.8 MB.
We record the total transmission delay required for the TS to transmit model updates from 20, 30, and 40  selected devices to the OS for outsourced auditing. 
The experimental results are shown in Fig. \ref{61117271}.
We observe that the transmission delay gradually decreases as the transmission bandwidth between the TS and the OS increases. 
Since each model update is approximately 41.8 MB, transmitting model updates from 20, 30, and 40 devices involves approximately 836 MB, 1254 MB, and 1672 MB of data, respectively.
Thus, under practical WAN scenarios with tens-to-hundreds of Mbps bandwidth, the  transmission delay between the TS and the OS  reaches the minute level.

\begin{table}[t]
  \caption{Measured peak memory footprint (MB) for different schemes.} 
  \vspace{-10pt}
\centering
\scalebox{0.8}{
\begin{tabular}{c|ccccccc}
\hline / & FedAvg& Krum & FLTrust & FL-Auditor & AlignIns& FLgym & FL-OA\\
\hline TS  & 1274.5 & 1279.0 &1318.4  & 1317.2 & 2608.6 & 1294.3 & 1278.5\\
 OS & -& - & -& 1365.7  & -&  - & 2622.4 \\
\hline
\end{tabular}}
\vspace{-5pt}
\label{6171327}
\end{table}

To evaluate the memory overhead of FL-OA, we measure the peak memory footprint on the servers and compare it with that of other schemes, as shown in Table \ref{6171327}.
The experiments are conducted on the CIFAR10 dataset, where 30 clients are selected to participate in training in each  round, and the other hyper-parameters are set to their default values.
Notably, FedAvg, Krum, FLTrust, AlignIns, and FLgym do not involve outsourced auditing.
Thus, their update auditing and aggregation  are executed on the TS.
We observe that, on the TS, most schemes exhibit comparable peak memory footprints, except for AlignIns.
This is because AlignIns compute and store an additional sign vector for each model update, thereby incurring a higher memory footprint.
For FL-Auditor and FL-OA, the auditing procedure is executed on the OS. 
The results show that the memory footprint of FL-OA  is higher than that of FL-Auditor on the OS. 
This is because FL-OA computes additional $\operatorname{PII}$ vectors for model updates to identify critical parameters, which introduces extra memory overhead for storing these vectors.

\begin{figure}[t]
  \centering   
    \subfloat[Server-side runtime]
      {\label{2232009}
  \centering \includegraphics[width=0.45\linewidth]{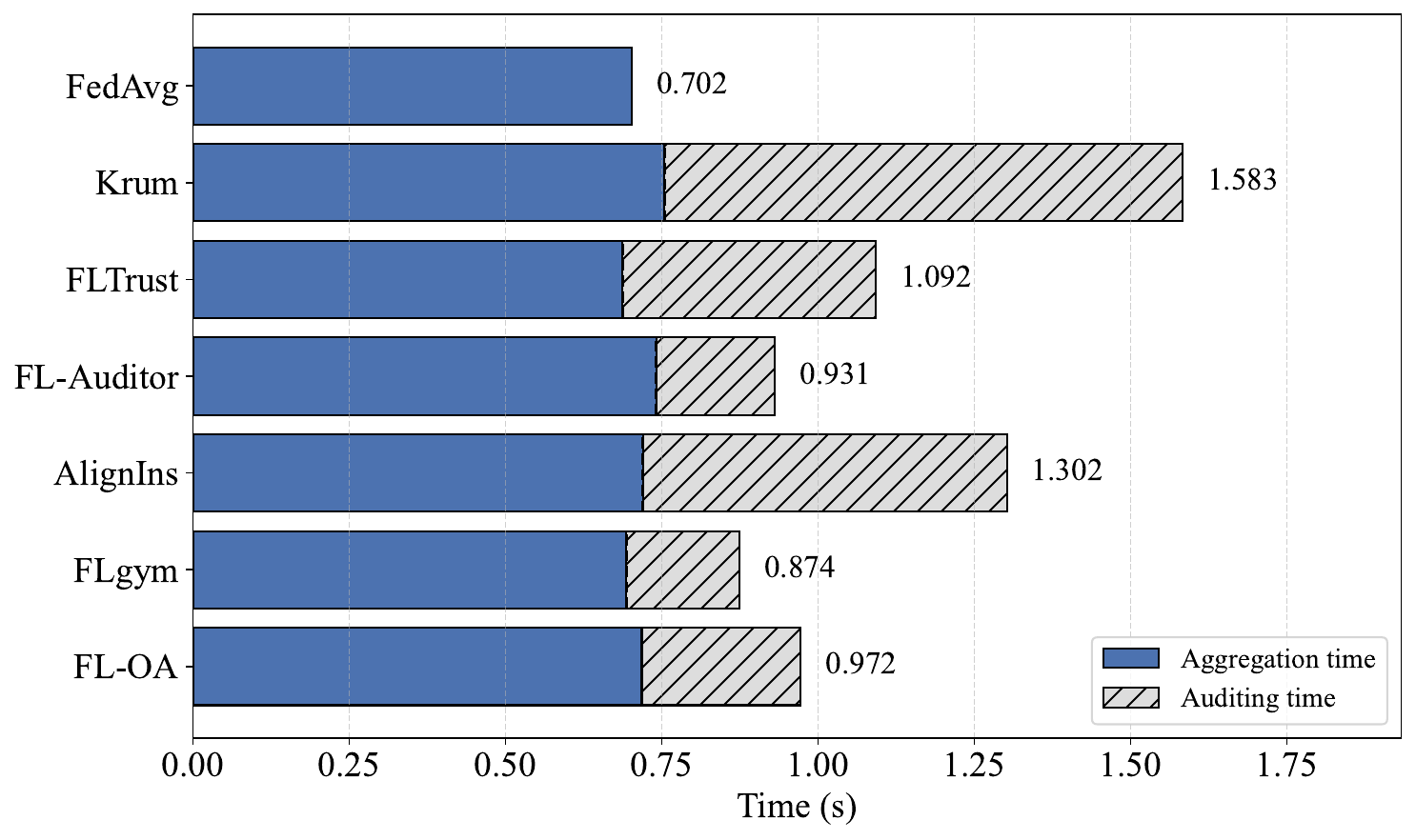}
  }
      \subfloat[Device-side runtime]
      {\label{2232008}
  \centering \includegraphics[width=0.462\linewidth]{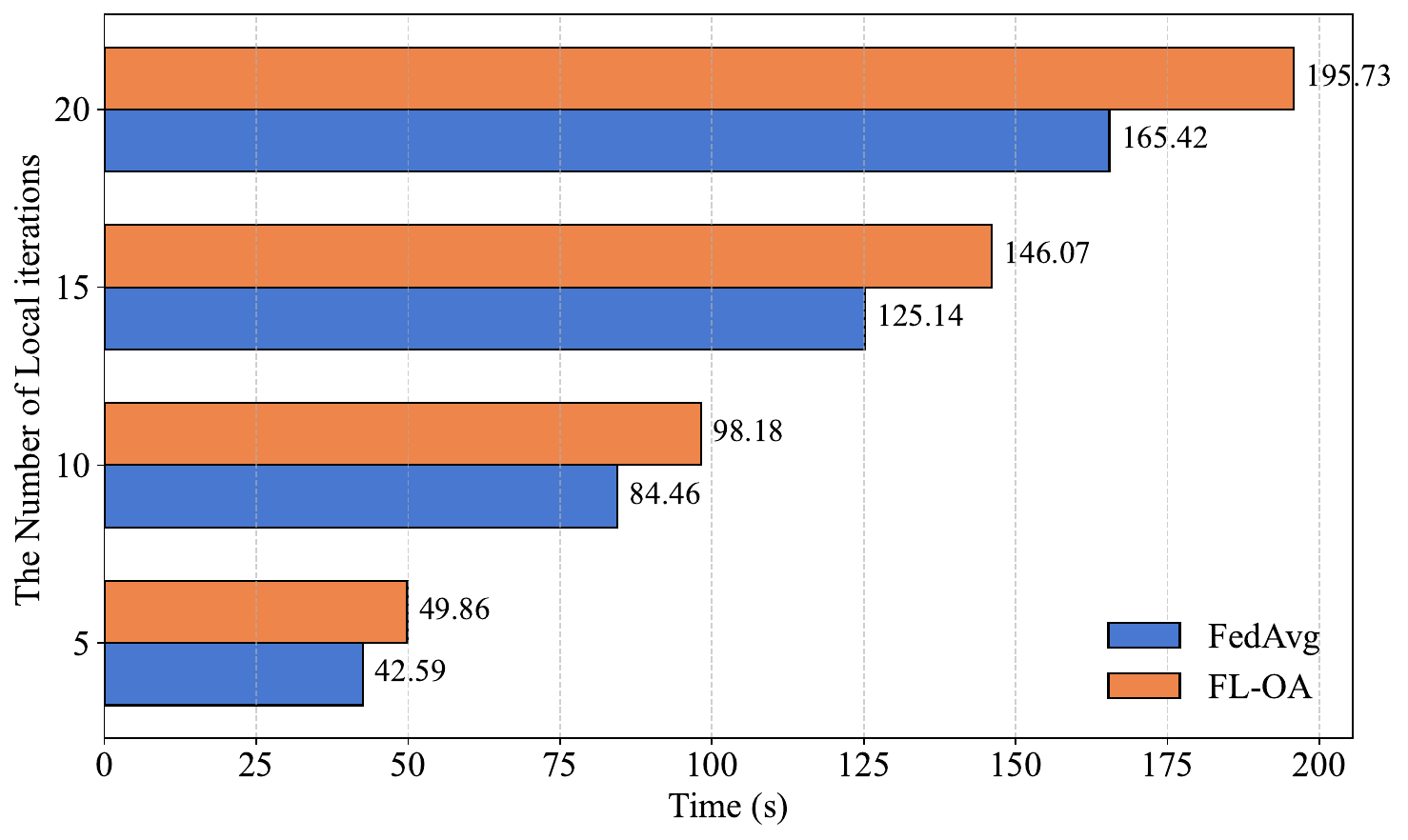}
  }
\caption{The runtime for different schemes in FL. (a) Server-side runtime for   different schemes. (b) Device-side runtime for FedAvg and FL-OA.}
\label{22320091}
\vspace{-10pt}
\end{figure}

\begin{table}[t]
  \caption{Actual per-round runtime   for  different schemes.} 
  \vspace{-10pt}
\centering
\scalebox{0.8}{
\begin{tabular}{c|ccccccc}
\hline / & FedAvg& Krum & FLTrust & FL-Auditor & AlignIns& FLgym & FL-OA\\
\hline Time (min)  & 4.07 & 4.13 & 4.09 & 7.42 & 4.11 & 4.09 & 7.54 \\  
\hline
\end{tabular}}
\vspace{-5pt}
\label{6171835}
\end{table}

To  evaluate the efficiency of FL-OA, we measure its  server-side runtime and device-side runtime , as shown in Fig. \ref{22320091}. 
The Server-side runtime  includes the time required for  update auditing, denoted as  $\textbf{T}_{\textit{aud}}$ and  update aggregation, denoted as $\textbf{T}_{\textit{agg}}$.
Notably, the runtimes measured in the evaluation does not include data transmission delay.
Specifically, on  CIFAR10 dataset, we compare the server-side runtime of FL-OA with that of FedAvg \cite{mcmahan2017communication} and other robust FL schemes under   same hardware environments.
In each round, 30 devices are selected to participate in training.
As shown in Fig. \ref{2232009}, the aggregation time $\textbf{T}_{\textit{agg}}$ of all robust schemes is comparable to that of FedAvg, since $\textbf{T}_{\textit{agg}}$ mainly corresponds to the cost of update aggregation.
For the auditing time,  FL-OA achieves a lower   time cost compared with Krum and AlignIns, indicating that the  auditing time introduced by FL-OA is acceptable.
In addition, we evaluate the device-side runtime of FL-OA. 
Fig. \ref{2232008} presents the device-side runtime of FL-OA and FedAvg under different numbers of local iterations.
The results show that  FL-OA introduces additional computational overhead compared with FedAvg. 
This is mainly because the gradient ascent step and the correction term are incorporated into local optimization,  enhancing the consistency of model updates among devices.

It is worth noting that we  measure the actual  per-round runtime of all schemes under the same hardware  and bandwidth settings. 
The per-round runtime consists ofthe device-side runtime, data transmission delay, and server-side runtime, as shown in Table \ref{6171835}.
We observe that FL-Auditor and FL-OA, both of which adopt outsourced auditing, require longer per-round runtime than the other schemes.
This is mainly caused by the additional data transmission delay between the task server and the outsourced server. 
Although FL-OA introduces extra runtime overhead, it significantly improves the robustness of the FL system against Byzantine attacks.

\section{CONCLUSION}

To defend against Byzantine attacks, FL-OA introduces an outsourced server that possesses an additional  root dataset to perform update auditing. 
This design enables the task server to conduct Byzantine-robust aggregation without directly acquiring  the   dataset,  thereby alleviating the difficulty faced by existing defense methods in satisfying the  additional dataset requirement.
 Furthermore, FL-OA introduces a  gradient ascent step and a correction term into the local training to regulate the direction of  model updates, which mitigates the divergence among benign updates caused by Non-IID data. 
To address the curse of dimensionality in auditing, FL-OA further designs a parameter importance indicator to extract critical
parameters of  model update for auditing. 
 Experimental results demonstrate that FL-OA can effectively defend against Byzantine attacks under various data distributions.

Notably, the effectiveness of FL-OA   relies on the assumption that the  OS  is honest and uncompromised.
This assumption is reasonable because the OS can be regarded as an independent third-party auditing entity, whose behavior is generally constrained by legal regulations and company reputation.
Nevertheless, we acknowledge that collusion between the OS and malicious intelligent devices may   occur in practical scenarios, which could lead to the following security risks.
\textit{(i)}  OS may intentionally assign higher trust scores to malicious devices, thereby  increasing the impact  of malicious updates on the global model. 
\textit{(ii)} OS may also reduce the trust scores of benign devices, causing useful benign updates to be weakened.
To mitigate the above risks, a viable  improvement is to extend FL-OA by introducing multiple independent outsourced servers.
By averaging the trust scores provided by these servers, FL-OA can reduce its reliance on a single OS and enhance its robustness against potential collusion between a single OS and malicious devices.

In addition to the  collusion issue, FL-OA  faces potential privacy leakage risks. 
Specifically, attackers may exploit the model updates submitted by  devices to launch privacy attacks, such as membership inference attacks, attribute inference attacks, and data reconstruction attacks, thereby illegally inferring or  recovering the devices' raw training data. 
For example, data reconstruction attacks \cite{NEURIPS2019_60a6c400}\cite{10495167} can recover original training samples from the model updates submitted by  devices. 
More importantly, poisoning attacks and privacy threats often coexist in practical FL scenarios \cite{10488898}.
Given the resource-constrained nature of intelligent devices,  it is imperative to integrate privacy-preserving mechanisms into
Byzantine attack detection in FL.
For example, it is worth investigating how to incorporate differential privacy into local training and  control the privacy budget and noise magnitude, thereby achieving an effective trade-off between privacy protection and global model performance.
Thus, in future work, we  will investigate lightweight privacy-preserving techniques suitable for intelligent devices to mitigate the privacy leakage risks during model update auditing.

\bibliographystyle{IEEEtran}
\bibliography{ref}

@inproceedings{FLtrust,
author = {Cao, Xiaoyu and Fang, Minghong and Liu, Jia and Gong, Neil},
year = {2021},
title = {FLTrust: Byzantine-robust Federated Learning via Trust Bootstrapping},
	booktitle = {Proc.  NDSS Symposium},
}

@inproceedings{1093,
	author = {Verleysen, Michel and Fran{\c{c}}ois, Damien},
	booktitle = {Proc. Comput. Intell. Bioinspired Syst.},
	pages = {758--770},
	title = {The Curse of Dimensionality in Data Mining and Time Series Prediction},
	year = {2005}}

@inproceedings{NIPS2017f4b9ec30,
  title={Machine learning with adversaries: Byzantine tolerant gradient descent},
  author={Blanchard, Peva and El Mhamdi, El Mahdi and Guerraoui, Rachid and Stainer, Julien},
  booktitle={Proc. Adv. Neural Inf. Process. Syst.},
  year={2017},
pages = {119--129},
}

@ARTICLE{10458320,
  author={Mu, Xutong and others},
  journal={IEEE Trans. Dependable Secure Comput.}, 
  title={FedDMC: Efficient and Robust Federated Learning via Detecting Malicious Clients}, 
  year={2024},
  volume={21},
  number={6},
  pages={5259-5274},
}

@ARTICLE{10475552,
  author={Zhang, Zhuangzhuang and Wu, Libing and He, Debiao and Li, Jianxin and Lu, Na and Wei, Xuejiang},
  journal={IEEE Trans. Sustainable Comput.}, 
  title={Using Third-Party Auditor to Help Federated Learning: An Efficient Byzantine-Robust Federated Learning}, 
  year={2024},
  volume={9},
  number={6},
  pages={848-861},
}

@inproceedings{sun2023fedspeed,
  title={Fedspeed: Larger local interval, less communication round, and higher generalization accuracy},
  author={Sun, Yan and Shen, Li and Huang, Tiansheng and Ding, Liang and Tao, Dacheng},
  booktitle={Proc. Int. Conf. Learn. Represent.},
  year={2023}
}

@article{10286887,
	author = {Zhou, Tailin and Zhang, Jun and Tsang, Danny H. K.},
	journal = {IEEE Trans. Mob. Comput.},
	number = {6},
	pages = {6731-6742},
	title = {FedFA: Federated Learning With Feature Anchors to Align Features and Classifiers for Heterogeneous Data},
	volume = {23},
	year = {2024},
}

@article{miao2022privacy,
	author = {Miao, Yinbin and Liu, Ziteng and Li, Hongwei and Choo, KimKwangRaymond and Deng, RobertH},
  journal={IEEE Trans. Inf. Forensics Secur.}, 
	pages = {2848--2861},
	title = {Privacy-preserving Byzantine-robust federated learning via blockchain systems},
	volume = {17},
	year = {2022}}

@inproceedings{mcmahan2017communication,
	author = {McMahan, Brendan and Moore, Eider and Ramage, Daniel and Hampson, Seth and Arcas, BlaiseAguera},
  booktitle={Proc. Artif. Intell. Statist.},
	pages = {1273--1282},
	title = {Communication-efficient learning of deep networks from decentralized data},
	year = {2017}}

@inproceedings{zhao2022penalizing,
	author = {Zhao, Yang and Zhang, Hao and Hu, Xiuyuan},
 booktitle={Proc. Int. Conf. Mach. Learn.},
	pages = {26982--26992},
	title = {Penalizing gradient norm for efficiently improving generalization in deep learning},
	year = {2022}}

@article{acar2021federated,
	author = {Acar, DurmusAlpEmre and Zhao, Yue and Navarro, RamonMatas and Mattina, Matthew and Whatmough, PaulN and Saligrama, Venkatesh},
	journal = {arXiv preprint arXiv:2111.04263},
	title = {Federated learning based on dynamic regularization},
	year = {2021}}

@inproceedings{li2020federated,
	author = {Li, Tian and Sahu, Anit Kumar and Zaheer, Manzil and Sanjabi, Maziar and Talwalkar, Ameet and Smith, Virginia},
	booktitle = {Proc. Mach. Learn. Syst.},
	pages = {429--450},
	title = {Federated optimization in heterogeneous networks},
	volume = {2},
	year = {2020}}

@article{wang2022threats,
	author = {Wang, Zhibo and Ma, Jingjing and Wang, Xue and Hu, Jiahui and Qin, Zhan and Ren, Kui},
	journal = {ACM Comput. Surv.},
	number = {7},
	pages = {1--36},
	publisher = {ACM New York, NY},
	title = {Threats to training: A survey of poisoning attacks and defenses on machine learning systems},
	volume = {55},
	year = {2022}}

@ARTICLE{kumar2023impact,
  author={Kumar, Kummari Naveen and Mohan, Chalavadi Krishna and Cenkeramaddi, Linga Reddy},
  journal={IEEE Trans. Pattern Anal. Mach. Intell.}, 
  title={The Impact of Adversarial Attacks on Federated Learning: A Survey}, 
  year={2024},
  volume={46},
  number={5},
  pages={2672-2691},
}

@ARTICLE{lu2024federated,
  author={Lu, Zili and Pan, Heng and Dai, Yueyue and Si, Xueming and Zhang, Yan},
  journal={IEEE Internet Things J.}, 
  title={Federated Learning With Non-IID Data: A Survey}, 
  year={2024},
  volume={11},
  number={11},
  pages={19188-19209},
}

@ARTICLE{huang2024federated,
  author={Huang, Wenke and others},
  journal={IEEE Trans. Pattern Anal. Mach. Intell.}, 
  title={Federated Learning for Generalization, Robustness, Fairness: A Survey and Benchmark}, 
  year={2024},
  volume={46},
  number={12},
  pages={9387-9406},
}

@article{hanzely2020federated,
  title={Federated learning of a mixture of global and local models},
  author={Hanzely, Filip and Richt{\'a}rik, Peter},
  journal={arXiv preprint arXiv:2002.05516},
  year={2020}
}

@article{krizhevsky2009learning,
  title={Learning multiple layers of features from tiny images},
  author={Krizhevsky, Alex and Hinton, Geoffrey and others},
  year={2009},
  publisher={Toronto, ON, Canada}
}

@article{cho2020client,
  title={Client selection in federated learning: Convergence analysis and power-of-choice selection strategies},
  author={Cho, Yae Jee and Wang, Jianyu and Joshi, Gauri},
  journal={arXiv preprint arXiv:2010.01243},
  year={2020}
}

@inproceedings{wang2020tackling,
  title={Tackling the objective inconsistency problem in heterogeneous federated optimization},
  author={Wang, Jianyu and Liu, Qinghua and Liang, Hao and Joshi, Gauri and Poor, HVincent},
  booktitle={Proc. Adv. Neural Inf. Process. Syst.},
  volume={33},
  pages={7611--7623},
  year={2020}
}

@inproceedings{fraboni2021free,
  title={Free-rider attacks on model aggregation in federated learning},
  author={Fraboni, Yann and Vidal, Richard and Lorenzi, Marco},
  booktitle={Proc. Artif. Intell. Statist.},
  pages={1846--1854},
  year={2021},
}

@inproceedings{yin2018byzantine,
  title={Byzantine-robust distributed learning: Towards optimal statistical rates},
  author={Yin, Dong and Chen, Yudong and Kannan, Ramchandran and Bartlett, Peter},
 booktitle={Proc. Int. Conf. Mach. Learn.},
  pages={5650--5659},
  year={2018},
}

@article{targ2016resnet,
  title={Resnet in resnet: Generalizing residual architectures},
  author={Targ, Sasha and Almeida, Diogo and Lyman, Kevin},
  journal={arXiv preprint arXiv:1603.08029},
  year={2016}
}

@inproceedings{shejwalkar2021manipulating,
  title={Manipulating the byzantine: Optimizing model poisoning attacks and defenses for federated learning},
  author={Shejwalkar, Virat and Houmansadr, Amir},
booktitle = {Proc.  NDSS Symposium},
  year={2021}
}

@article{yu2020salvaging,
  title={Salvaging federated learning by local adaptation},
  author={Yu, Tao and Bagdasaryan, Eugene and Shmatikov, Vitaly},
  journal={arXiv preprint arXiv:2002.04758},
  year={2020}
}

@ARTICLE{9451544,
  author={Li, Zewen and Liu, Fan and Yang, Wenjie and Peng, Shouheng and Zhou, Jun},
  journal={IEEE Trans. Neural Networks Learn. Syst.}, 
  title={A Survey of Convolutional Neural Networks: Analysis, Applications, and Prospects}, 
  year={2022},
  volume={33},
  number={12},
  pages={6999-7019},
}

@INPROCEEDINGS{9798217,
  author={Zheng, Xu and Dong, Qihao and Fu, Anmin},
  booktitle={Proc.  Conf. Comput. Commun. Workshops}, 
  title={WMDefense: Using Watermark to Defense Byzantine Attacks in Federated Learning}, 
  year={2022},
  pages={1-6},
}

@ARTICLE{10713463,
  author={Dong, Qihao and others},
  journal={IEEE Trans. Inf. Forensics Secur.}, 
  title={CareFL: Contribution Guided Byzantine-Robust Federated Learning}, 
  year={2024},
  volume={19},
  number={},
  pages={9714-9729},
}

@article{xia2024byzantine,
  title={Byzantine-resilient secure aggregation for federated learning without privacy compromises},
  author={Xia, Yue and Hofmeister, Christoph and Egger, Maximilian and Bitar, Rawad},
  journal={arXiv preprint arXiv:2405.08698},
  year={2024}
}

@inproceedings{andriushchenko2022towards,
  title={Towards understanding sharpness-aware minimization},
  author={Andriushchenko, Maksym and Flammarion, Nicolas},
 booktitle={Proc. Int. Conf. Mach. Learn.},
  pages={639--668},
  year={2022},
}

@article{lin2016dirichlet,
  title={On the dirichlet distribution},
  author={Lin, Jiayu},
  journal={Master’s thesis, Dept. Math. Statist., Queen's Univ.},
  volume={40},
  year={2016}
}

@article{zhao2018federated,
  title={Federated learning with non-iid data},
  author={Zhao, Yue and Li, Meng and Lai, Liangzhen and Suda, Naveen and Civin, Damon and Chandra, Vikas},
  journal={arXiv preprint arXiv:1806.00582},
  year={2018}
}

@inproceedings{qu2022generalized,
  title={Generalized federated learning via sharpness aware minimization},
  author={Qu, Zhe and Li, Xingyu and Duan, Rui and Liu, Yao and Tang, Bo and Lu, Zhuo},
 booktitle={Proc. Int. Conf. Mach. Learn.},
  pages={18250--18280},
  year={2022},
}

@inproceedings{beyer1999nearest,
  title={When is “nearest neighbor” meaningful?},
  author={Beyer, Kevin and Goldstein, Jonathan and Ramakrishnan, Raghu and Shaft, Uri},
  booktitle={Proc. Int. Conf. Database Theory},
  pages={217--235},
  year={1999},
}

@inproceedings{huang2023multi,
  title={Multi-metrics adaptively identifies backdoors in federated learning},
  author={Huang, Siquan and Li, Yijiang and Chen, Chong and Shi, Leyu and Gao, Ying},
  booktitle={Proc. IEEE/CVF Conf. Comput. Vis. Pattern Recognit. },
  pages={4652--4662},
  year={2023}
}

@inproceedings{krauss2023mesas,
  title={Mesas: Poisoning defense for federated learning resilient against adaptive attackers},
  author={Krau{\ss}, Torsten and Dmitrienko, Alexandra},
  booktitle={Proc. ACM Conf. Comput. Commun. Secur.},
  pages={1526--1540},
  year={2023}
}

@ARTICLE{11202428,
  author={Zhang, Hongliang and Yu, Zhongyuan and Wang, Guijuan and Xu, Fenghua and Zhang, Yongzhao and Hu, Chunqiang and Wang, Xiaofen and Yu, Jiguo},
  journal={IEEE Trans. Dependable Secure Comput.}, 
  title={Toward Model-Contrastive Federated Learning With Lightweight Privacy Preservation and Poisoning Attack Detection}, 
  year={2026},
  volume={23},
  number={2},
  pages={1830-1846},
  doi={10.1109/TDSC.2025.3620529}}

@inproceedings{zhang2022neurotoxin,
  title={Neurotoxin: Durable backdoors in federated learning},
  author={Zhang, Zhengming and others},
 booktitle={Proc. Int. Conf. Mach. Learn.},
  pages={26429--26446},
  year={2022},
}

@inproceedings{fang2023vulnerability,
  title={On the vulnerability of backdoor defenses for federated learning},
  author={Fang, Pei and Chen, Jinghui},
  booktitle={Proc. AAAI Conf. Artif. Intell.},
  volume={37},
  number={10},
  pages={11800--11808},
  year={2023}
}

@inproceedings{xu2025detecting,
  title={Detecting backdoor attacks in federated learning via direction alignment inspection},
  author={Xu, Jiahao and Zhang, Zikai and Hu, Rui},
  booktitle={Proc. IEEE/CVF Conf. Comput. Vis. Pattern Recognit.},
  pages={20654--20664},
  year={2025}
}

@ARTICLE{10648998,
  author={Liu, Jingwei and Wu, Yufeng and Du, Wei and Sun, Rong and Xu, Guangxia and Liu, Lei and Wu, Celimuge},
  journal={IEEE Trans. Consum. Electron.}, 
  title={Byzantine-Robust Hierarchical Aggregation for Cross-Device Federated Learning in Consumer IoT}, 
  year={2025},
  volume={71},
  number={2},
  pages={6359-6370},
}

@ARTICLE{11421423,
  author={Guo, Weian and others},
  journal={IEEE Trans. Consum. Electron.}, 
  title={Byzantine-Resilient Federated Learning with Trust-Aware Task Scheduling for Heterogeneous UAV Swarms}, 
  year={2026},
  volume={},
  number={},
  pages={1-1},
}

@ARTICLE{10495004,
  author={Pei, Jiaming and Xue, Rubing and Liu, Chao and Wang, Lukun},
  journal={IEEE Trans. Consum. Electron.}, 
  title={Toward Byzantine-Resilient Secure AI: A Federated Learning Communication Framework for 6G Consumer Electronics}, 
  year={2024},
  volume={70},
  number={3},
  pages={5719-5728},
}

@ARTICLE{10492865,
  author={Pei, Jiaming and Liu, Wenxuan and Li, Jinhai and Wang, Lukun and Liu, Chao},
  journal={IEEE Trans. Consum. Electron.}, 
  title={A Review of Federated Learning Methods in Heterogeneous Scenarios}, 
  year={2024},
  volume={70},
  number={3},
  pages={5983-5999},
}

@ARTICLE{10891500,
  author={Bai, Jun and Wu, Di and Zeng, Shan and Zhao, Yao and Qu, Youyang and Yu, Shui},
  journal={IEEE Trans. Consum. Electron.}, 
  title={Non-IID Free Federated Learning With Fuzzy Optimization for Consumer Electronics Systems}, 
  year={2025},
  volume={71},
  number={2},
  pages={7032-7044},
}

@article{mackiewicz1993principal,
  title={Principal components analysis (PCA)},
  author={Ma{\'c}kiewicz, Andrzej and Ratajczak, Waldemar},
  journal={Computers \& Geosciences},
  volume={19},
  number={3},
  pages={303--342},
  year={1993},
  publisher={Elsevier}
}

@article{van2008visualizing,
  title={Visualizing data using t-SNE.},
  author={Van der Maaten, Laurens and Hinton, Geoffrey},
  journal={Journal of machine learning research},
  volume={9},
  number={11},
  year={2008}
}

@inproceedings{netzer2011reading,
  title={Reading digits in natural images with unsupervised feature learning},
  author={Netzer, Yuval and others},
  booktitle={NIPS Workshops.},
  volume={2011},
  number={2},
  pages={4},
  year={2011},
}

@article{darlow2018cinic,
  title={Cinic-10 is not imagenet or cifar-10},
  author={Darlow, Luke N and Crowley, Elliot J and Antoniou, Antreas and Storkey, Amos J},
   journal= {arXiv preprint arXiv:1810.03505},
  year={2018}
}

@article{le2015tiny,
  title={Tiny imagenet visual recognition challenge},
  author={Le, Yann and Yang, Xuan and others},
  journal={CS 231N},
  volume={7},
  number={7},
  pages={3},
  year={2015}
}

@ARTICLE{11298320,
  author={Xiao, Ke and Wang, Qiyuan and Anagnostopoulos, Christos},
  journal={IEEE Trans. Inf. Forensics Secur.}, 
  title={FLgym: Toward Robust and Byzantine-Resilient Federated Learning}, 
  year={2026},
  volume={21},
  number={},
  pages={404-416},
}

@article{uddin2025systematic,
  title={A systematic literature review of robust federated learning: Issues, solutions, and future research directions},
  author={Uddin, Md Palash and Xiang, Yong and Hasan, Mahmudul and Bai, Jun and Zhao, Yao and Gao, Longxiang},
  journal={ACM Comput. Surv.},
  volume={57},
  number={10},
  pages={1--62},
  year={2025},
}

@inproceedings{li2021model,
  title={Model-contrastive federated learning},
  author={Li, Qinbin and He, Bingsheng and Song, Dawn},
  booktitle={Proc. IEEE/CVF Conf. Comput. Vis. Pattern Recognit. },
  pages={10713--10722},
  year={2021}
}

@inproceedings{NEURIPS2019_60a6c400,
 author = {Zhu, Ligeng and Liu, Zhijian and Han, Song},
 booktitle = {Proc. NIPS},
 title = {Deep Leakage from Gradients},
 volume = {32},
 year = {2019}
}

@ARTICLE{10495167,
  author={Sotthiwat, Ekanut and Zhen, Liangli and Zhang, Chi and Li, Zengxiang and Goh, Rick Siow Mong},
  journal={IEEE Trans. Neural Netw. Learn. Syst.}, 
  title={Generative Image Reconstruction From Gradients}, 
  year={2025},
  volume={36},
  number={1},
  pages={21-31},
}

@ARTICLE{10488898,
  author={Huang, Ren-Yi and Samaraweera, Dumindu and Chang, J. Morris},
  journal={Computer}, 
  title={Exploring Threats, Defenses, and Privacy-Preserving Techniques in Federated Learning: A Survey}, 
  year={2024},
  volume={57},
  number={4},
  pages={46-56},
}

\end{document}